\documentclass[10pt,twocolumn,letterpaper]{article}

\usepackage{iccv}
\usepackage{times}
\usepackage{epsfig}
\usepackage{graphicx}
\usepackage{amsmath}
\usepackage{amssymb}

\makeatletter
\@namedef
{ver@everyshi.sty}{}
\makeatother

\usepackage{lipsum}  
\usepackage{booktabs}  
\usepackage{xcolor}  
\usepackage{subcaption}  
\usepackage{multirow}  
\usepackage{makecell}  
\usepackage{pgfplots}  
\usepackage{tabularx}  
\usepackage[nolist,nohyperlinks]{acronym}  
\usepackage{placeins}  

\pgfplotsset{compat=newest}

\graphicspath{{./graphics/}}
\usetikzlibrary{calc}

\newcommand{\bp}{\mathbf{p}}

\newcommand{\bS}{\mathbf{S}}

\newcommand{\figref}[1]{Fig.~\ref{#1}}
\newcommand{\secref}[1]{Section~\ref{#1}}

\newcommand{\tabref}[1]{Table~\ref{#1}}

\makeatletter
\DeclareRobustCommand\onedot{\futurelet\@let@token\@onedot}
\def\@onedot{\ifx\@let@token.\else.\null\fi\xspace}
\def\eg{e.g\onedot} 
\def\ie{i.e\onedot}

\def\etal{et~al\onedot}

\makeatother

\newcommand{\boldparagraph}[1]{\vspace{0.15cm}\noindent{\bf #1:} }

\definecolor{darkgreen}{rgb}{0,0.7,0}
\definecolor{real_img}{RGB}{215,255,170}
\definecolor{syn_img}{RGB}{19,73,189}
\definecolor{misc_img}{RGB}{144,12,63}
\definecolor{real_plot}{RGB}{100,163,29}
\definecolor{syn_plot}{RGB}{19,73,189}
\definecolor{misc_plot}{RGB}{144,12,63}
\definecolor{cd_plot}{RGB}{27,158,119}
\definecolor{mae_plot}{RGB}{217,95,2}
\definecolor{mse_plot}{RGB}{117,112,179}
\definecolor{kitti}{RGB}{189, 0, 38}
\definecolor{nuscenes}{RGB}{254, 204, 92}
\definecolor{carla}{RGB}{116, 196, 118}
\definecolor{builder}{RGB}{0, 109, 44}
\definecolor{misc1}{RGB}{150, 150, 150}
\definecolor{misc2}{RGB}{82, 82, 82}
\definecolor{misc3}{RGB}{0, 0, 0}

\newcommand{\Real}{\textit{Real}}
\newcommand{\Syn}{\textit{Synthetic}}
\newcommand{\Misc}{\textit{Misc}}

\begin{acronym}
    \acro{bev}[BEV]{bird's-eye view}
    \acro{CD}[CD]{\textit{Chamfer} distance}
    \acro{CNN}[CNN]{convolutional neural network}
    \acroplural{CNN}[CNNs]{convolutional neural networks}
    \acro{C2ST}[C2ST]{Classifier Two-Sample Test}
    \acroplural{C2ST}[C2STs]{Classifier Two-Sample Tests}
    \acro{DNN}[DNN]{deep neural network}
    \acro{EMD}[EMD]{\textit{Earth Mover's} distance}
    \acro{FID}[FID]{Fr\'{e}chet Inception Distance}
    \acro{FPD}[FPD]{Fr\'{e}chet point cloud distance}
    \acro{GAN}[GAN]{generative adversarial network}
    \acroplural{GAN}[GANs]{generative adversarial networks}
    \acro{IS}[IS]{Inception Score}
    \acro{lidar}[LiDAR]{light detection and ranging}
    \acro{MAE}[MAE]{mean absolute error}
    \acro{MLP}[MLP]{multi-layer perceptron}
    \acro{MOS}[MOS]{mean opinion score}
    \acro{MSE}[MSE]{mean squared error}
    \acro{PSNR}[PSNR]{peak signal-to-noise ratio}
    \acro{SSIM}[SSIM]{structural similarity index}
    \acro{tSNE}[t-SNE]{t-Distributed Stochastic Neighbor Embedding}
    \acro{UDA}[UDA]{unsupervised domain adaptation}
\end{acronym}

\usepackage[breaklinks=true,bookmarks=false]{hyperref}
\iccvfinalcopy 

\ificcvfinal\fi

\begin{document}

\title{Quantifying point cloud realism through\\adversarially learned latent representations}
\author{
Larissa~T.~Triess$^{1,2 [\href{https://orcid.org/0000-0003-0037-8460}{0000-0003-0037-8460}]}$
\and
David~Peter$^{1 [\href{https://orcid.org/0000-0001-7950-9915}{0000-0001-7950-9915}]}$
\and
Stefan~A.~Baur$^{1 [\href{https://orcid.org/0000-0002-0735-8713}{0000-0002-0735-8713}]}$
\and
J.~Marius~Z\"ollner$^{2,3 [\href{https://orcid.org/0000-0001-6190-7202}{0000-0001-6190-7202}]}$
\and
\\
$^{1}$Mercedes-Benz AG, Research and Development, Stuttgart, Germany\\
$^{2}$Karlsruhe Institute of Technology, Karlsruhe, Germany\\
$^{3}$Research Center for Information Technology, Karlsruhe, Germany\\
{\small\tt larissa.triess@daimler.com}
}

\maketitle
\ificcvfinal\thispagestyle{empty}\fi


\begin{abstract}

Judging the quality of samples synthesized by generative models can be tedious and time consuming, especially for complex data structures, such as point clouds.
This paper presents a novel approach to quantify the realism of local regions in LiDAR point clouds.
Relevant features are learned from real-world and synthetic point clouds by training on a proxy classification task.
Inspired by fair networks, we use an adversarial technique to discourage the encoding of dataset-specific information.
The resulting metric can assign a quality score to samples without requiring any task specific annotations.

In a series of experiments, we confirm the soundness of our metric by applying it in controllable task setups and on unseen data.
Additional experiments show reliable interpolation capabilities of the metric between data with varying degree of realism.
As one important application, we demonstrate how the local realism score can be used for anomaly detection in point clouds.

\end{abstract}

\FloatBarrier


\begin{figure}[t]
    \centering
    \includegraphics[width=0.96\linewidth]{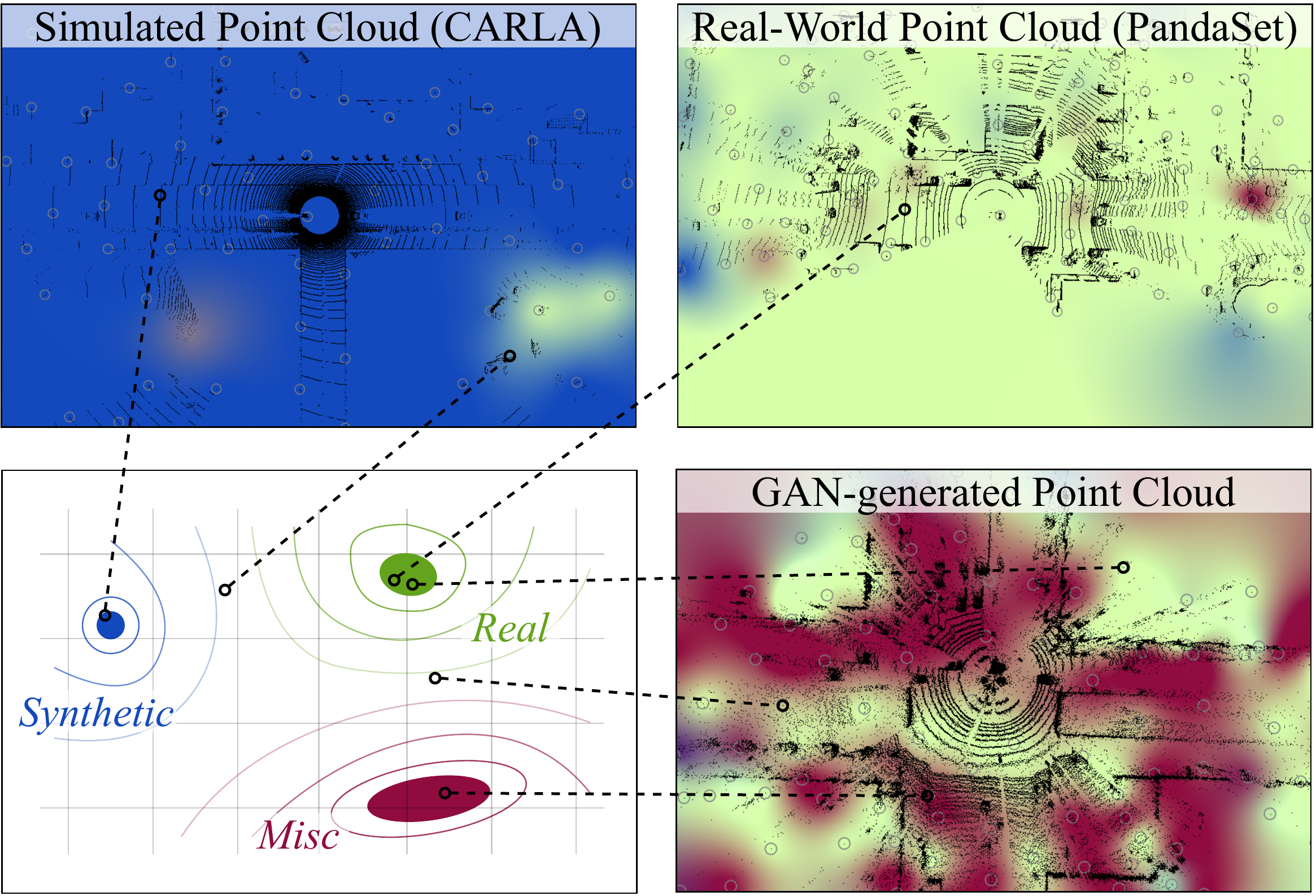}
    \caption{
    \textbf{Proposed Approach}:
    The realism measure has a tripartite understanding of the 3D-world, schematically illustrated on the bottom left.
    The other images show the color-coded metric scores at discrete query point locations (gray circles).
    Local regions in the simulated sample (top left) are largely predicted as being of synthetic origin (blue), while regions in the real-world sample (top right) are predicted as being realistic (green).
    The bottom right image shows a GAN-generated sample which has large areas with high distortion levels that neither appear realistic nor synthetic.
    The metric therefore assigns high misc scores (red).
    }
    \label{fig:eyecatcher}
\end{figure}

\section{Introduction}

Generative models, such as \acp{GAN}, are often used to synthesize realistic training data samples to improve the performance of perception networks.
Assessing the quality of such synthesized samples is a crucial part of the process which is usually done by experts, a cumbersome and time consuming approach.
Though a lot of work has been conducted to determine the quality of generated images, very little work is published about how to quantify the realism of LiDAR point clouds.
Visual inspection of such data is expensive and not very reliable given that the interpretation of 3D point data is rather unnatural for humans.
Because of their subjective nature, it is difficult to compare different generative approaches with a qualitative measure.
This paper closes the gap and introduces a quantitative evaluation for LiDAR point clouds.

In recent years, a large amount of evaluation measures for \acp{GAN} emerged~\cite{Borji2019CVIU}.
Many of them are image specific and cannot be applied to point clouds.
Existing work on generating realistic LiDAR point clouds mostly relies on qualitative measures to evaluate the generation quality.
Alternatively, some works apply annotation transfer~\cite{Sallab2019ICMLwork} or use the \acl{EMD} as an evaluation criterion~\cite{Caccia2019IROS}.
However, these methods require either annotations associated with the data or a matching target, \ie Ground Truth, for the generated sample.
Both are often not feasible when working with large-scale data generation.

The generation capabilities of the models are often directly graded by how much the performance of perception networks increases after being trained with the additional generated data.
This procedure serves as an indication for realism, but cannot solely verify realism and fidelity~\cite{Triess2019IV,Sallab2019ICMLwork}.
The improved perception accuracy is primarily caused by the larger quantity of training data.
Only measuring the performance difference in downstream perception tasks is therefore not an adequate measure for the realism of the data itself.

This paper proposes a novel learning-based approach for robust quantification of LiDAR point cloud quality.
The metric is trained to learn relevant features via a proxy classification task.
To avoid learning global scene context, we use hierarchical feature set learning to confine features locally in space.
This locality aspect additionally enables the detection and localization of anomalies and other sensor effects within the scenery.
To discourage the network from encoding dataset-specific information, we use an adversarial learning technique which enables robust quantification of unseen data distributions.
The resulting metric does not require any additional annotations.


\section{Related Work}
\label{sec:related}

\subsection{Point Cloud Quality Measures} 
\label{sec:related_pc_quality}

A decisive element to process unordered point sets is the ability to operate in a permutation-invariant fashion, such that the ordering of points does not matter.
\acf{EMD} and \acf{CD} are often used to compare such sets.
\ac{EMD} measures the distance between the two sets by attempting to transform one set into the other, while \ac{CD} measures the squared distance between each point in one set and its nearest neighbor in the other set.
Both metrics are often used to measure the reconstruction of single object shapes, like those from ShapeNet~\cite{Achlioptas2018ICLRWORK}.
Caccia~\etal~\cite{Caccia2019IROS} use the metrics as a measure of reconstruction quality on entire scenes captured with a LiDAR scanner.
This procedure is only applicable to paired translational \acp{GAN} or supervised approaches, because it requires a known target, \ie Ground Truth to measure the reconstruction error.
It is therefore not useable for unpaired or unsupervised translation on which many domain adaptation tasks are based.

In previous work, we use the \acf{MOS} testing to verify the realism of generated LiDAR point clouds~\cite{Triess2019IV}.
It was previously introduced in~\cite{Ledig2017CVPR} to provide a qualitative measure for realism in RGB images.
In contrast to~\cite{Ledig2017CVPR}, where untrained people were used to determine the realism, \cite{Triess2019IV}~requires LiDAR experts for the testing process to assure a high enough sensor domain familiarity of the test persons.

Some domain adaptation \acp{GAN} are used to improve the performance of downstream perception tasks by training the perception network with the generated data~\cite{Sixt2018FRA,Sallab2019ICMLwork}.
The generation capabilities of the \ac{GAN} are derived by how much the perception performance on the target domain changes when trained with the domain adapted data in addition or instead of the source data.
These capabilities include the ability to generate samples that are advantageous for perception network training but might not be suitable to quantify the actual generation quality.
Inspecting example images in state-of-the-art literature~\cite{Caccia2019IROS,Sallab2019ICMLwork} and closely analyzing our own generated data, we observe that the GAN-generated point clouds are usually very noisy, especially at object boundaries (see bottom right in~\figref{fig:eyecatcher}).
Therefore, we find that solely analyzing \ac{GAN} generation performance on downstream perception tasks is not enough to claim realistic generation properties.

\subsection{GAN Evaluation Measures} 
\label{sec:related_gan_measures}

The \ac{GAN} objective function is not suited as a general quality measure or as a measure of sample diversity, as it can only measure how well the generator and the discriminator perform relative to the opponent.
Therefore, a considerable amount of literature deals with how to evaluate generative models and propose various evaluation measures.
The most important ones are summarized in extensive survey papers \cite{Lucic2018NIPS,Xu2018ArXiv,Borji2019CVIU}.
They can be divided into two major categories: qualitative and quantitative measures.

Qualitative evaluation~\cite{Goodfellow2014NIPS,Huang2017CVPR,Zhang2017ICCV,Srivastava2017NIPS,Lin2018NIPS,Chen2016NIPS,Mathieu2016NIPS} uses visual inspection of a small collection of examples by humans and is therefore of subjective nature.
It is a simple way to get an initial impression of the performance of a generative model but cannot be performed in an automated fashion.
The subjective nature makes it difficult to compare performances across different works, even when a large inspection group, such as via Mechanical Turk, is used.
Furthermore, it is expensive and time-consuming.

Quantitative evaluation, on the other hand, is performed over a large collection of examples, often in an automated fashion.
The two most popular quantitative metrics are the \ac{IS}~\cite{Salimans2016NIPS} and the \ac{FID}~\cite{Heusel2017NIPS}.
In their original form, these and other ImageNet-based metrics~\cite{Gurumurthy2017CVPR,Che2017ICLR,Zhou2018ICLR} are exclusively applicable to camera image data, as they are based on features learned from the ImageNet dataset~\cite{ImageNet2009} and can therefore not be directly applied to LiDAR scans.

Therefore, \cite{Shu2019ICCV} proposes \acf{FPD}, which can measure the quality of GAN-generated 3D point clouds.
Based on the same principle as \ac{FID}, \ac{FPD} calculates the 2-Wasserstein distance between real and fake Gaussian measures in the feature spaces extracted by PointNet~\cite{Charles2017CVPR}.
In contrast to our method, \ac{FPD} requires labels on the target domain to train the feature extractor.
Further, it is only possible to compare a sample to one particular distribution and therefore makes it difficult to obtain a reliable measure on unseen data.

There also exists a number of modality independent GAN metrics, such as
distribution-based~\cite{Tolstikhin2017NIPS,Gretton2012JMLR,Arora2018ICLR,Richardson2018NIPS},
classification-based~\cite{Santurkar2018ICML,Lehmann2006,Yang2017ICLR},
and model comparison~\cite{Im2016ArXiv,Olsson2018ArXiv,Zhang2018ArXiv} approaches.
However, we do not further consider them in this work, since their focus is not on judging the quality of individual samples, but rather on capturing sample diversity and mode collapse in entire distributions.

\subsection{Metric Learning} 
\label{sec:related_metric_learning}

The goal of deep metric learning is to learn a feature embedding, such that similar data samples are projected close to each other while dissimilar data samples are projected far away from each other in the high-dimensional feature space.
Common methods use siamese networks trained with contrastive losses to distinguish between similar and dissimilar pairs of samples~\cite{Chicco2021}.
Thereupon, triplet loss architectures train multiple parallel networks with shared weights to achieve the feature embedding~\cite{Hoffer2015SIMBAD,Dong2018ECCV}.
This work uses an adversarial training technique to push features in a similar or dissimilar embedding.

\subsection{Contributions} 
\label{sec:related_contributions}

The aim of this paper is to provide a reliable metric that gives a quantitative estimate about the realism of generated LiDAR data.
The contributions of this work are threefold.
First and foremost, we present a novel way to learn a measure of realism in point clouds.
This is achieved by learning hierarchical point set features on a proxy classification task.
Second, we utilize an adversarial technique from the fairness-in-machine-learning domain in order to eliminate dataset-specific information.
This allows the measure to be used on unseen datasets.
Finally, we demonstrate how the fine-grained local realism score can be used for anomaly detection in LiDAR scans.


\section{Method}
\label{sec:method}

\subsection{Objective and Properties} 
\label{sec:method_properties}

The aim of this work is to provide a method to estimate the level of realism for arbitrary LiDAR point clouds.
Since there is no obvious definition of what realistic point clouds look like, we design the metric to learn relevant realism features from the distribution of real-world LiDAR data directly.
The output of the metric can then be interpreted as a distance measure between the input and the learned distribution in a high dimensional space.

Regarding the discussed aspects of existing point cloud quality measures and GAN measures, we expect a useful LiDAR point cloud metric to be:

\boldparagraph{Quantitative} 
The realism score is a quantitative measure that determines the distance of the input sample to the internal representation of the learned realistic distribution.
The score has well defined lower and upper bounds that reach from $0$ (unrealistic) to $1$ (realistic).

\boldparagraph{Universal} 
The metric has to be applicable to any LiDAR input and therefore must be independent from any application or task.
This means no explicit ground truth information is required.

\boldparagraph{Transferable} 
The metric must give a reliable and robust prediction for all inputs, independent on whether the data distribution of the input sample is known by the metric or not.
This makes the metric transferable to new and unseen data.

\boldparagraph{Local} 
The metric should be able to compute spatially local realism scores for smaller regions within a point cloud.
It is also expected to focus on identifying the quality of the point cloud properties while ignoring global scene properties as much as possible.

\boldparagraph{Flexible} 
Point clouds are usually sets of unordered points with varying size.
Therefore, it is crucial to have a processing that is permutation-invariant and independent of the number of points to process.

\boldparagraph{Fast} 
Speed is not the most important property, but a fast computation time allows the metric to run in parallel to the training of a neural network for LiDAR data generation.
This enables monitoring of generated sample quality over the training time of the network.

\vspace{0.15cm}
We implement our metric in such a way that the described properties are fulfilled.
To differentiate the metric from a GAN discriminator, we want to stress that a discriminator is not \textit{transferable} to unseen data.

\subsection{Architecture} 
\label{sec:method_architecture}

\begin{figure*}
	\centering
	\includegraphics[width=0.96\linewidth]{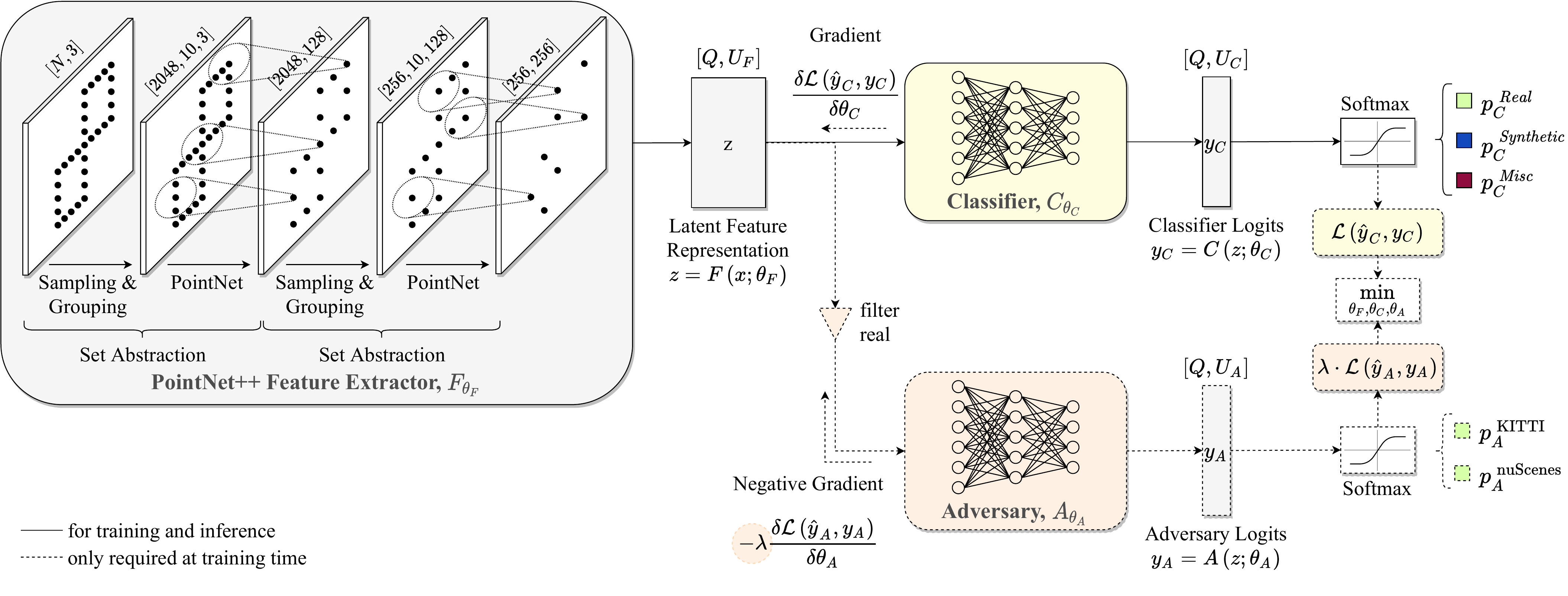}
	\caption{
		\textbf{Architecture}:
		The feature extractor $F_{\theta_F}$ uses hierarchical feature set learning from PointNet++~\cite{Qi2017NIPS} to encode information about each of the $Q$ query points and their nearest neighbors.
		The neighborhood features $z$ are then passed to two identical networks, the classifier $C_{\theta_C}$ and the adversary $A_{\theta_A}$.
		The classifier outputs probability scores $\bp_C$ for each category -- \Real{}, \Syn{}, \Misc{} -- while the adversary outputs probability scores $\bp_A$ for each dataset.
		For both network outputs a multi-class cross-entropy loss is minimized.
		To let the classifier perform as good as possible while the adversary should perform as bad as possible, the gradient is inverted between the adversary input and the feature extractor~\cite{Beutel2017FAT,Raff2018DSAA}.
		The input to the adversary is limited to samples from the \Real{} category.
		$\lambda$ is a factor that regulates the influence of the two losses, weighting the ratio of accuracy versus fairness.
		In our experiments we use a factor of $\lambda=0.3$.
		For more information we refer to the supplementary material.
	}
	\label{fig:architecture}
\end{figure*}

\figref{fig:architecture}~shows the architecture of our approach.
The following describes the components and presents how each part is designed to contribute towards achieving the desired metric properties.
The underlying idea of the metric design is to compute a distance measure between different data distributions of \textit{realistic} and \textit{unrealistic} LiDAR point cloud compositions.
The network learns features indicating realism from data distributions by using a proxy classification task.
Specifically, the network is trained to classify point clouds from different datasets into three categories: \Real{}, \Syn{}, \Misc{}.
The premise is the possibility to divide the probability space of LiDAR point clouds into those that derive from real-world data (\Real{}), those that derive from simulations (\Syn{}), and all the others (\Misc{}), \eg distorted or randomized data.
Refer to \figref{fig:eyecatcher} for an impression.
By acquiring the prior information about the tripartite data distribution, the metric does not require any target information or labels for inference.

The features are obtained with hierarchical feature set learning, explained in~\secref{sec:method_feature_extractor}.
\secref{sec:method_fairness}~outlines our adversarial learning technique.

\subsubsection{Feature Extractor}
\label{sec:method_feature_extractor}

The left block in \figref{fig:architecture} visualizes the PointNet++~\cite{Qi2017NIPS} concept of the feature extractor $F_{\theta_F}$.
We use two abstraction levels, sampling 2048 and 256 query points, respectively, with 10 nearest neighbors each.
Keeping the number of neighbors and abstraction levels low limits the network to only encode information about \textit{local} LiDAR-specific statistics instead of global scenery information.
On the other hand, the high amount of query points helps to cover many different regions within the point cloud and guarantees the \textit{local} aspect of our method.
We use a 3-layer \ac{MLP} in each abstraction level with filter sizes of $\left[64, 64, 128\right]$ and $\left[128, 128, 256\right]$, respectively.
This results in the neighborhood features, a latent space representation $z=F(x,\theta_F)$ of size $\left[Q, U_F\right]$ with $U_F=256$ features for each of the $Q=256$ query points.
The features $z$ are then fed to a densely connected classifier $C_{\theta_C}$ (yellow block).
It consists of a hidden layer with 128 units and 50\% dropout, and the output layer with $U_C$ units.

As the classifier output, we obtain the probability vector $\bp_{C,q} =\operatorname{softmax}(y_C) \in [0,1]^{U_C}$ per query point~$q$.
The vector has $U_C=3$ entries for each of the categories \Real{}, \Syn{} and \Misc{}.
The component $p_{C,q}^\Real{}$ quantifies the degree of realism in each local region~$q$.
The scores $\bS= \frac{1}{Q} \sum_q \bp_{C,q}$ for the entire scene are given by the mean over all query positions.
Here, $S^\Real{}$ is a measure for the degree of realism of the entire point cloud.
A score of $0$ indicates low realism while $1$ indicates high realism.

\subsubsection{Fair Categorization}
\label{sec:method_fairness}

To obtain a \textit{transferable} metric network, we decided to leverage a concept often used to design fair network architectures~\cite{Beutel2017FAT,Raff2018DSAA}.
The idea is to force the feature extractor to encode only information into the latent representation $z$ that is relevant for realism estimation.
This means, we actively discourage the feature extractor from encoding information that is specific to the distribution of a single dataset.
In other words, using fair networks terminology~\cite{Beutel2017FAT}, we treat the concrete dataset name as a sensitive attribute.
With this procedure we improve the generalization ability towards unknown data.

To achieve this behavior, we add a second output head, the adversary $A_{\theta_A}$, which is only required at training time (see orange block in~\figref{fig:architecture}).
Its layers are identical to the one of the classifier, except for the number of units in the output layer.
Following the designs proposed in~\cite{Beutel2017FAT,Raff2018DSAA}, we train all network components by minimizing the losses for both heads, $\mathcal{L}_C=\mathcal{L}\left(y_C,\hat{y}_C\right)$ and $\mathcal{L}_A=\mathcal{L}\left(y_A,\hat{y}_A\right)$, but reversing the gradient in the path between the adversary input and the feature extractor.
The goal is for $C$ to predict the category $y_C$ and for $A$ to predict the dataset $y_A$ as good as possible, but for $F$ to make it hard for $A$ to predict $y_A$.
Training with the reversed gradient results in $F$ encoding as little information as possible for predicting $y_A$.
The training objective is formulated as
\begin{equation}
	\begin{aligned}
		\min_{\theta_F,\theta_C,\theta_A}
		  & \mathcal{L} \Big( C\big( F(x;\theta_F);\theta_C \big), \hat{y}_C \Big) \\
        & \quad + \mathcal{L} \Big( A\big( J_\lambda[ F(x;\theta_F) ];\theta_A \big), \hat{y}_A \Big)
	\end{aligned}
\end{equation}
with $\theta$ being the trainable variables and $J_\lambda$ a special function such that
\begin{equation}
    J_\lambda[F] = F \quad \text{but} \quad \nabla J_\lambda[F] = -\lambda \cdot \nabla F \quad .
\end{equation}
The factor $\lambda$ determines the ratio of accuracy and fairness.

In contrast to the application examples from related literature~\cite{Beutel2017FAT,Raff2018DSAA}, our requested attribute, the category, can always be retrieved from the sensitive attribute, the specific dataset.
Therefore, we consider only the data samples coming from the \Real{} category in the adversary during training, indicated by the triangle in~\figref{fig:architecture}.
Considering samples from all categories leads to unwanted decline in classifier performance (see ablation study in \secref{sec:experiments_fairness}).
Our filter extension forces the feature extractor to encode only common features within the \Real{} category, but keep all other features in the other categories.
In principal, it is possible to use a separate adversary for each category, but this enlarges the training setup unnecessarily since the main focus is to determine the realism of the input.


\section{Experiments}
\label{sec:experiments}

\secref{sec:experiments_datasets} and \secref{sec:experiments_baseline} present the datasets and baselines used in our experiments.
We provide extensive evaluations to demonstrate the capability and applicability of our method.
First, we present the classification performance on completely unseen data distributions (\secref{sec:experiments_classification}).
Then, we show that the metric provides a credible measure of realism when applied to generated point clouds (\secref{sec:experiments_reconstruction}).
In~\secref{sec:experiments_fairness} we show the feature embedding capabilities of the fair training strategy in an ablation study.
\secref{sec:experiments_transition}~shows that our metric is capable to interpolate the space between the training categories in a meaningful way.
As an important application, \secref{sec:experiments_anomaly}~demonstrates how our method can be used for anomaly detection in point clouds.
In the end, we elaborate on the limitations of our method (\secref{sec:experiments_limitations}).
Additional visualizations and implementation details are provided in the supplementary material.

\subsection{Datasets} 
\label{sec:experiments_datasets}

\begin{table}[b]
	\centering
	\caption{
		\textbf{Training Datasets}:
		The table lists the datasets that are used as support sets to train the metric network.
		The rightmost column shows the number of samples in the training split.
	}
	\label{tab:datasets_training}

	\begin{tabular}{llr}
		\toprule
		Category				& Support Set						& Train Split \\
		\midrule
		\multirow{2}{*}{\Real{}}	& KITTI~\cite{Geiger2013IJRR}		&      18,329 \\
								& nuScenes~\cite{Caesar2020CVPR}	&      28,130 \\
		\hline
		\multirow{2}{*}{\Syn{}}	& CARLA~\cite{Dosovitskiy2017}		&     106,503 \\
								& GeometricSet						&      18,200 \\
		\hline
		\Misc{}					& Misc 1, 2, 3					&    $\infty$ \\
		\bottomrule
	\end{tabular}

\end{table}

\begin{table}[b]
	\centering
	\caption{
		\textbf{Evaluation Datasets}:
		The table lists the datasets that are used to evaluate the performance of the metric.
		Second column from the right shows the number of samples used to train the up-sampling networks.
		At the bottom, one example per dataset is depicted.
	}
	\label{tab:datasets_evaluation}

	\begin{tabular}{llrc}
		\toprule
		Category	& Dataset						& Train Split	& Example \\
		\midrule
		\Real{}		& PandaSet~\cite{PandaSet2020}	& 7,440			& (a) \\
		\hline
		\Misc{}		& Misc~4						& -				& (b) \\
		\bottomrule
	\end{tabular}

	\vspace{4pt}

	\begin{tabular}{l@{\hskip2pt}c l@{\hskip2pt}c}
		(a) & \includegraphics[width=0.35\linewidth]{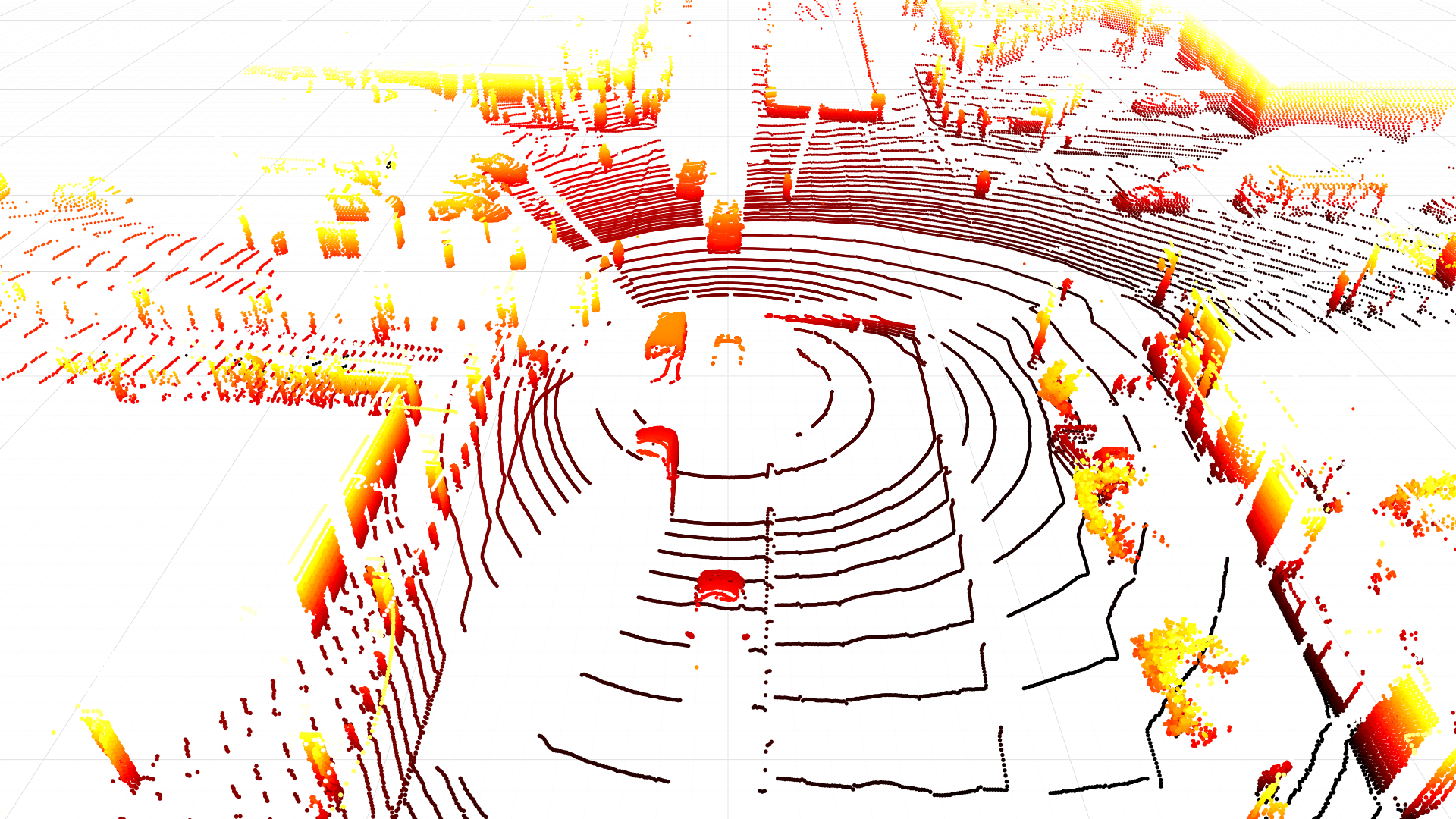} &
		(b) & \includegraphics[width=0.35\linewidth]{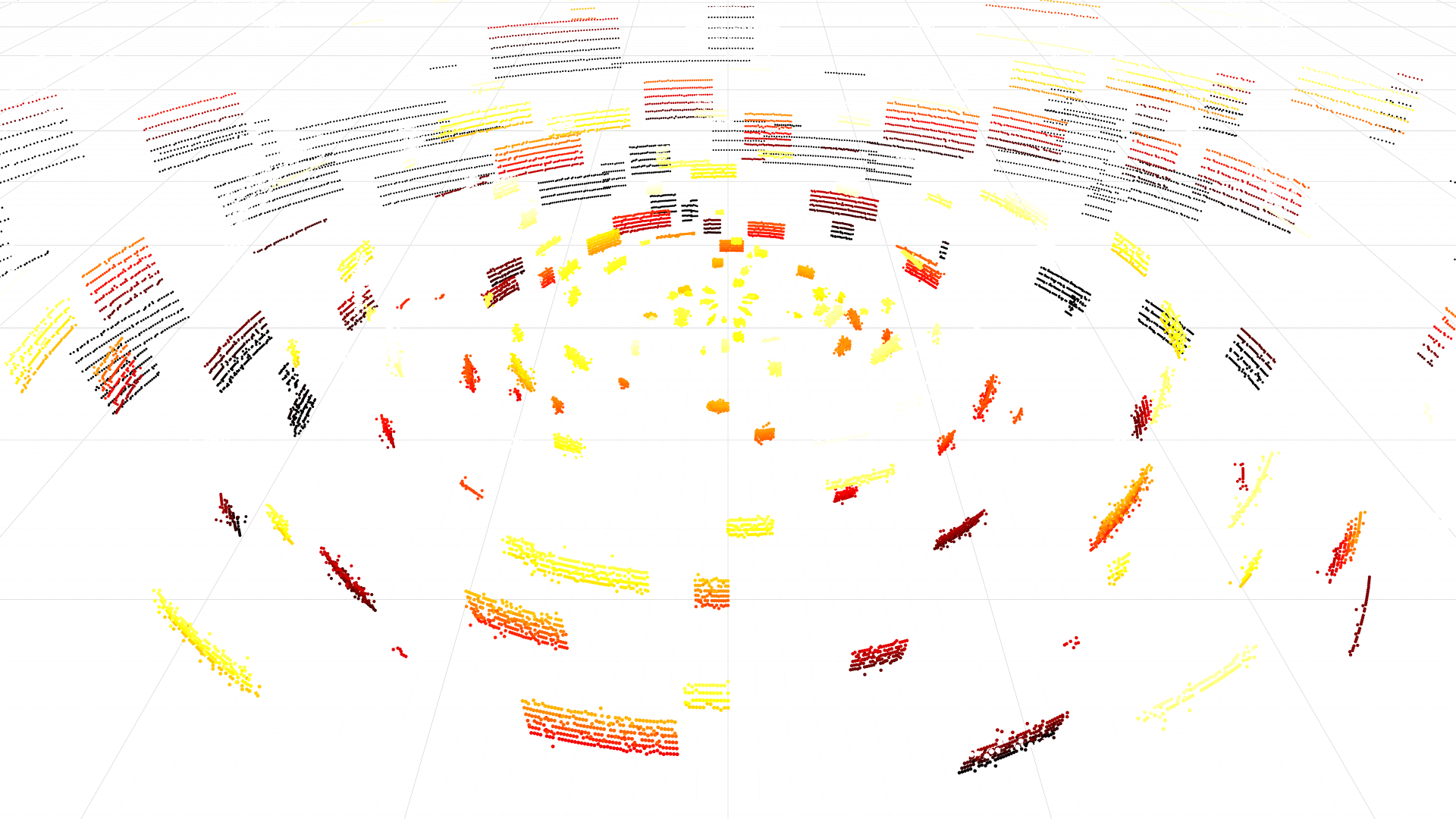} \\
	\end{tabular}

\end{table}

\begin{figure*}
    \centering

    \begin{tabularx}{\linewidth}{c@{\hskip1pt}c@{\hskip3pt} c@{\hskip3pt} c@{\hskip1pt}c@{\hskip3pt} c@{\hskip3pt} c@{\hskip1pt}c@{\hskip1pt}c}
        KITTI & nuScenes && CARLA & GeometricSet && Misc~1 & Misc~2 & Misc~3 \\

        \includegraphics[width=0.135\linewidth]{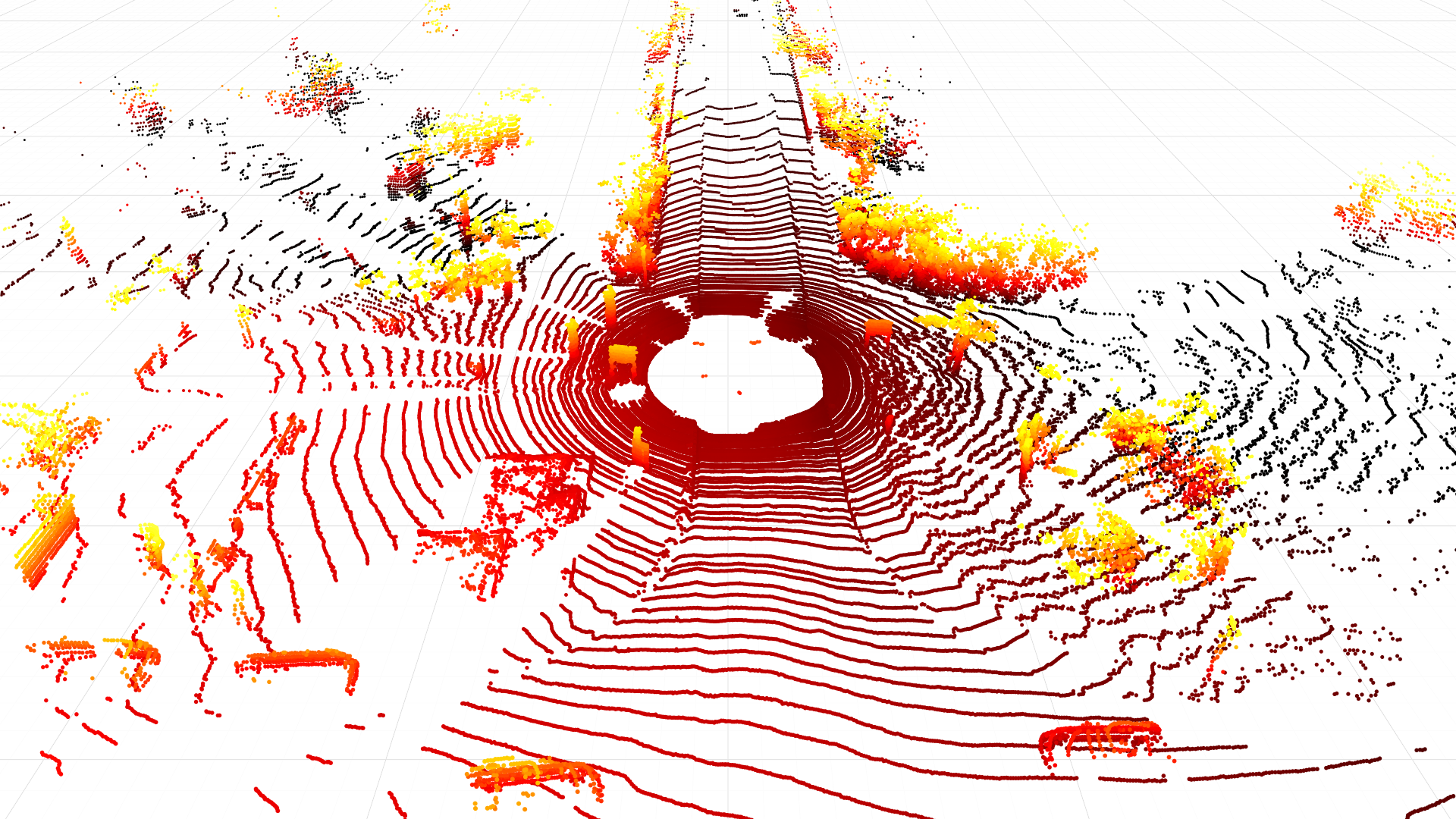} & \includegraphics[width=0.135\linewidth]{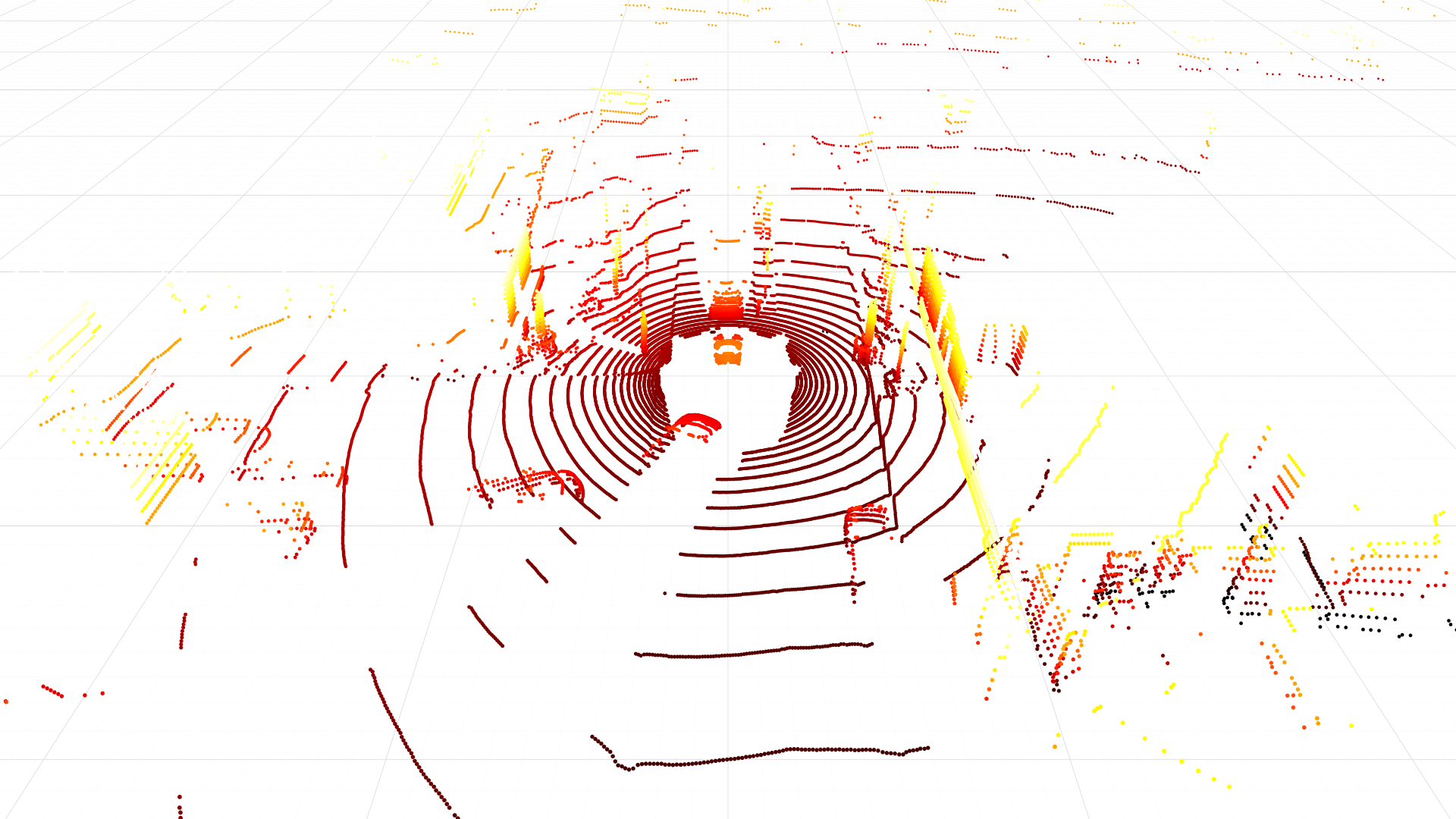} && \includegraphics[width=0.135\linewidth]{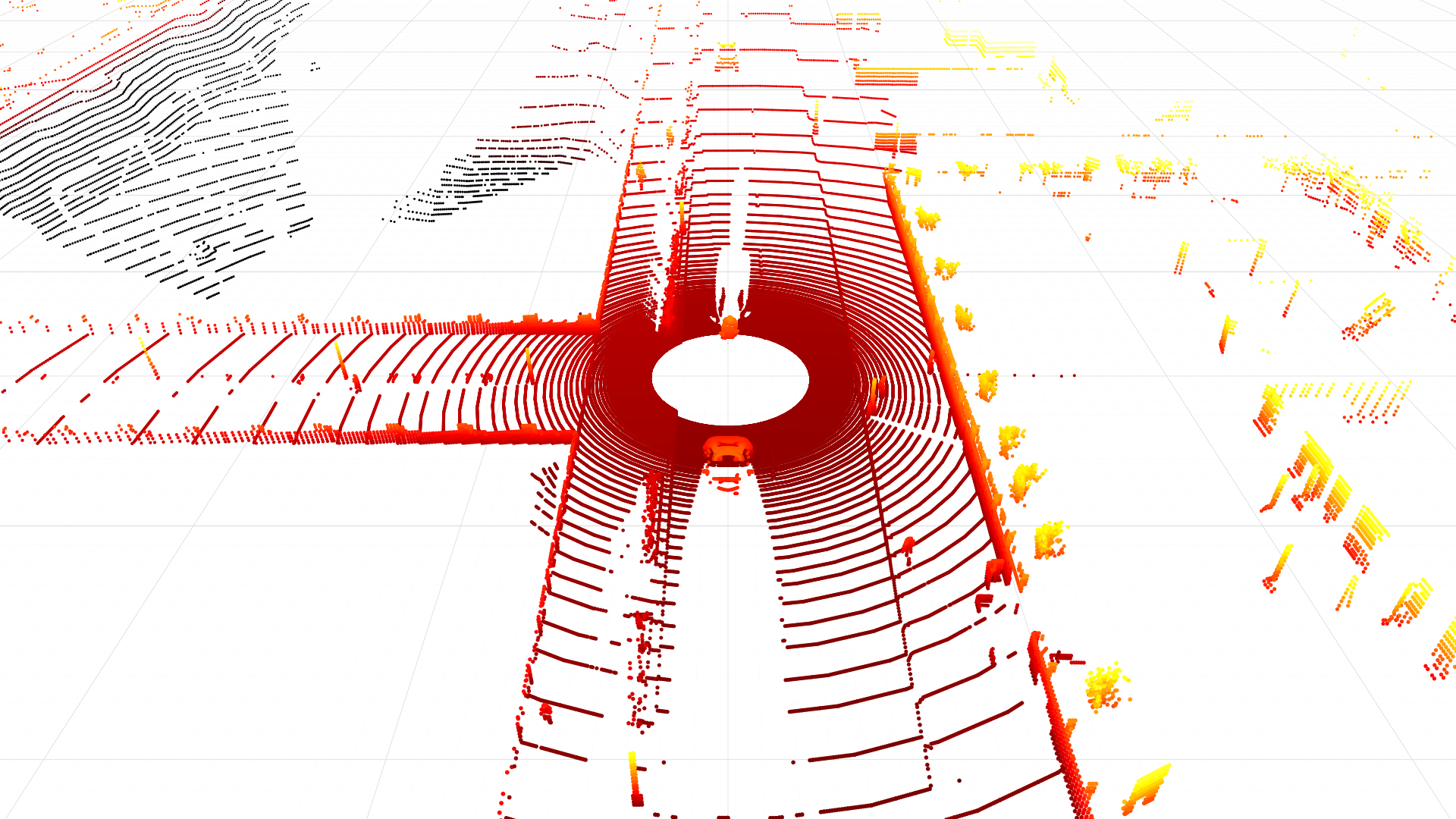} & \includegraphics[width=0.135\linewidth]{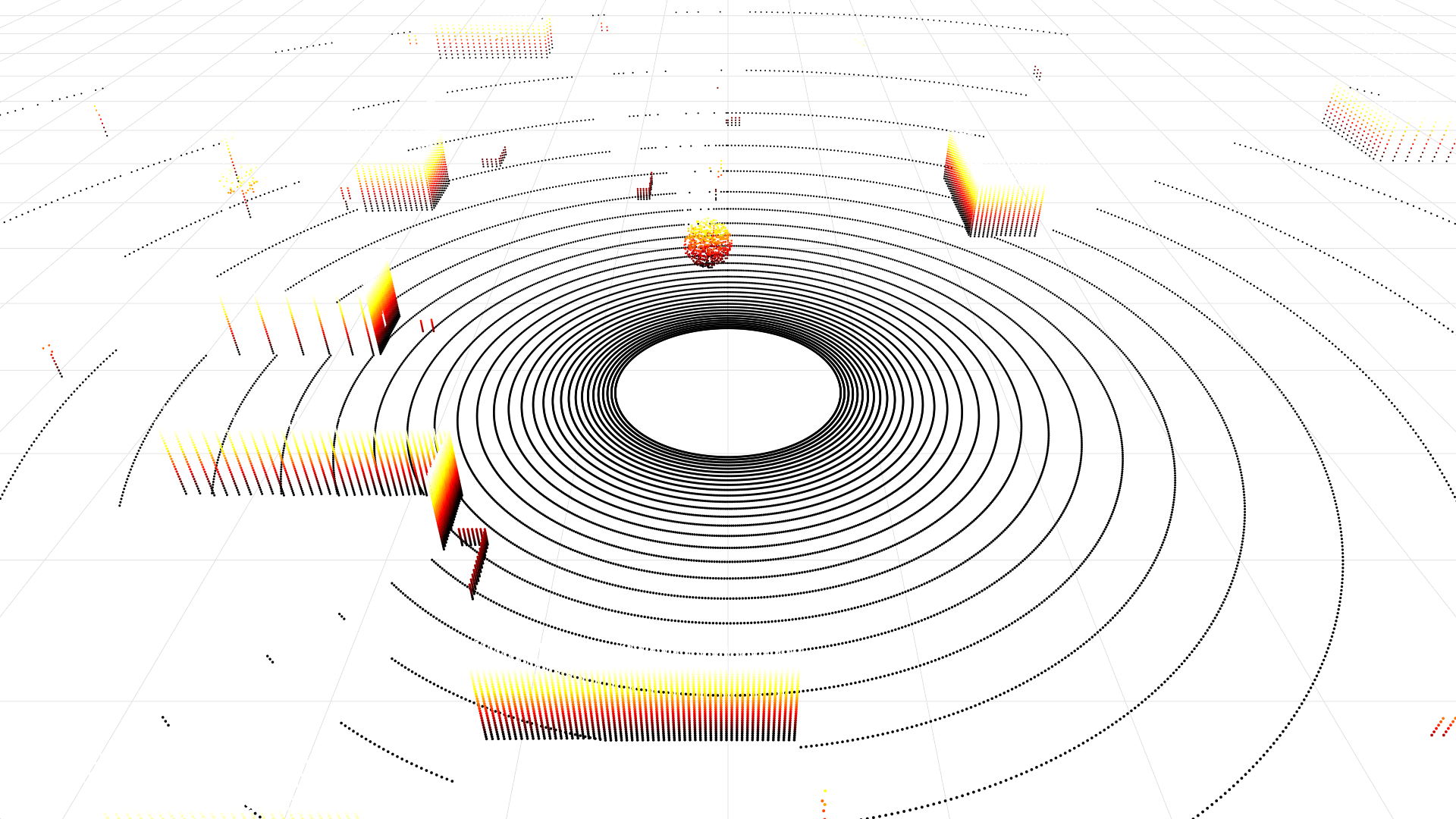} && \includegraphics[width=0.135\linewidth]{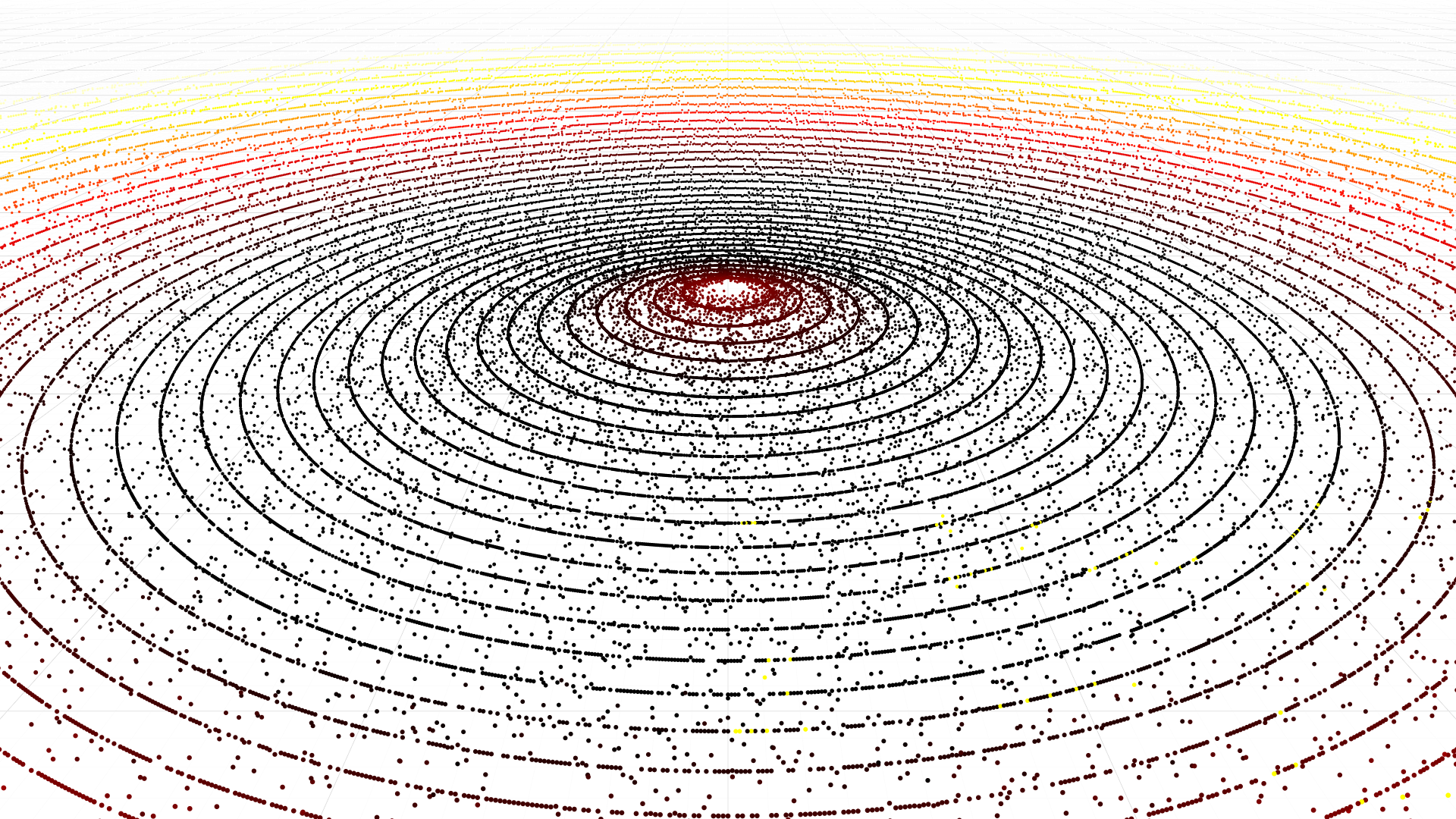} & \includegraphics[width=0.135\linewidth]{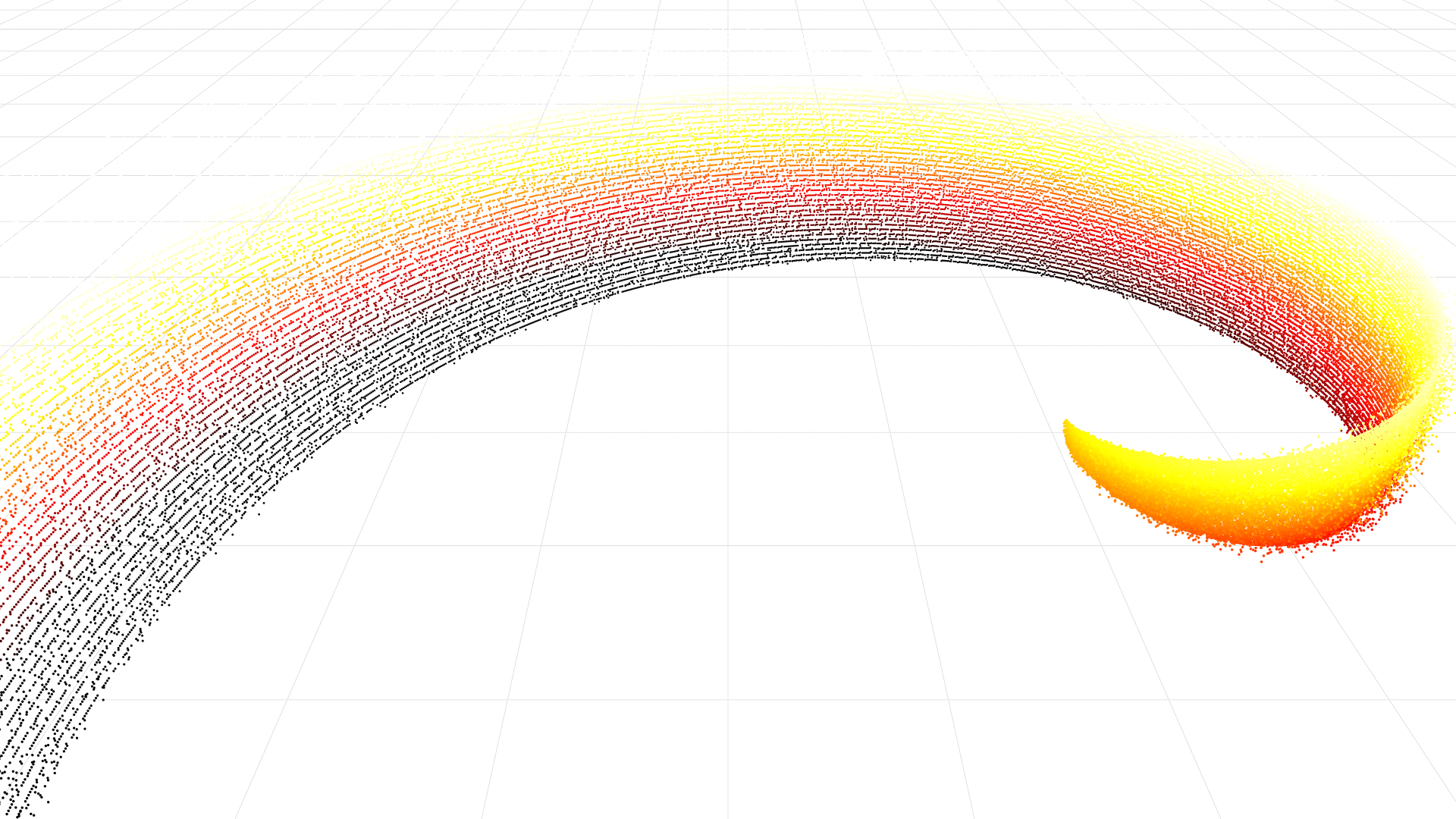} &
        \includegraphics[width=0.135\linewidth]{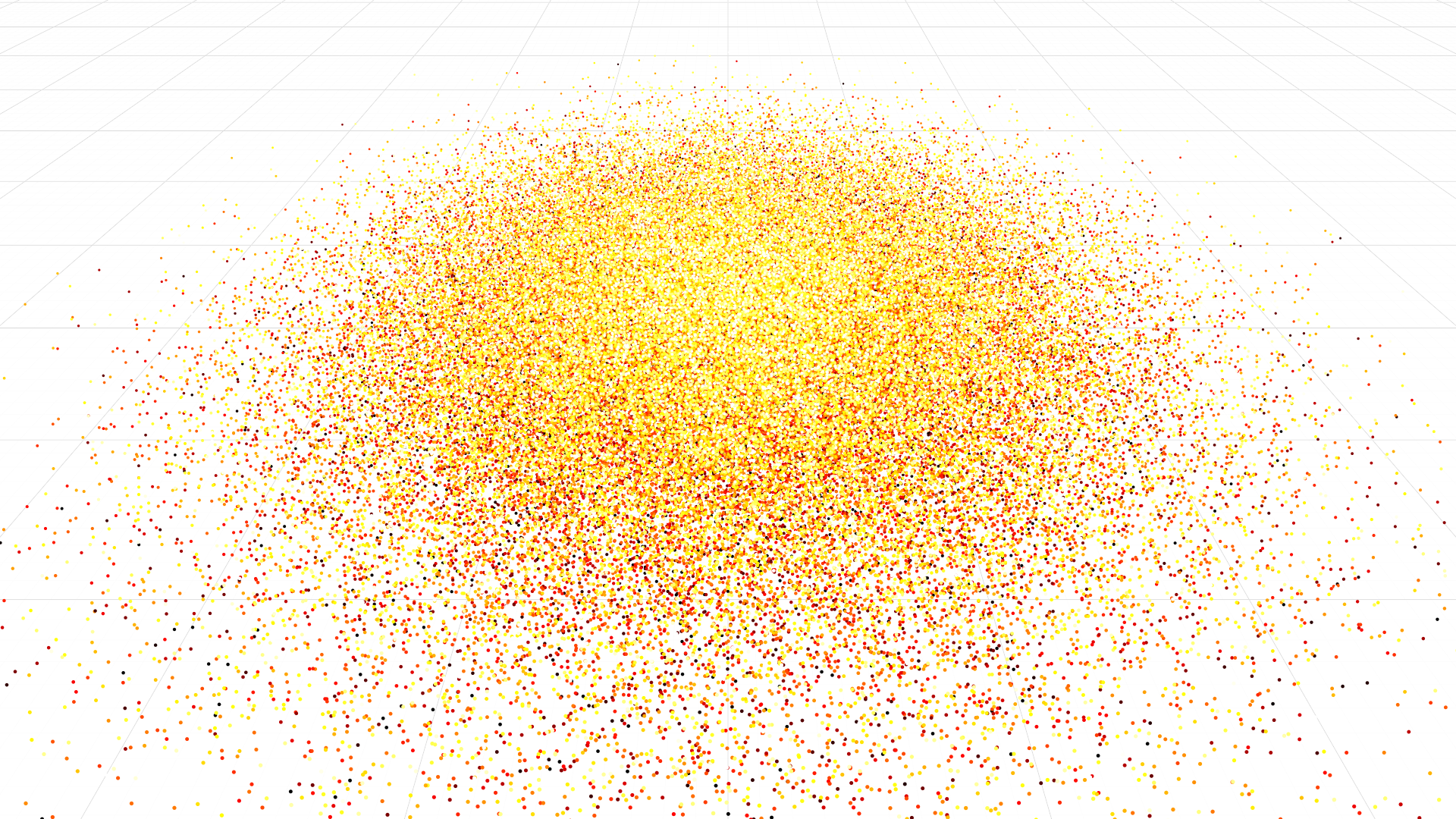} \\

        \cline{1-2}\cline{4-5}\cline{7-9}
        \multicolumn{2}{c}{\Real{}} && \multicolumn{2}{c}{\Syn{}} && \multicolumn{3}{c}{\Misc{}}
    \end{tabularx}

    \caption{
        \textbf{Support Sets Showcase}:
        The height is color coded from black/red small to yellow/white large.
    }
    \label{fig:examples_support_center}
\end{figure*}

We use two different groups of datasets.
\tabref{tab:datasets_training}~lists the datasets that are used as support sets for training the metric network.
Visualizations for each dataset are provided in~\figref{fig:examples_support_center}.
\tabref{tab:datasets_evaluation}~contains the datasets that are exclusively used to evaluate the metric, where PandaSet is used to train the super-resolution networks that are needed for some of the following experiments.
With the strict separation of training and evaluation datasets, additionally to the mandatory training and test splits, we can demonstrate that our method is a useful measure on unknown data distributions.
In both cases alike, the datasets are divided into three categories: \Real{}, \Syn{}, \Misc{}.

Publicly available real-world datasets are used in \Real{} for training (KITTI, nuScenes), and evaluation (PandaSet).

For \Syn{}, we use the CARLA simulator where we implement the sensor specifications of a Velodyne HDL-64 sensor to create ray-traced range measurements.
GeometricSet is the second support set in this category.
Here, simple geometric objects, such as spheres and cubes are randomly scattered on a ground plane in three dimensional space and ray-traced in a scan pattern.

Finally, we add a third category, \Misc{}, to allow the network to represent absurd data distributions, as they often occur during GAN~trainings or with data augmentation.
Therefore, \Misc{} contains randomized data that is generated at training or evaluation time.
Misc~1 and Misc~2 are generated by linearly increasing the depth over the rows or columns of a virtual LiDAR scanner, respectively.
Misc~3 is a simple Gaussian noise with varying standard deviations.
Misc~4 is created by setting patches of varying height and width of the LiDAR depth projection to the same distance.
Misc~4 is only used for evaluation.
Varying degrees of Gaussian noise are added to the Euclidean distances of Misc~\{1,2,4\}.

It is certainly possible to add more support sets to each category, for example by including more datasets or by augmenting existing ones.
However, we want to stress that our method is a reliable metric even when using just a small number of support sets, as demonstrated by the experiments below.
In addition to the training data listed in the tables, we use 600 samples from a different split of each dataset to obtain our evaluation results.
No annotations or additional information are required to train or apply the metric, all operations are based on the $xyz$ coordinates of the point clouds.

\subsection{Baselines and Evaluation} 
\label{sec:experiments_baseline}

We evaluate our metric by applying it to the generated outputs of different versions of LiDAR up-sampling networks.
The task of up-sampling is easy to understand and offers the possibility to compare against other distance metrics by exploiting the high-resolution Ground Truth.
Furthermore, it renders the difficulty for the networks to synthesize realistic high-resolution LiDAR outputs and is therefore a suitable testing candidate for our realism metric.

As a baseline, we compute the \acf{CD} between the high-resolution prediction and the Ground Truth and compare against our metric results and human visual judgment.
Since we follow the procedure proposed in~\cite{Triess2019IV}, where the data is represented by cylindrical depth projections, we can compute additional distance measures.
Specifically, we compute the \ac{MSE} and \ac{MAE} from the dense point cloud representation between the reconstruction and the target.

\subsection{Metric Results} 
\label{sec:experiments_classification}

\begin{figure}[b]
    \centering

    \begin{tikzpicture}
    \pgfplotsset{
        width=7.4cm,
        compat=newest,
        grid style={dashed,gray!30},
        select coords between index/.style 2 args={
            x filter/.code={
                \ifnum\coordindex<#1\def\pgfmathresult{}\fi\ifnum\coordindex>#2\def\pgfmathresult{}\fi
            }
        }
    }

    \begin{axis}[
        name=plottop,
        height=2.0cm,
        ytick={0,1},
        yticklabels={
          {\color{misc_plot}Misc~4},
          {\color{real_plot}PandaSet}
        },
        yticklabel pos=right,
        xticklabels=\empty,
        xmin=0,
        xmax=1,
        grid=both,
        legend style={at={(0.5,1.2)},anchor=south},
        legend columns=-1,
        legend cell align={left},
    ]

        \addplot[only marks,color=real_plot,mark=*,error bars/.cd,x dir=both,x explicit]
        table[y expr=\coordindex,x=real_mean,x error=real_stddev,col sep=comma,select coords between index={0}{1}]
        {data/metric_training_results.csv};

        \addplot[only marks,color=syn_plot,mark=triangle*,error bars/.cd,x dir=both,x explicit]
        table[y expr=\coordindex,x=syn_mean,x error=syn_stddev,col sep=comma,select coords between index={0}{1}]
        {data/metric_training_results.csv};

        \addplot[only marks,color=misc_plot,mark=square*,error bars/.cd,x dir=both,x explicit]
        table[y expr=\coordindex,x=misc_mean,x error=misc_stddev,col sep=comma,select coords between index={0}{1}]
        {data/metric_training_results.csv};

        \legend{$S^\Real{}$,$S^\Syn{}$,$S^\Misc{}$}
    \end{axis}

    \begin{axis}[
        name=plotbottom,
        at={($(plottop.south)-(0,0.4cm)$)},
        anchor=north,
        height=4cm,
        xlabel=Mean Softmax Score $S$,
        ytick={2,3,4,5,6,7,8},
        yticklabels={
          {\color{misc_plot}Misc~3},
          {\color{misc_plot}Misc~2},
          {\color{misc_plot}Misc~1},
          {\color{syn_plot}GeometricSet},
          {\color{syn_plot}CARLA},
          {\color{real_plot}nuScenes},
          {\color{real_plot}KITTI}
        },
        yticklabel pos=right,
        xmin=0,
        xmax=1,
        grid=both,
    ]

        \addplot[only marks,color=real_plot,mark=*,error bars/.cd,x dir=both,x explicit]
        table[y expr=\coordindex,x=real_mean,x error=real_stddev,col sep=comma,select coords between index={2}{9}]
        {data/metric_training_results.csv};

        \addplot[only marks,color=syn_plot,mark=triangle*,error bars/.cd,x dir=both,x explicit]
        table[y expr=\coordindex,x=syn_mean,x error=syn_stddev,col sep=comma,select coords between index={2}{9}]
        {data/metric_training_results.csv};

        \addplot[only marks,color=misc_plot,mark=square*,error bars/.cd,x dir=both,x explicit]
        table[y expr=\coordindex,x=misc_mean,x error=misc_stddev,col sep=comma,select coords between index={2}{9}]
        {data/metric_training_results.csv};

    \end{axis}
\end{tikzpicture}

    \caption{
        \textbf{Metric Results}:
        Shown the metric output~$S$ (mark) and the standard deviation (whisker) for all three categories on different datasets.
        The upper part shows the unseen datasets, PandaSet and Misc~4.
        They are both correctly classified as \Real{} and \Misc{}, respectively.
        The results of the support sets are depicted in the lower part.
        The text coloring indicates the correct class.
    }
    \label{fig:experiments_metric_training_results}
\end{figure}
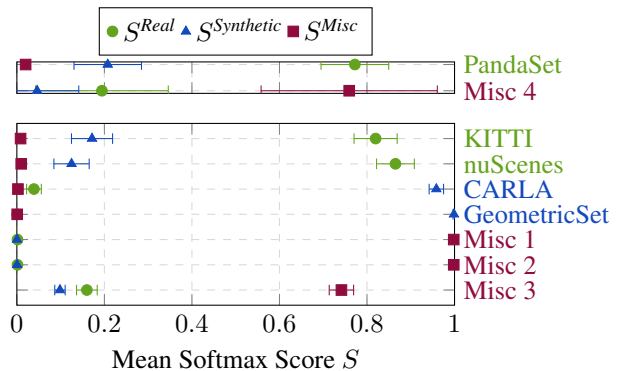

We run our metric network on the test split of the evaluation datasets and, as a reference, on the support sets as well.
\figref{fig:experiments_metric_training_results}~shows the mean of the metric scores $S$ for each of the three categories.
The observation is that our method correctly identifies the unknown PandaSet data as being realistic by outputting a high \Real{} score.
Misc~4 is also correctly classified, as indicated by the high \Misc{} scores.

\subsection{Metric Verification} 
\label{sec:experiments_reconstruction}

In this section we demonstrate how our quality measure reacts to data that was generated by a neural network and compare it to traditional evaluation measures, as introduced in~\secref{sec:experiments_baseline}.

We compare bilinear interpolation to two \ac{CNN} versions and one \ac{GAN}.
The \acp{CNN} are trained with $\mathcal{L}_1$ and $\mathcal{L}_2$ loss, respectively.
For the \ac{GAN} experiments, we adapt SRGAN~\cite{Ledig2017CVPR} from which the generator is also used in the \ac{CNN} trainings.
We conduct the experiments for $2\!\times$, $4\!\times$, and $8\!\times$ up-sampling and compare the predictions of our metric to the qualitative \ac{MOS} ranking in~\cite{Triess2019IV}.
Here, we only present the results for $4\!\times$ up-sampling, the other results, implementation details, and qualitative results are given in the appendix.

\figref{fig:experiments_upsampling_metric_4x}~shows the realism score on the right side (higher is better) and the baseline metrics on the left (smaller is better).
The realism score for the original data (GT), is displayed for reference and achieves the best results with the highest realism score and lowest reconstruction error.
The methods are ranked from top to bottom by increasing realism as perceived by human experts which matches the ordering by \acl{MOS}~\cite{Triess2019IV}.
In general, the baseline metrics do not show clear correlation to the degree of realism and fail to produce a reasonable ordering of the methods.
Our realism score, on the other hand, sorts the up-sampling methods according to human visual judgment.
These results align with the ones in~\cite{Triess2019IV}, where the \ac{MOS} study showed that a low reconstruction error does not necessarily imply a good generation quality.

The GAN obtains the lowest realism score which aligns with visual inspections of the generated data.
An example scene is visualized in the bottom right of~\figref{fig:eyecatcher} and further images are provided in the appendix.
We attribute the high levels of noise and the low generation quality to two facts:
First, due to the limited size of PandaSet, the \ac{GAN} could only be trained with $\sim\!7,000$ samples, which is very small for such large data complexities.
Second, the SRGAN architecture is not specifically optimized for LiDAR point clouds.

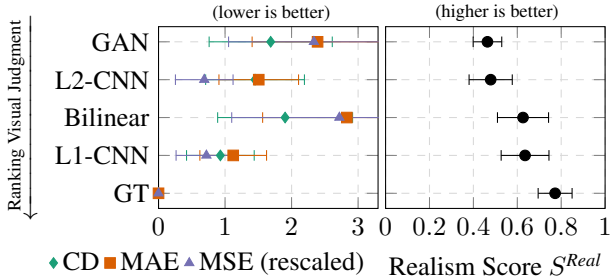
\begin{figure}
    \centering

    \pgfplotstableread[col sep=comma]{data/upsampling_metric_results_4x.csv}\datatable

\begin{tikzpicture}
    \pgfplotsset{
        compat=newest,
        grid style={dashed,gray!30},
    }

    \node[] at (1.5,2.6) {\scriptsize (lower is better)\par};

    \begin{axis}[
        name=plot1,
        height=4cm,
        width=4.5cm,
        ytick=data,
        yticklabels from table={\datatable}{Method},
        ylabel=$\xleftarrow{\text{Ranking Visual Judgment}}$,
        xmin=0,
        xmax=3.3,
        grid=both,
        legend style={at={(1.0,-0.3)},anchor=east,draw=none},
        legend columns=-1,
        legend cell align={left},
    ]

        \addplot[only marks,color=cd_plot,mark=diamond*,error bars/.cd,x dir=both,x explicit]
        table[y expr=\coordindex,x=CD_mean,x error=CD_std,col sep=comma] {data/upsampling_metric_results_4x.csv};

        \addplot[only marks,color=mae_plot,mark=square*,error bars/.cd,x dir=both,x explicit]
        table[y expr=\coordindex,x=MAE_mean,x error=MAE_std,col sep=comma] {data/upsampling_metric_results_4x.csv};

        \addplot[only marks,color=mse_plot,mark=triangle*,error bars/.cd,x dir=both,x explicit]
        table[y expr=\coordindex,x expr=\thisrow{MSE_mean}/50,x error expr=\thisrow{MSE_std}/50,col sep=comma] {data/upsampling_metric_results_4x.csv};

        \legend{CD,MAE,MSE (rescaled)}
    \end{axis}

    \node[] at (4.5,2.6) {\scriptsize (higher is better)\par};

    \begin{axis}[
        name=plot2,at={($(plot1.east)+(0.1cm,0)$)},
        anchor=west,
        height=4cm,
        width=4.5cm,
        xlabel=Realism Score $S^\Real{}$,
        ytick=data,
        yticklabels=\empty,
        xmin=0,
        xmax=1,
        grid=both,
    ]

        \addplot[black,mark=*,only marks,error bars/.cd,x dir=both,x explicit]
        table[y expr=\coordindex,x=Real_mean,x error=Real_std,col sep=comma] {data/upsampling_metric_results_4x.csv};

    \end{axis}
\end{tikzpicture}

    \caption{
        \textbf{Metric Scores for Up-Sampling Methods}:
        The diagram lists four methods to perform $4\times$ LiDAR scan upsampling, in addition to the high-resolution \textit{Ground Truth} (GT) data.
        The left side shows different baseline measures, \ie reconstruction errors (lower is better), whereas the right side shows our realism score (higher is better).
        Methods are ordered by their respective human visual judgment ratings.
    }
    \label{fig:experiments_upsampling_metric_4x}
\end{figure}

\subsection{Ablation Study} 
\label{sec:experiments_fairness}


The proposed approach uses the fairness setup with a filter extension to embed features related to realism and omit dataset-specific information.
To demonstrate the advantage of the adversarial feature learning and the necessity of our extension, we train two additional metric networks, one without the adversary and one with the adversary but without the extension.

\begin{figure}
	\centering

	\begin{tabular}{llll}
		\textcolor{kitti}{\rule{2.4mm}{2.4mm}} KITTI & \textcolor{nuscenes}{\rule{2.4mm}{2.4mm}} nuScenes & \textcolor{carla}{\rule{2.4mm}{2.4mm}} Carla & \textcolor{builder}{\rule{2.4mm}{2.4mm}} GeometricSet
	\end{tabular}
	\begin{tabular}{lll}
		\textcolor{misc1}{\rule{2.4mm}{2.4mm}} Misc 1 & \textcolor{misc2}{\rule{2.4mm}{2.4mm}} Misc 2 & \textcolor{misc3}{\rule{2.4mm}{2.4mm}} Misc 3
	\end{tabular}

	\begin{subfigure}{0.32\linewidth}
		\centering
		\includegraphics[width=\linewidth]{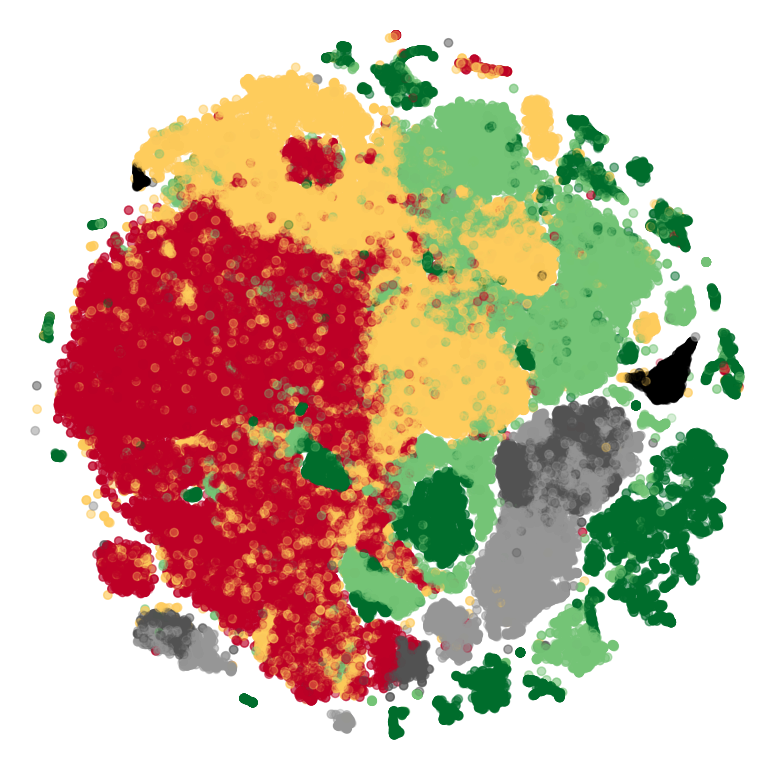}
		\caption{\label{fig:experiments_feature_embedding_none}No Adversary}
	\end{subfigure}%
	\hspace{4pt}%
	\begin{subfigure}{0.32\linewidth}
		\centering
		\includegraphics[width=\linewidth]{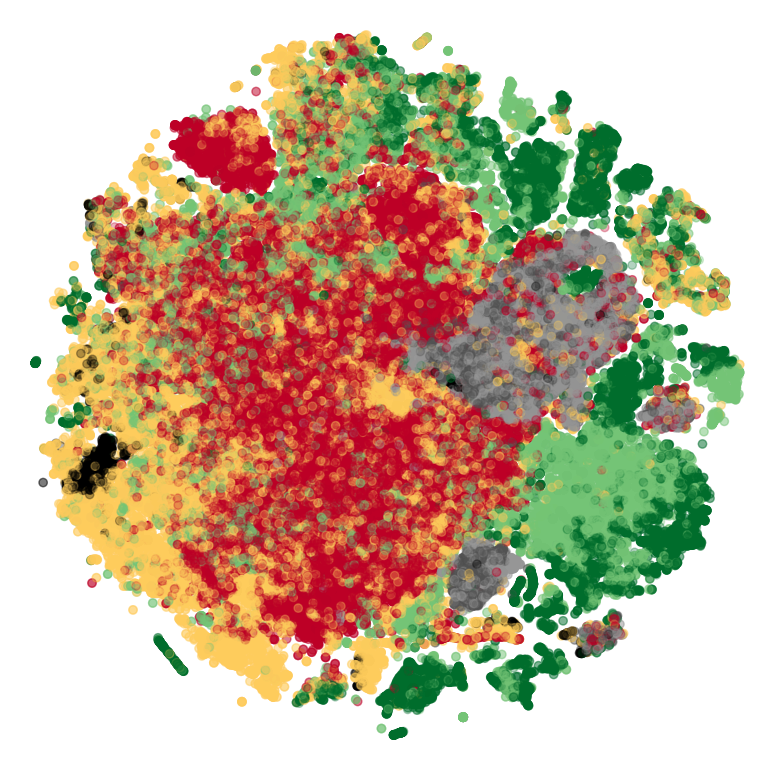}
		\caption{\label{fig:experiments_feature_embedding_full}Full Adversary}
	\end{subfigure}%
	\hspace{4pt}%
	\begin{subfigure}{0.32\linewidth}
		\centering
		\includegraphics[width=\linewidth]{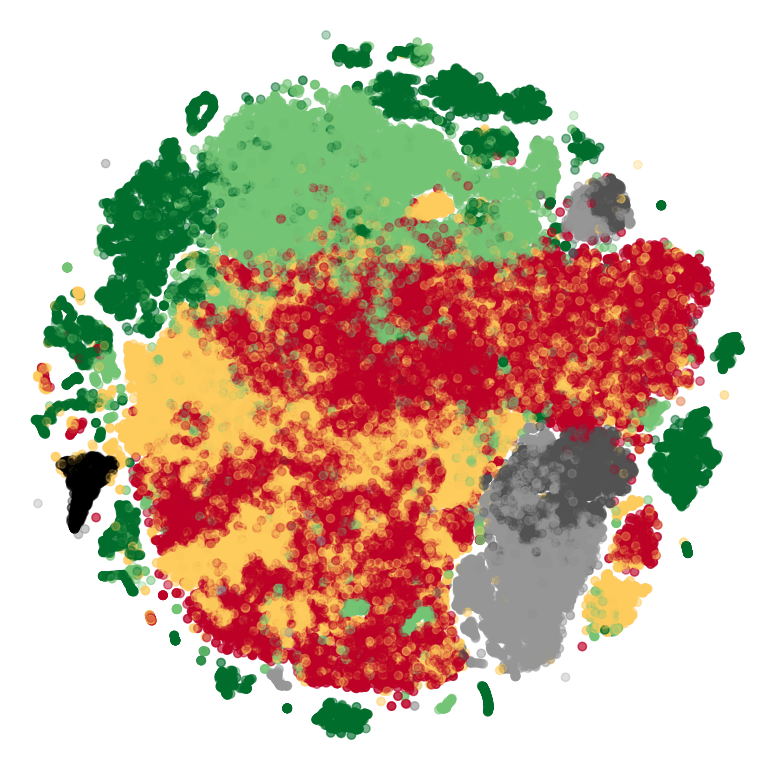}
		\caption{\label{fig:experiments_feature_embedding_masked}Ours}
	\end{subfigure}

	\caption{
		\textbf{Learned Feature Embedding}:
		Shown are the \ac{tSNE} plots for the feature embeddings $z$ of three versions of the metric network.
		(\subref{fig:experiments_feature_embedding_none})~represents the learned features when the metric is trained without an adversary.
		(\subref{fig:experiments_feature_embedding_full})~visualizes when an adversary was used during training, as described in the related literature.
		Our proposed approach is depicted by~(\subref{fig:experiments_feature_embedding_masked}), where the adversary only focuses on data samples from the \Real{} category.
	}

	\label{fig:experiments_feature_embedding}
\end{figure}

\figref{fig:experiments_feature_embedding}~shows plots of the \ac{tSNE} of the neighborhood features $z$, extracted from the three different metric versions.
Each support set is represented by a different color.
If two points or clusters are close, it means their encoding is more similar to those with higher spacing.

The \ac{tSNE} plots show that KITTI and nuScenes form two separable clusters when trained without the adversary~(\figref{fig:experiments_feature_embedding_none}).
In our approach (\figref{fig:experiments_feature_embedding_masked}), they are mixed and form a single cluster.
This shows that our proposed approach matches features from KITTI and nuScenes to the same distribution, whereas the network without the fair setup does not.
Consequently, the fair setup enforces the feature extractor to only encode the overlap of the real dataset features.

\figref{fig:experiments_feature_embedding_full}~visualizes the encoding of the network trained with the adversary but without the filter extension.
Clusters are not clearly separable, even those of support sets from different categories, and more points are randomly scattered.
This indicates impaired classification capabilities and justifies the use of the filter extension.

\subsection{Feature Continuity} 
\label{sec:experiments_transition}

In this section we demonstrate the continuity in transition between support sets in the feature space.
This is an important property when applying the metric to data that does not belong to one of the chosen categories, but stem from a transition in between.
One example is a point cloud generated by a GAN that learns a sim-to-real mapping, but does not yet work perfectly.
For the sake of simplicity, we take the CARLA test split and add different levels of noise to the distance measurements of the point clouds in order to simulate different generative network states in a controllable manner.
The noise is normally distributed with zero mean and varying standard deviation~$\sigma$.

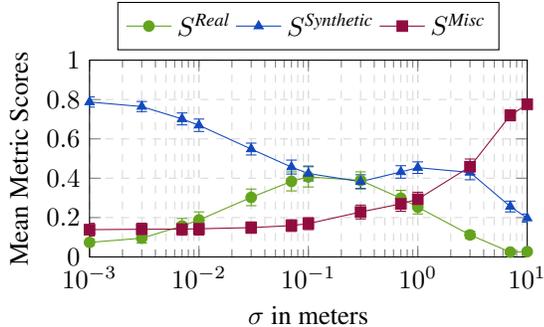
\begin{figure}
    \centering

    \begin{tikzpicture}
    \pgfplotsset{
        width=7.4cm,
        height=4.2cm,
        compat=newest,
        grid style={dashed,gray!30},
    }

    \begin{semilogxaxis}[
        xlabel=$\sigma$ in meters,
        ylabel=Mean Metric Scores,
        ymin=0,
        ymax=1,
        xmin=1e-3,
        xmax=1e1,
        grid=both,
        legend style={at={(0.5,1.05)},anchor=south},
        legend columns=-1,
        legend cell align={left},
    ]

        \addplot[color=real_plot,mark=*,error bars/.cd,y dir=both,y explicit] table[x=Level,y=Real_mean,y error=Real_stddev,col sep=comma] {data/carla_gaussian_metric_results.csv};

        \addplot[color=syn_plot,mark=triangle*,error bars/.cd,y dir=both,y explicit] table[x=Level,y=Syn_mean,y error=Syn_stddev,col sep=comma] {data/carla_gaussian_metric_results.csv};

        \addplot[color=misc_plot,mark=square*,error bars/.cd,y dir=both,y explicit] table[x=Level,y=Misc_mean,y error=Misc_stddev,col sep=comma] {data/carla_gaussian_metric_results.csv};

        \legend{$S^\Real{}$,$S^\Syn{}$,$S^\Misc{}$}
    \end{semilogxaxis}
\end{tikzpicture}

    \caption{
        \textbf{Point Cloud Distortion}:
        Increasing levels of Gaussian noise are applied to the distance measurements of CARLA point clouds.
        For small additive noise levels of a few centimeters, the synthetic CARLA data appears more realistic.
        For higher $\sigma$, the noise is dominant, as indicated by decreasing \Real{} and \Syn{} scores.
    }
    \label{fig:experiments_carla_gaussian_levels}

\end{figure}

\figref{fig:experiments_carla_gaussian_levels}~shows the mean metric scores $S$ over a wide range of additive noise levels applied to CARLA data.
Notably, at low noise levels, the \Real{} score increases.
This reflects the fact that ideal synthetic data needs a certain level of range noise in order to appear more realistic.
On the other hand, at high noise levels, the data barely possesses any structure, as indicated by high \Misc{} and low \Real{} and \Syn{} scores.

We interpret the smooth transitions between the states as an indication for a disentangled latent representation within our metric network.
Further, it shows the necessity for all three support sets when working with real-world and synthetic data.

\subsection{Anomaly Detection} 
\label{sec:experiments_anomaly}

\begin{figure}
	\centering

	\begin{tabular}{lll}
		\textcolor{real_img}{\rule{2.4mm}{2.4mm}} $p_C^\Real{}$ & \textcolor{syn_img}{\rule{2.4mm}{2.4mm}} $p_C^\Syn{}$ & \textcolor{misc_img}{\rule{2.4mm}{2.4mm}} $p_C^\Misc{}$
	\end{tabular}

	\begin{tabular}{c@{\hskip1pt}c}
		\begin{subfigure}{0.48\linewidth}
			\centering
			\begin{tikzpicture}
    \begin{axis}[
        width=1.35\linewidth,
        enlargelimits=false,
        axis on top,
        axis equal image,
        ticks=none,
    ]

    \addplot graphics[xmin=-60,xmax=60,ymin=-40,ymax=40] {experiments_anomaly_pandaset_seq-043_fid-000026.png};
    \addplot graphics[xmin=-2.66,xmax=2.66,ymin=-1.4,ymax=1.4] {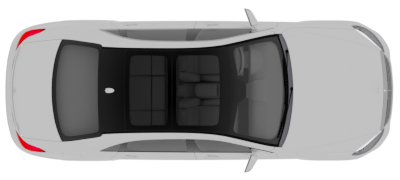};

    \end{axis}
\end{tikzpicture}
			\caption{\label{fig:experiments_anomaly_pandaset}PandaSet}
		\end{subfigure}%
		&%
		\begin{subfigure}{0.48\linewidth}
			\centering
			\begin{tikzpicture}
    \begin{axis}[
        width=1.35\linewidth,
        enlargelimits=false,
        axis on top,
        axis equal image,
        ticks=none,
    ]

    \addplot graphics[xmin=-60,xmax=60,ymin=-40,ymax=40] {experiments_anomaly_kitti_seq-08_fid-000094.png};
    \addplot graphics[xmin=-2.66,xmax=2.66,ymin=-1.4,ymax=1.4] {ego_vehicle_top_view_ortho.png};

    \end{axis}
\end{tikzpicture}
			\caption{\label{fig:experiments_anomaly_kitti}KITTI}
		\end{subfigure}%
		\\%
		\begin{subfigure}{0.48\linewidth}
			\centering
			\begin{tikzpicture}
    \begin{axis}[
        width=1.35\linewidth,
        enlargelimits=false,
        axis on top,
        axis equal image,
        ticks=none,
    ]

    \addplot graphics[xmin=-60,xmax=60,ymin=-40,ymax=40] {experiments_anomaly_nuscenes_scene-0012_sample-000000.png};
    \addplot graphics[xmin=-2.66,xmax=2.66,ymin=-1.4,ymax=1.4] {ego_vehicle_top_view_ortho.png};

    \end{axis}
\end{tikzpicture}
			\caption{\label{fig:experiments_anomaly_nuscenes}nuScenes}
		\end{subfigure}%
		&%
		\begin{subfigure}{0.48\linewidth}
			\centering
			\begin{tikzpicture}
    \begin{axis}[
        width=1.35\linewidth,
        enlargelimits=false,
        axis on top,
        axis equal image,
        ticks=none,
    ]

    \addplot graphics[xmin=-60,xmax=60,ymin=-40,ymax=40] {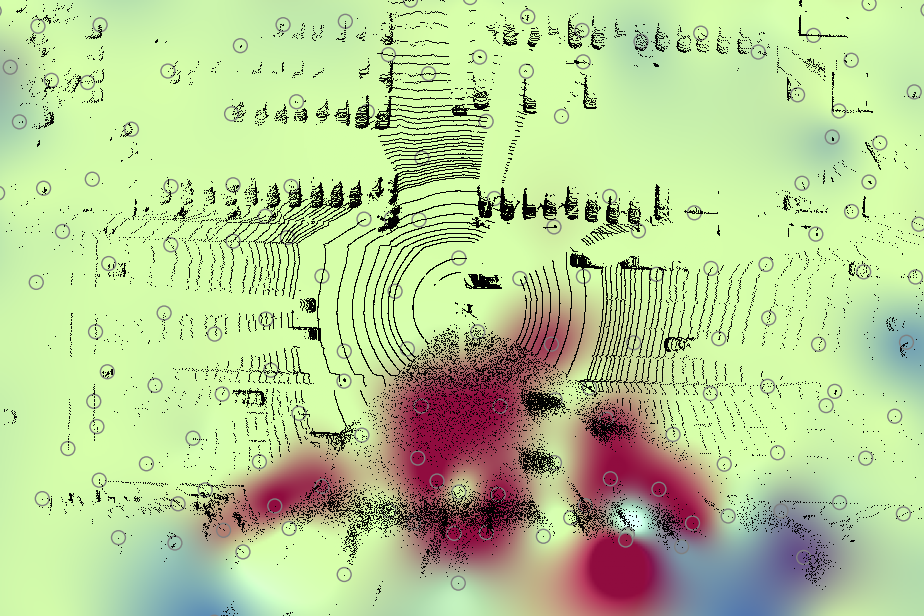};

    \draw[dashed] (0,0) -- (-39,-40);
    \draw[dashed] (0,0) -- (60,-38);

    \addplot graphics[xmin=-2.66,xmax=2.66,ymin=-1.4,ymax=1.4] {ego_vehicle_top_view_ortho.png};

    \end{axis}
\end{tikzpicture}
			\caption{\label{fig:experiments_anomaly_pandaset_noise}PandaSet with noise}
		\end{subfigure}%
		\\
	\end{tabular}

	\caption{
		\textbf{Localization of Anomalies}:
		Example scenes with low \Real{} scores.
		The colors represent interpolated $p_C$ scores which are discrete values located at the query points (gray circles).
		In~(\subref{fig:experiments_anomaly_pandaset}), the purple area marks a road section with extreme elevation changes.
		In the lower half of~(\subref{fig:experiments_anomaly_kitti}), the metric highlights seemingly floating branches of a huge tree, that enter the LiDAR field-of-view from above.
		(\subref{fig:experiments_anomaly_nuscenes})~shows an unusual scene in a dead end road with steep hills surrounding the car.
		(\subref{fig:experiments_anomaly_pandaset_noise})~illustrates a PandaSet sample where the region indicated by dashed lines has been manually distorted with additive Gaussian noise.
	}

	\label{fig:experiments_anomaly_detection}
\end{figure}

Given the locality aspect of our method, we are able to find anomalies or regions with differing appearance within a single scan.
\figref{fig:experiments_anomaly_detection}~shows three examples where our metric outputs the lowest \Real{} score within all test scans and one example where a part of the horizontal field of view is distorted with noise.
The method can successfully identify anomalies within the support sets, \ie KITTI and nuScenes.
To a lesser extent, the method is also capable of identifying similar unusual constellations in completely unseen data (PandaSet).
Weird sensor effects, such as the region altered with additive Gaussian noise in the PandaSet scan, are also detected by the metric (purple areas).

\subsection{Limitations} 
\label{sec:experiments_limitations}

LiDAR data is sparse and our approach is dependent on measurements.
Wherever there is no measurement data, the metric cannot give information about the data quality.
This limits the ability for reliable prediction of very sparse data, for example at high distances $\left(>100~m\right)$ or when information is lost in advance to metric score computation.
On the other hand, this enables the processing of point clouds that are not a full $360^{\circ}$ scan of the world.


\section{Conclusion}

This paper presented a novel metric to quantify the degree of realism of local regions in LiDAR point clouds.
In extensive experiments, we demonstrated the reliability and applicability of our metric on unseen data.
Through adversarial learning, we obtain a feature encoding that is able to adequately capture data realism instead of focusing on dataset specifics.
Our approach provides reliable interpolation capabilities between various levels of realism without requiring annotations.
The predictions of our method correlate well with visual judgment, unlike reconstruction errors serving as a proxy for realism.
In addition, we demonstrated that the local realism score can be used to detect anomalies.

\section*{Acknowledgment}

The research leading to these results is funded by the German Federal Ministry for Economic Affairs and Energy within the project “KI Delta Learning” (F\"orderkennzeichen 19A19013A).

{\small
\bibliographystyle{bibliography/ieee_fullname}
\bibliography{bibliography/bib_long,bibliography/refs}
}

\newpage
\appendix
\section{Appendix}

This appendix covers details of the \ac{DNN} architectures, hyperparameters, evaluation, and additional qualitative results.
\secref{app:method}~contains detailed listings of the hyperparameters of the metric network and an analysis of the training extend as additional information to the architecture in~\secref{sec:method_architecture}.
\secref{app:evaluation}~gives detailed information about the architecture and hyperparameters of the up-sampling networks and contains the additional evaluation results from~\secref{sec:experiments_reconstruction}.
In \secref{app:qualitative}, we provide additional qualitative results to \secref{sec:experiments_transition}, \secref{sec:experiments_anomaly}, and \secref{sec:experiments_limitations}.

\subsection{Metric Implementation, Hyperparameters, and Analysis}
\label{app:method}

\subsubsection{Architecture}

\begin{table*}
    \centering

    \caption{
        \textbf{Network Architecture}:
        Detailed network architecture and input format definition.
        The ID of each row is used to reference the output of the row. $\uparrow$ indicates that the layer directly above is an input.
        $N$ denotes the number of LiDAR measurements.
        $Q_i$ are the number of query points at abstraction level $i$.
        $K_i$ are the number of nearest neighbors to search at abstraction level $i$.
        $U_C$ and $U_A$ are the number of output units of the classifier and adversary, respectively.
    }
    \label{tab:implementation_metric}

    \begin{tabularx}{\linewidth}{rlll p{6.8cm}}
        \toprule
        \textbf{ID} & \textbf{Inputs} & \textbf{Operation} & \textbf{Output Shape} & \textbf{Description} \\
        \midrule
        1 & LiDAR & $x$, $y$, $z$ & $[N \times 3]$ & Position of each point relative to sensor origin \\
        \midrule
        \multicolumn{5}{c}{\textbf{Feature Extractor: Abstraction Module 1}} \\
        \midrule
        2 & $\uparrow$, $Q_1$ & Farthest point sampling & $[2048]$ & Indices of $Q_1$ query points \\
        3 & 1, $\uparrow$ & Group & $[2048 \times 3]$ & Grouped sampled points \\
        4 & 1, 2, $K_1$ & Nearest neighbor search & $[2048 \times 10]$ & Indices of the $K_1$ nearest neigbors per query \\
        5 & 1, 2, $\uparrow$ & Group & $[2048 \times 10 \times 3]$ & Grouped neighborhoods \\
        6 & $\uparrow$ & Neighborhood normalization & $[2048 \times 10 \times 3]$ & Translation normalization towards query point \\
        7 & $\uparrow$ & (Conv+LeakyReLU) $\times\!2$ & $[2048 \times 10 \times 64]$ & Kernel size $1\!\times\!1$, stride 1 \\
        8 & $\uparrow$ & Conv+LeakyReLU & $[2048 \times 10 \times 128]$ & Kernel size $1\!\times\!1$, stride 1 \\
        9 & $\uparrow$ & ReduceMax & $[2048 \times 128]$ & Maximum over neighborhood features \\
        \midrule
        \multicolumn{5}{c}{\textbf{Feature Extractor: Abstraction Module 2}} \\
        \midrule
        10 & 3, $Q_2$ & Farthest point sampling & $[256]$ & Indices of $Q_2$ query points \\
        11 & 3, 10, $K_2$ & Nearest neighbor search & $[256 \times 10]$ & Indices of the $K_2$ nearest neighbors per query \\
        12 & 3, 10, $\uparrow$ & Group & $[256 \times 10 \times 3]$ & Grouped neighborhoods \\
        13 & $\uparrow$ & Neighborhood normalization & $[256 \times 10 \times 3]$ & Translation normalization towards query point \\
        14 & 9, 11 & Group & $[256 \times 10 \times 128]$ & Grouped features \\
        15 & 13, $\uparrow$ & Concat features & $[256 \times 10 \times 131]$ & Grouped features with $xyz$ \\
        16 & $\uparrow$ & (Conv+LeakyReLU) $\times\!2$ & $[256 \times 10 \times 128]$ & Kernel size $1\!\times\!1$, stride 1 \\
        17 & $\uparrow$ & Conv+LeakyReLU & $[256 \times 10 \times 256]$ & Kernel size $1\!\times\!1$, stride 1 \\
        18 & $\uparrow$ & ReduceMax & $[256 \times 256]$ & Maximum over neighborhood features $\rightarrow$ latent representation $z$\\
        \midrule
        \multicolumn{5}{c}{\textbf{Classifier / Adversary}} \\
        \midrule
        19 & $\uparrow$ & Dense+LeakyReLU & $[256 \times 128]$ & \\
        20 & $\uparrow$ & Dropout & $[256 \times 128]$ & Dropout ratio 50\% \\
        21 & $\uparrow$ & Dense & $[256 \times U_{C,A}]$ & Output logits vector $y_{C,A}$ \\
        22 & $\uparrow$ & Softmax & $[256 \times U_{C,A}]$ & Output probability vector $p_{C,A}$ \\
        \bottomrule
    \end{tabularx}

\end{table*}

\tabref{tab:implementation_metric}~lists all layers, inputs, and operations of our \ac{DNN} architecture.
We use TensorFlow to implement online data processing, neural network weight optimization, and network inference.
The implementation is oriented on the original PointNet++ implementation~\cite{Qi2017NIPS}\footnote{PointNet++ code \url{https://github.com/charlesq34/pointnet2}}.
The Adam optimizer is used for optimization.
We use an initial learning rate of $1e^{-3}$ with exponential warm-up and decay.

In contrast to PointNet++, we use KNN search instead of radius search.
PointNet++ operates on point clouds from the ShapeNet dataset, which contains uniformly sampled points on object surfaces.
In LiDAR point clouds, points are not uniformly distributed and with increasing distance to the sensor, also the distance between neighboring points increase.
We found KNN search more practical to obtain meaningful neighborhoods in LiDAR point clouds compared to radius search.

The outputs of the classifier $C$ and adversary $A$ have $U_C$ and $U_A$ channels, respectively.
The classifier outputs the scores for each of the $U_C=3$ categories, namely \Real{}, \Syn{}, \Misc{}.
In the adversary, the output has $U_A=7$ channels, one for each of the support sets.
However, \figref{fig:architecture} shows only outputs for the two support sets from the \Real{} category.
For simplicity, the details of the implementation are not visualized in the respective figure and are indicated by the filter triangle in the path between the feature extractor and the adversary.
The filter is implemented as a class weighting when computing the loss from the adversary output.
The class weights $w$ for each dataset $d$ are set to
\begin{equation*}
    w_d =
    \begin{cases}
        1, & \text{if } d \in \Real{} \\
        0, & \text{otherwise}
    \end{cases}
    \quad .
\end{equation*}
We found this the easiest and most stable way to implement the desired behavior in TensorFlow graph mode.

\subsubsection{Choosing the correct loss factor $\lambda$}

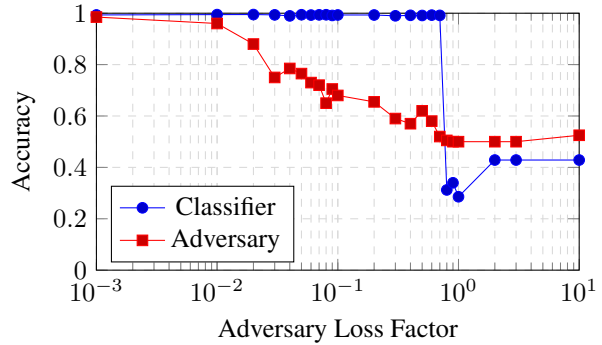
\begin{figure}
    \centering

    \begin{tikzpicture}
        \pgfplotsset{
            width=8cm,
            height=5cm,
            compat=newest,
            grid style={dashed,gray!30},
        }

        \begin{semilogxaxis}[
            xlabel=Adversary Loss Factor,
            ylabel=Accuracy,
            xmin=1e-3,
            xmax=1e1,
            ymin=0,
            ymax=1,
            grid=both,
            legend pos=south west,
        ]

            \addplot table[x=factor,y=classifier,col sep=comma] {data/accuracy_over_factor.csv};
            \addplot table[x=factor,y=adversary,col sep=comma] {data/accuracy_over_factor.csv};

            \legend{Classifier,Adversary}
        \end{semilogxaxis}
    \end{tikzpicture}

    \caption{
        \textbf{Ratio of accuracy and fairness}:
        Accuracy of classifier and adversary over the loss factor~$\lambda$.
        At small $\lambda$, both accuracies are high which means high classification performance but no fairness.
        With increasing $\lambda$ the network gets fairer while maintaining its high level of classification accuracy.
        At a certain point the network becomes unstable and crashes into chance level performance in the classifier.
    }
    \label{fig:experiments_accuracy_over_factor}
\end{figure}

\figref{fig:experiments_accuracy_over_factor}~shows the accuracy of the two output heads over varying loss factor~$\lambda$.
A good ratio between network accuracy and fairness is indicated by a large offset between the classifier and the adversary accuracy.
We use a factor of $\lambda=0.3$ for all our experiments in the main paper.
Here, classifier accuracy is at 99\% while adversary accuracy has dropped to almost chance level (59\%).
A perfectly fair setup at maximum performance is equal to 100\% classifier accuracy and 50\% adversary accuracy, in our case.

The reason why the adversary accuracy drops so suddenly when increasing $\lambda$ is caused by the direct correlation between the desired and the sensitive attribute, as explained in~\secref{sec:method_fairness}.

\subsubsection{Theoretical Lower Bound}

Classification accuracy $\operatorname{ACC}$ is defined in range $[0,1]$.
However, the lower bound is actually $\frac{1}{U}$ with $U$ being the number of classes.
Considering the fair learning setup with $U_C=3$ and $U_A=7$, without our class weighting extension, one would assume that if the network has the best possible performance and the best possible fairness, the accuracies result to $\operatorname{ACC}_C \approx 1$ and $\operatorname{ACC}_A \approx \frac{1}{U_A}=\frac{1}{7}$.
Due to the direct correspondence between categories and datasets this is not the case for the adversary.
The resulting confusion matrices are schematically illustrated in~\figref{fig:lower_bound_confusion}.
The pseudo-diagonal for the adversary is caused by the perfect classification capabilities of the classifier, which prevents a perfect confusion in the adversary.
The new lower bound of the adversary accuracy in this state can now be formulated as
\begin{equation*}
    \operatorname{ACC}_A \approx \frac{U_C}{U_A}=\frac{3}{7}\approx42.9\%
\end{equation*}
which is considerably higher than $\operatorname{ACC}_A \approx \frac{1}{U_A}=\frac{1}{7}\approx14.3\%$.

\begin{figure}
    \centering

    \resizebox{\linewidth}{!}{
    \begin{tikzpicture}
    \tikzset{
        field/.pic = {
            \def\w{0.7};
            \draw[pic actions] (-\w/2,-\w/2) rectangle +(\w,\w);
            \node[draw=none, text=black, opacity=1.0, anchor=center] at (0,0) {\tikzpictext};
        }
    }

    \foreach \x in {1,2,3} {
        \pic[draw=none, pic text={$\tilde{K}_\x$}] at (-0.7+\x*0.7,2*0.7) {field};
    }
    \pic[draw=none, pic text={$K_1$}] at (-1*0.7,0.7) {field};
    \pic[draw=none, pic text={$K_2$}] at (-1*0.7,0) {field};
    \pic[draw=none, pic text={$K_3$}] at (-1*0.7,-0.7) {field};
    \pic[fill=blue!10!white, pic text={$1$}] at (0*0.7,0.7) {field};
    \pic[fill=white, pic text={$0$}] at (1*0.7,0.7) {field};
    \pic[fill=white, pic text={$0$}] at (2*0.7,0.7) {field};
    \pic[fill=white, pic text={$0$}] at (0*0.7,0) {field};
    \pic[fill=orange!20!white, pic text={$1$}] at (1*0.7,0) {field};
    \pic[fill=white, pic text={$0$}] at (2*0.7,0) {field};
    \pic[fill=white, pic text={$0$}] at (0*0.7,-0.7) {field};
    \pic[fill=white, pic text={$0$}] at (1*0.7,-0.7) {field};
    \pic[fill=red!10!white, pic text={$1$}] at (2*0.7,-0.7) {field};

    \pic[draw=none, pic text={$\tilde{K}_{11}$}] at (5+0*0.7,4*0.7) {field};
    \pic[draw=none, pic text={$\tilde{K}_{12}$}] at (5+1*0.7,4*0.7) {field};
    \pic[draw=none, pic text={$\tilde{K}_{21}$}] at (5+2*0.7,4*0.7) {field};
    \pic[draw=none, pic text={$\tilde{K}_{22}$}] at (5+3*0.7,4*0.7) {field};
    \pic[draw=none, pic text={$\tilde{K}_{31}$}] at (5+4*0.7,4*0.7) {field};
    \pic[draw=none, pic text={$\tilde{K}_{32}$}] at (5+5*0.7,4*0.7) {field};
    \pic[draw=none, pic text={$\tilde{K}_{33}$}] at (5+6*0.7,4*0.7) {field};
    \pic[draw=none, pic text={$K_{11}$}] at (5-0.7,3*0.7) {field};
    \pic[draw=none, pic text={$K_{12}$}] at (5-0.7,2*0.7) {field};
    \pic[draw=none, pic text={$K_{21}$}] at (5-0.7,1*0.7) {field};
    \pic[draw=none, pic text={$K_{22}$}] at (5-0.7,0*0.7) {field};
    \pic[draw=none, pic text={$K_{31}$}] at (5-0.7,-1*0.7) {field};
    \pic[draw=none, pic text={$K_{32}$}] at (5-0.7,-2*0.7) {field};
    \pic[draw=none, pic text={$K_{33}$}] at (5-0.7,-3*0.7) {field};
    \pic[fill=blue!10!white, pic text={$\frac{1}{2}$}] at (5+0*0.7,3*0.7) {field};
    \pic[fill=blue!10!white, pic text={$\frac{1}{2}$}] at (5+1*0.7,3*0.7) {field};
    \foreach \x in {2,...,6}{
        \pic[fill=white, pic text={$0$}] at (5+\x*0.7,3*0.7) {field};
    }
    \pic[fill=blue!10!white, pic text={$\frac{1}{2}$}] at (5+0*0.7,2*0.7) {field};
    \pic[fill=blue!10!white, pic text={$\frac{1}{2}$}] at (5+1*0.7,2*0.7) {field};
    \foreach \x in {2,...,6}{
        \pic[fill=white, pic text={$0$}] at (5+\x*0.7,2*0.7) {field};
    }
    \pic[fill=orange!20!white, pic text={$\frac{1}{2}$}] at (5+2*0.7,0.7) {field};
    \pic[fill=orange!20!white, pic text={$\frac{1}{2}$}] at (5+3*0.7,0.7) {field};
    \foreach \x in {0,1,4,5,6}{
        \pic[fill=white, pic text={$0$}] at (5+\x*0.7,0.7) {field};
    }
    \pic[fill=orange!20!white, pic text={$\frac{1}{2}$}] at (5+2*0.7,0) {field};
    \pic[fill=orange!20!white, pic text={$\frac{1}{2}$}] at (5+3*0.7,0) {field};
    \foreach \x in {0,1,4,5,6}{
        \pic[fill=white, pic text={$0$}] at (5+\x*0.7,0) {field};
    }
    \foreach \x in {0,...,3}{
        \pic[fill=white, pic text={$0$}] at (5+\x*0.7,-0.7) {field};
    }
    \foreach \x in {4,5,6}{
        \pic[fill=red!10!white, pic text={$\frac{1}{3}$}] at (5+\x*0.7,-0.7) {field};
    }
    \foreach \x in {0,...,3}{
        \pic[fill=white, pic text={$0$}] at (5+\x*0.7,-2*0.7) {field};
    }
    \foreach \x in {4,5,6}{
        \pic[fill=red!10!white, pic text={$\frac{1}{3}$}] at (5+\x*0.7,-2*0.7) {field};
    }
    \foreach \x in {0,...,3}{
        \pic[fill=white, pic text={$0$}] at (5+\x*0.7,-3*0.7) {field};
    }
    \foreach \x in {4,5,6}{
        \pic[fill=red!10!white, pic text={$\frac{1}{3}$}] at (5+\x*0.7,-3*0.7) {field};
    }

    \pic[draw=none, pic text={$\tilde{K}_1$}] at (5+0.5*0.7,5*0.7) {field};
    \draw[blue] (5-0.3,4.5*0.7) -- +(2*0.7-0.2,0);

    \pic[draw=none, pic text={$\tilde{K}_2$}] at (5+2.5*0.7,5*0.7) {field};
    \draw[orange] (5+1.5*0.7+0.1,4.5*0.7) -- +(2*0.7-0.2,0);

    \pic[draw=none, pic text={$\tilde{K}_3$}] at (5+5*0.7,5*0.7) {field};
    \draw[red] (5+3.5*0.7+0.1,4.5*0.7) -- +(3*0.7-0.2,0);

    \pic[draw=none, pic text={$K_1$}] at (5-2*0.7,3*0.7-0.4) {field};
    \draw[blue] (5-1.5*0.7,2*0.7-0.3) -- +(0,2*0.7-0.2);

    \pic[draw=none, pic text={$K_2$}] at (5-2*0.7,1*0.7-0.4) {field};
    \draw[orange] (5-1.5*0.7,-0.3) -- +(0,2*0.7-0.2);

    \pic[draw=none, pic text={$K_3$}] at (5-2*0.7,-2*0.7) {field};
    \draw[red] (5-1.5*0.7,-3*0.7-0.3) -- +(0,3*0.7-0.2);

    \end{tikzpicture}
    }  

    \caption{
    \textbf{Lower Bound Confusion Matrix}:
    The figure shows schematic confusion matrices for the classifier (left) and the adversary (right).
    Here, the classifier has $U_C=3$ categories ($K_1$, $K_2$, $K_3$) and the adversary has $U_A=7$ output channels for the respective datasets of the categories.
    If the network is trained with maximum accuracy (100\%), the classifier confusion matrix is a diagonal matrix, as shown on the left.
    Assuming the adversary is maximally confused in this state, a pseudo-diagonal confusion emerges, increasing the theoretical lower bound of adversary accuracy from $\frac{1}{U_A}$ to $\frac{U_C}{U_A}$.
    }
    \label{fig:lower_bound_confusion}
\end{figure}

Our proposed method does not consider the datasets from other categories than \Real{}.
Therefore, the theoretical lower bound of the adversary remains at
\begin{equation*}
    \operatorname{ACC}_A \approx \frac{1}{\# \text{~datasets in \Real{}}}=\frac{1}{2}=50\% \quad .
\end{equation*}

\subsection{Metric Verification on Reconstructed Data}
\label{app:evaluation}

This section gives additional details on the up-sampling experiments for metric verification of~\secref{sec:experiments_reconstruction}.
The up-sampling process is based on cylindrical depth projections of the LiDAR point clouds.
Only the vertical resolution of the LiDAR images is enhanced.
The bilinear interpolation is a traditional approach for which we directly used the resize method from TensorFlow (\texttt{tf.image.resize(images, size, method=ResizeMethod.BILINEAR)}).
For all other experiments, we used the generator from the SRGAN architecture~\cite{Ledig2017CVPR} and for the GAN experiments, also the discriminator architecture.
After being processed by the super-resolution networks, the generated point clouds are converted back into lists of points and are fed to the metric network for realism judgement.

\subsubsection{Implementation Details}

\begin{table*}
    \centering

    \caption{
        \textbf{SRGAN Generator Architecture}:
        Detailed network architecture and input format definition of the SRGAN generator~\cite{Ledig2017CVPR}.
        The ID of each row is used to reference the output of the row.
        $\uparrow$ indicates that the layer directly above is an input.
        $N$ denotes the number of measured LiDAR points.
        $H$ denotes the number of layers in the LiDAR sensor and $W$ are the number of layer pulses fired per $360^\circ$ revolution.
        The cylindrical depth projection is either retrieved directly from the raw image of the sensor or with a back-projection by computing $(r,\varphi,\theta)$ from $(x,y,z)$.
        Missing measurements are set to a constant distance in the dense projection and are masked in the loss computation.
    }
    \label{tab:implementation_eval_generator}

    \begin{tabularx}{\linewidth}{rlll p{6.8cm}}
        \toprule
        \textbf{ID} & \textbf{Inputs} & \textbf{Operation} & \textbf{Output Shape} & \textbf{Description} \\
        \midrule
        \multicolumn{5}{c}{\textbf{Input features from LiDAR scan}} \\
        \midrule
        1 & LiDAR & $x$, $y$, $z$ & $[N \times 3]$ & Position of each point relative to sensor origin \\
        2 & $\uparrow$ & Projection $(x,y,z) \rightarrow (r,\varphi,\theta)$ & $[H,W,1]$ & Cylindrical depth projection $r$ with $\theta$ over $H$ and $\varphi$ over $W$ \\
        \midrule
        \multicolumn{5}{c}{\textbf{Residual blocks}} \\
        \midrule
        3 & $\uparrow$ & Conv+ParametricReLU & $[H,W,64]$ & Kernel size $9\!\times\!9$, stride 1 \\
        4 & $\uparrow$ & Conv+BN+ParametricReLU & $[H,W,64]$ & Kernel size $3\!\times\!3$, stride 1 \\
        5 & $\uparrow$ & Conv+BN & $[H,W,64]$ & Kernel size $3\!\times\!3$, stride 1 \\
        6 & $\uparrow$, 3 & Add & $[H,W,64]$ & Element-wise addition \\
        7 & $\uparrow$ & Repeat steps (4-6) & $[H,W,64]$ & $\times\!16$ repetition of residual blocks \\
        8 & $\uparrow$ & Conv+BN & $[H,W,64]$ & Kernel size $3\!\times\!3$, stride 1 \\
        9 & $\uparrow$, 3 & Add & $[H,W,64]$ & Element-wise addition \\
        \midrule
        \multicolumn{5}{c}{\textbf{Super-resolution blocks}} \\
        \midrule
        10 & $\uparrow$ & Conv & $[H,W,256]$ & Kernel size $3\!\times\!3$, stride 1 \\
        11 & $\uparrow$ & SubpixelShuffle & $[2 \cdot H,W,128]$ & Reshape by moving values from the channel dimension to the spatial dimension \\
        12 & $\uparrow$ & ParametricReLU & $[2 \cdot H,W,128]$ & \\
        13 & $\uparrow$ & Repeat steps (10-12) & $[f_\text{up} \cdot H,W,128]$ & $\times\!\log_2 f_\text{up}$ repetition with $f_\text{up}$ being the desired up-sampling factor, \ie $f_\text{up}=\{2,4,8\}$ \\
        14 & $\uparrow$ & Conv & $[f_\text{up} \cdot H,W,1]$ & Kernel size $9\!\times\!9$, stride 1 \\
        \bottomrule
    \end{tabularx}

\end{table*}

\begin{table*}
    \centering

    \caption{
        \textbf{SRGAN Discriminator Architecture}:
        Detailed network architecture and input format definition of the SRGAN discriminator~\cite{Ledig2017CVPR}.
        The input to the network is either the ground truth $r^\text{gt}$ or the prediction from the generator $r^\text{hr}$.
    }
    \label{tab:implementation_eval_discriminator}

    \begin{tabularx}{\linewidth}{rlll p{6.8cm}}
        \toprule
        \textbf{ID} & \textbf{Inputs} & \textbf{Operation} & \textbf{Output Shape} & \textbf{Description} \\
        \midrule
        1 & LiDAR & $r^\text{gt}$ or $r^\text{hr}$ & $[f_\text{up} \cdot H,W,1]$ & High-resolution cylindrical depth projection \\
        \midrule
        \multicolumn{5}{c}{\textbf{Conv blocks}} \\
        \midrule
        2 & $\uparrow$ & Conv+LeakyReLU & $[f_\text{up} \cdot H,W,64]$ & Kernel size $3\!\times\!3$, stride 1 \\
        3 & $\uparrow$ & Conv+BN+LeakyReLU & $[\frac{f_\text{up}}{2} H,\frac{1}{4} W,64]$ & Kernel size $5\!\times\!5$, strides $2\!\times\!4$ \\
        4 & $\uparrow$ & Conv+BN+LeakyReLU & $[\frac{f_\text{up}}{2} H,\frac{1}{4} W,128]$ & Kernel size $3\!\times\!3$, stride 1 \\
        5 & $\uparrow$ & Conv+BN+LeakyReLU & $[\frac{f_\text{up}}{4} H,\frac{1}{8} W,128]$ & Kernel size $3\!\times\!3$, stride 2 \\
        6 & $\uparrow$ & Conv+BN+LeakyReLU & $[\frac{f_\text{up}}{4} H,\frac{1}{8} W,256]$ & Kernel size $3\!\times\!3$, stride 1 \\
        7 & $\uparrow$ & Conv+BN+LeakyReLU & $[\frac{f_\text{up}}{4} H,\frac{1}{16} W,256]$ & Kernel size $3\!\times\!3$, strides $1\!\times\!2$ \\
        8 & $\uparrow$ & Conv+BN+LeakyReLU & $[\frac{f_\text{up}}{4} H,\frac{1}{16} W,512]$ & Kernel size $3\!\times\!3$, stride 1 \\
        9 & $\uparrow$ & Conv+BN+LeakyReLU & $[\frac{f_\text{up}}{8} H,\frac{1}{32} W,512]$ & Kernel size $3\!\times\!3$, stride 2 \\
        \midrule
        \multicolumn{5}{c}{\textbf{Reduction}} \\
        \midrule
        10 & $\uparrow$ & Flatten & $[\frac{f_\text{up}}{2} \cdot H \cdot W]$ & \\
        11 & $\uparrow$ & Dense+LeakyReLU & $[1024]$ & \\
        12 & $\uparrow$ & Dense & $[1]$ & \\
        \bottomrule
    \end{tabularx}

\end{table*}

\tabref{tab:implementation_eval_generator}~lists all layers, inputs, and operations of the SRGAN generator architecture.
In the $\mathcal{L}_{\{1,2\}}$-CNN trainings, a weighted $\mathcal{L}_\alpha$ loss is minimized.
The objective is formulated as
\begin{equation*}
    \min_{\theta_G} \mathcal{L}_\alpha = \min_{\theta_G} \frac{1}{\alpha |\gamma|} \sum_{(i,j)\in\gamma} \left| r_{i,j}^\text{gt} - r_{i,j}^\text{hr} \right|
\end{equation*}
with the set of measured points $\gamma$, and $r^\text{gt}$ being the high-resolution Ground Truth and $r^\text{hr}$ the prediction
\begin{equation*}
    r^\text{hr} = G_{\theta_G} \left( r^\text{lr} \right)
\end{equation*}
from the low-resolution input $r^\text{lr}$.

\tabref{tab:implementation_eval_discriminator}~lists all layers, inputs, and operations of the SRGAN discriminator architecture.
Here, an adversarial loss, defined as
\begin{equation*}
    \min_{\theta_G} \max_{\theta_D}
    \left\{
    \log{\left[ D_{\theta_D} \left( r^\text{gt} \right) \right]} +
    \log{\left[ 1 - D_{\theta_D} \left( G_{\theta_G} \left( r^\text{lr} \right) \right) \right]}
    \right\}
\end{equation*}
is minimized.
The Adam optimizer is used for optimization with an initial learning rate of $1e^{-3}$.

\subsubsection{Additional Experimental Results}

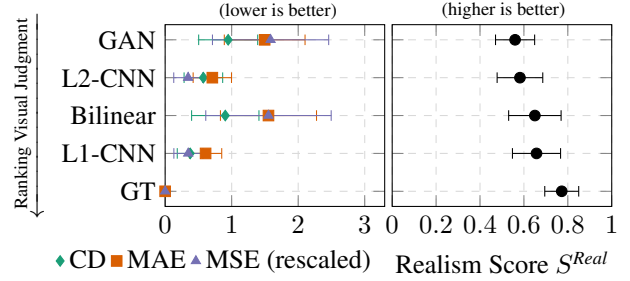
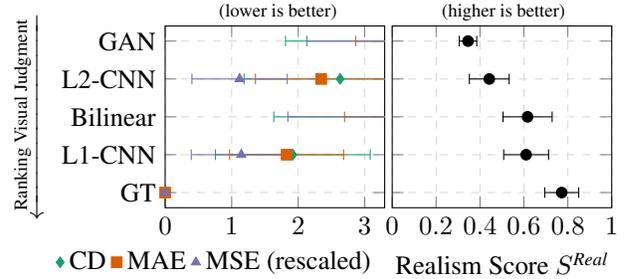
\begin{figure}
    \centering

    \begin{subfigure}{\linewidth}
        \centering
        \pgfplotstableread[col sep=comma]{data/upsampling_metric_results_2x.csv}\datatable

\begin{tikzpicture}
    \pgfplotsset{
        compat=newest,
        grid style={dashed,gray!30},
    }

    \node[] at (1.5,2.6) {\scriptsize (lower is better)\par};

    \begin{axis}[
        name=plot1,
        height=4cm,
        width=4.5cm,
        ytick=data,
        yticklabels from table={\datatable}{Method},
        ylabel=$\xleftarrow{\text{Ranking Visual Judgment}}$,
        xmin=0,
        xmax=3.3,
        grid=both,
        legend style={at={(1.0,-0.3)},anchor=east,draw=none},
        legend columns=-1,
        legend cell align={left},
    ]

        \addplot[only marks,color=cd_plot,mark=diamond*,error bars/.cd,x dir=both,x explicit]
        table[y expr=\coordindex,x=CD_mean,x error=CD_std,col sep=comma] {data/upsampling_metric_results_2x.csv};

        \addplot[only marks,color=mae_plot,mark=square*,error bars/.cd,x dir=both,x explicit]
        table[y expr=\coordindex,x=MAE_mean,x error=MAE_std,col sep=comma] {data/upsampling_metric_results_2x.csv};

        \addplot[only marks,color=mse_plot,mark=triangle*,error bars/.cd,x dir=both,x explicit]
        table[y expr=\coordindex,x expr=\thisrow{MSE_mean}/50,x error expr=\thisrow{MSE_std}/50,col sep=comma] {data/upsampling_metric_results_2x.csv};

        \legend{CD,MAE,MSE (rescaled)}
    \end{axis}

    \node[] at (4.5,2.6) {\scriptsize (higher is better)\par};

    \begin{axis}[
        name=plot2,at={($(plot1.east)+(0.1cm,0)$)},
        anchor=west,
        height=4cm,
        width=4.5cm,
        xlabel=Realism Score $S^\Real{}$,
        ytick=data,
        yticklabels=\empty,
        xmin=0,
        xmax=1,
        grid=both,
    ]

        \addplot[black,mark=*,only marks,error bars/.cd,x dir=both,x explicit]
        table[y expr=\coordindex,x=Real_mean,x error=Real_std,col sep=comma] {data/upsampling_metric_results_2x.csv};

    \end{axis}
\end{tikzpicture}
        \caption{\label{fig:experiments_upsampling_metric_2x}$f_\text{up}=2$}
    \end{subfigure}

    \vspace{1em}

    \begin{subfigure}{\linewidth}
        \centering
        \pgfplotstableread[col sep=comma]{data/upsampling_metric_results_8x.csv}\datatable

\begin{tikzpicture}
    \pgfplotsset{
        compat=newest,
        grid style={dashed,gray!30},
    }

    \node[] at (1.5,2.6) {\scriptsize (lower is better)\par};

    \begin{axis}[
        name=plot1,
        height=4cm,
        width=4.5cm,
        ytick=data,
        yticklabels from table={\datatable}{Method},
        ylabel=$\xleftarrow{\text{Ranking Visual Judgment}}$,
        xmin=0,
        xmax=3.3,
        grid=both,
        legend style={at={(1.0,-0.3)},anchor=east,draw=none},
        legend columns=-1,
        legend cell align={left},
    ]

        \addplot[only marks,color=cd_plot,mark=diamond*,error bars/.cd,x dir=both,x explicit]
        table[y expr=\coordindex,x=CD_mean,x error=CD_std,col sep=comma] {data/upsampling_metric_results_8x.csv};

        \addplot[only marks,color=mae_plot,mark=square*,error bars/.cd,x dir=both,x explicit]
        table[y expr=\coordindex,x=MAE_mean,x error=MAE_std,col sep=comma] {data/upsampling_metric_results_8x.csv};

        \addplot[only marks,color=mse_plot,mark=triangle*,error bars/.cd,x dir=both,x explicit]
        table[y expr=\coordindex,x expr=\thisrow{MSE_mean}/50,x error expr=\thisrow{MSE_std}/50,col sep=comma] {data/upsampling_metric_results_8x.csv};

        \legend{CD,MAE,MSE (rescaled)}
    \end{axis}

    \node[] at (4.5,2.6) {\scriptsize (higher is better)\par};

    \begin{axis}[
        name=plot2,at={($(plot1.east)+(0.1cm,0)$)},
        anchor=west,
        height=4cm,
        width=4.5cm,
        xlabel=Realism Score $S^\Real{}$,
        ytick=data,
        yticklabels=\empty,
        xmin=0,
        xmax=1,
        grid=both,
    ]

        \addplot[black,mark=*,only marks,error bars/.cd,x dir=both,x explicit]
        table[y expr=\coordindex,x=Real_mean,x error=Real_std,col sep=comma] {data/upsampling_metric_results_8x.csv};

    \end{axis}
\end{tikzpicture}
        \caption{\label{fig:experiments_upsampling_metric_8x}$f_\text{up}=8$}
    \end{subfigure}

    \caption{
        \textbf{Metric Scores over Up-Sampling Methods}:
        The two plots each show four methods to perform LiDAR scan up-sampling and the high-resolution Ground Truth.
        $f_\text{up}$ is the up-sampling factor.
        The left side shows different reconstruction errors (lower is better) whereas the right side shows our realism score (higher is better).
    }
    \label{fig:experiments_upsampling_metric}
\end{figure}

In~\secref{sec:experiments_reconstruction}, we only showed the metric results on the $4\!~\!\times$ up-sampling results.
\figref{fig:experiments_upsampling_metric}~additionally shows the metric evaluation results on $2\!~\!\times$ and $8\!~\!\times$ up-sampling.
With increasing factor $f_\text{up}=\{2,4,8\}$ the generation quality of the networks degrades.
Depending on the up-sampling method this degradation is stronger or weaker, but the overall ranking of the methods remains the same.

\subsection{Qualitative Results}
\label{app:qualitative}

\begin{figure*}
    \centering

    \begin{tabular}{c@{\hskip3pt}c@{\hskip3pt}c@{\hskip3pt}c}

        \textcolor{real_img}{\rule{2.4mm}{2.4mm}} 0.04,
        \textcolor{syn_img}{\rule{2.4mm}{2.4mm}} 0.96,
        \textcolor{misc_img}{\rule{2.4mm}{2.4mm}} 0.00
        &
        \textcolor{real_img}{\rule{2.4mm}{2.4mm}} 0.15,
        \textcolor{syn_img}{\rule{2.4mm}{2.4mm}} 0.64,
        \textcolor{misc_img}{\rule{2.4mm}{2.4mm}} 0.21
        &
        \textcolor{real_img}{\rule{2.4mm}{2.4mm}} 0.27,
        \textcolor{syn_img}{\rule{2.4mm}{2.4mm}} 0.56,
        \textcolor{misc_img}{\rule{2.4mm}{2.4mm}} 0.21
        &
        \textcolor{real_img}{\rule{2.4mm}{2.4mm}} 0.30,
        \textcolor{syn_img}{\rule{2.4mm}{2.4mm}} 0.49,
        \textcolor{misc_img}{\rule{2.4mm}{2.4mm}} 0.21
        \\
        \begin{subfigure}{0.24\linewidth}
            \centering
            \frame{\includegraphics[width=\linewidth]{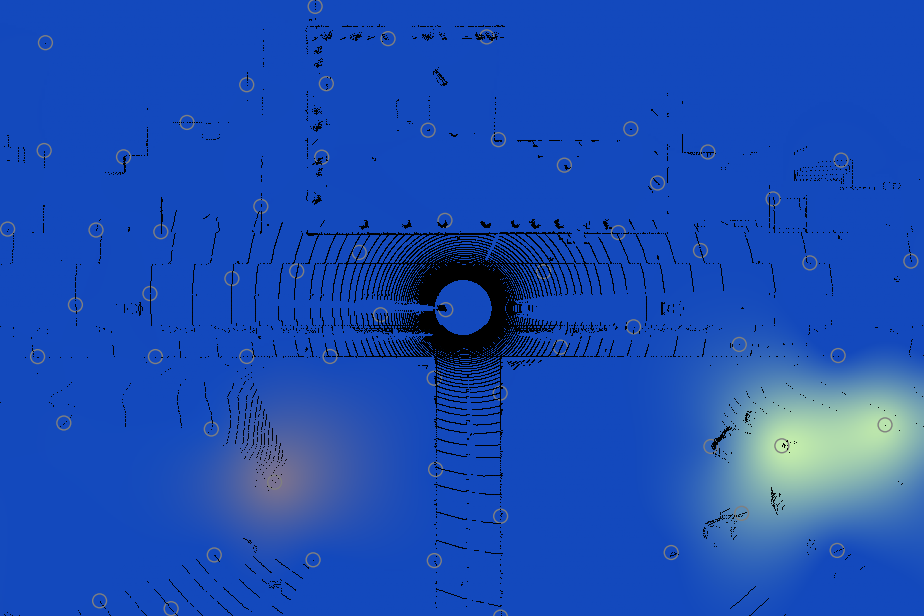}}
            \caption{No noise}
        \end{subfigure}
        &
        \begin{subfigure}{0.24\linewidth}
            \centering
            \frame{\includegraphics[width=\linewidth]{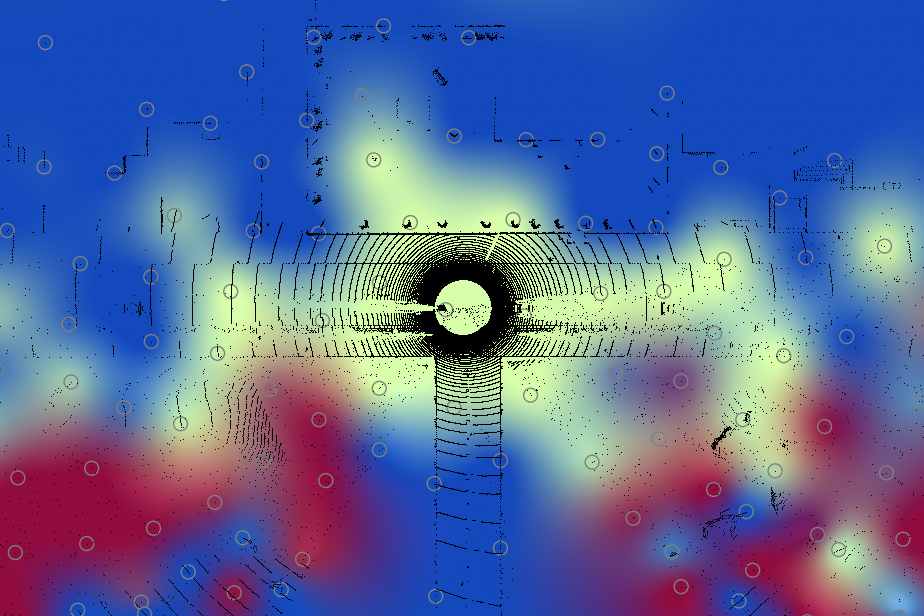}}
            \caption{$\sigma=0.01$}
        \end{subfigure}
        &
        \begin{subfigure}{0.24\linewidth}
            \centering
            \frame{\includegraphics[width=\linewidth]{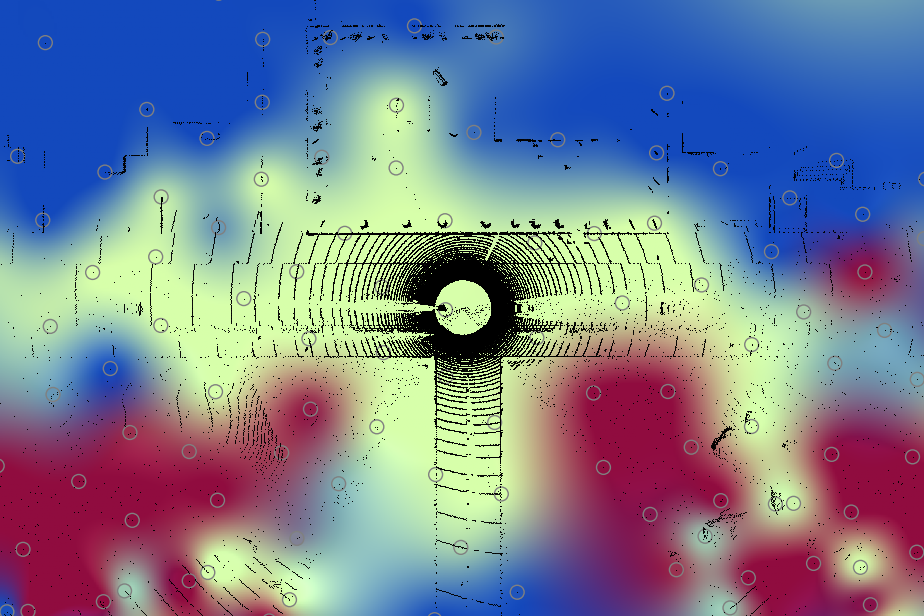}}
            \caption{$\sigma=0.03$}
        \end{subfigure}
        &
        \begin{subfigure}{0.24\linewidth}
            \centering
            \frame{\includegraphics[width=\linewidth]{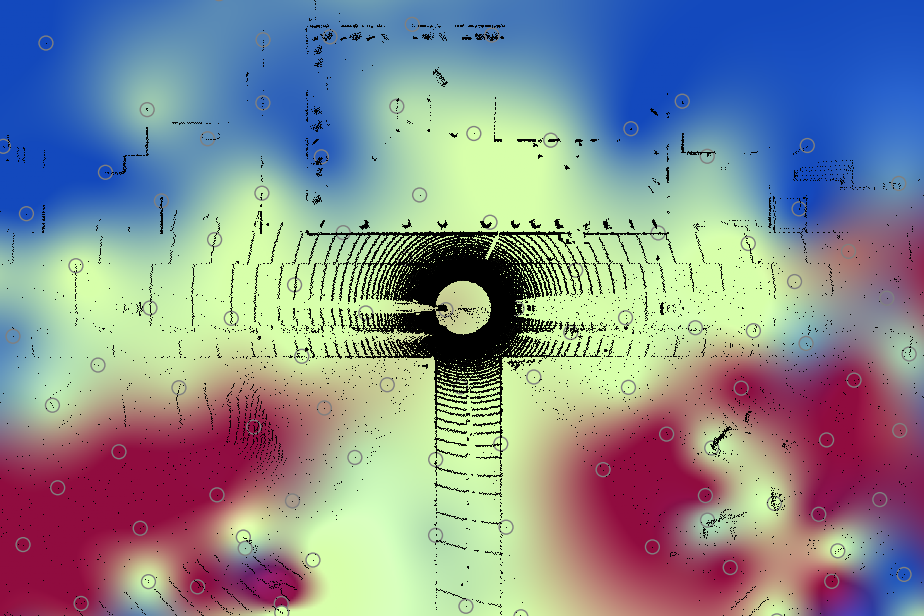}}
            \caption{$\sigma=0.07$}
        \end{subfigure}
        \\
        \\
        \textcolor{real_img}{\rule{2.4mm}{2.4mm}} 0.33,
        \textcolor{syn_img}{\rule{2.4mm}{2.4mm}} 0.44,
        \textcolor{misc_img}{\rule{2.4mm}{2.4mm}} 0.23
        &
        \textcolor{real_img}{\rule{2.4mm}{2.4mm}} 0.35,
        \textcolor{syn_img}{\rule{2.4mm}{2.4mm}} 0.38,
        \textcolor{misc_img}{\rule{2.4mm}{2.4mm}} 0.27
        &
        \textcolor{real_img}{\rule{2.4mm}{2.4mm}} 0.26,
        \textcolor{syn_img}{\rule{2.4mm}{2.4mm}} 0.42,
        \textcolor{misc_img}{\rule{2.4mm}{2.4mm}} 0.32
        &
        \textcolor{real_img}{\rule{2.4mm}{2.4mm}} 0.24,
        \textcolor{syn_img}{\rule{2.4mm}{2.4mm}} 0.44,
        \textcolor{misc_img}{\rule{2.4mm}{2.4mm}} 0.32
        \\
        \begin{subfigure}{0.24\linewidth}
            \centering
            \frame{\includegraphics[width=\linewidth]{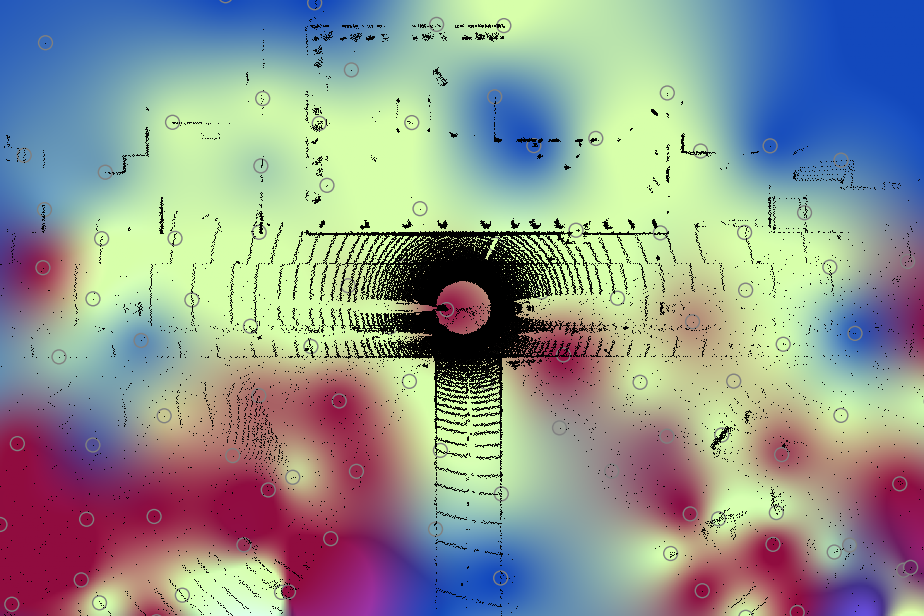}}
            \caption{$\sigma=0.1$}
        \end{subfigure}
        &
        \begin{subfigure}{0.24\linewidth}
            \centering
            \frame{\includegraphics[width=\linewidth]{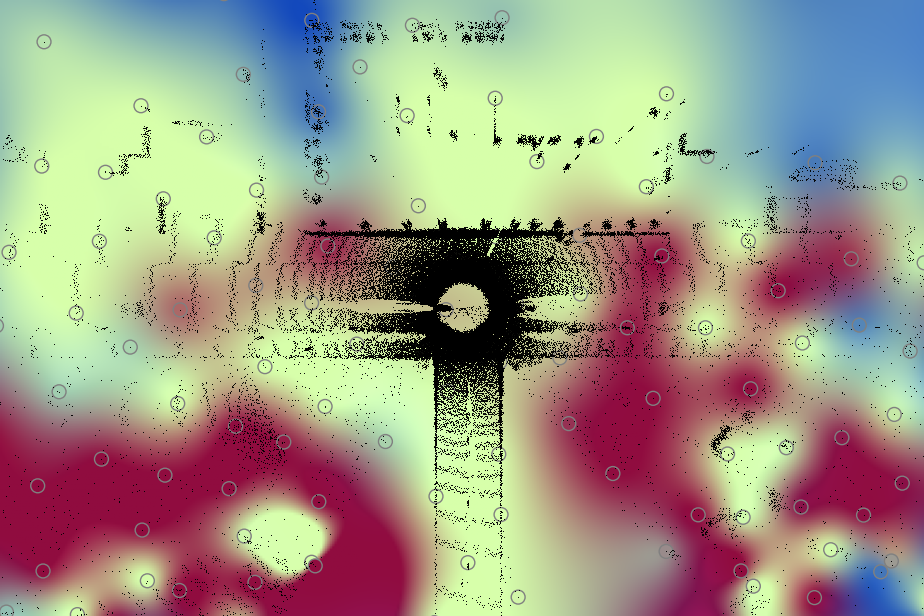}}
            \caption{$\sigma=0.3$}
        \end{subfigure}
        &
        \begin{subfigure}{0.24\linewidth}
            \centering
            \frame{\includegraphics[width=\linewidth]{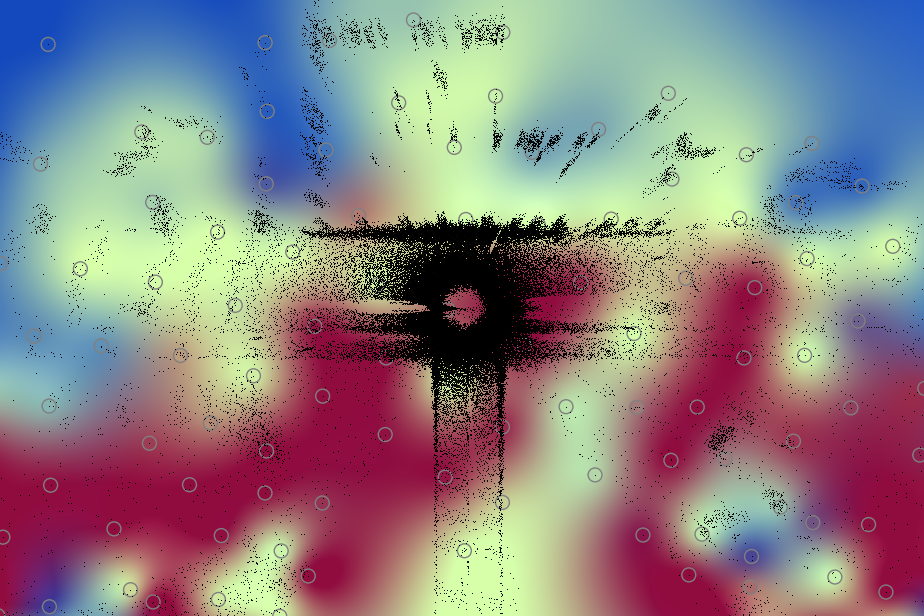}}
            \caption{$\sigma=0.7$}
        \end{subfigure}
        &
        \begin{subfigure}{0.24\linewidth}
            \centering
            \frame{\includegraphics[width=\linewidth]{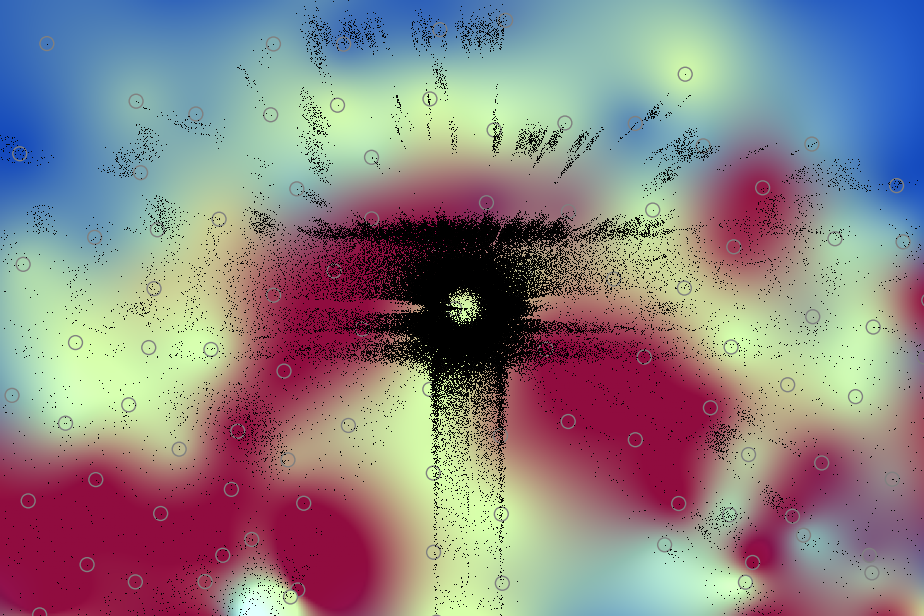}}
            \caption{$\sigma=1.0$}
        \end{subfigure}
        \\
        \\
        \textcolor{real_img}{\rule{2.4mm}{2.4mm}} 0.12,
        \textcolor{syn_img}{\rule{2.4mm}{2.4mm}} 0.38,
        \textcolor{misc_img}{\rule{2.4mm}{2.4mm}} 0.49
        &
        \textcolor{real_img}{\rule{2.4mm}{2.4mm}} 0.03,
        \textcolor{syn_img}{\rule{2.4mm}{2.4mm}} 0.27,
        \textcolor{misc_img}{\rule{2.4mm}{2.4mm}} 0.70
        &
        \textcolor{real_img}{\rule{2.4mm}{2.4mm}} 0.02,
        \textcolor{syn_img}{\rule{2.4mm}{2.4mm}} 0.20,
        \textcolor{misc_img}{\rule{2.4mm}{2.4mm}} 0.78
        &
        \\
        \begin{subfigure}{0.24\linewidth}
            \centering
            \frame{\includegraphics[width=\linewidth]{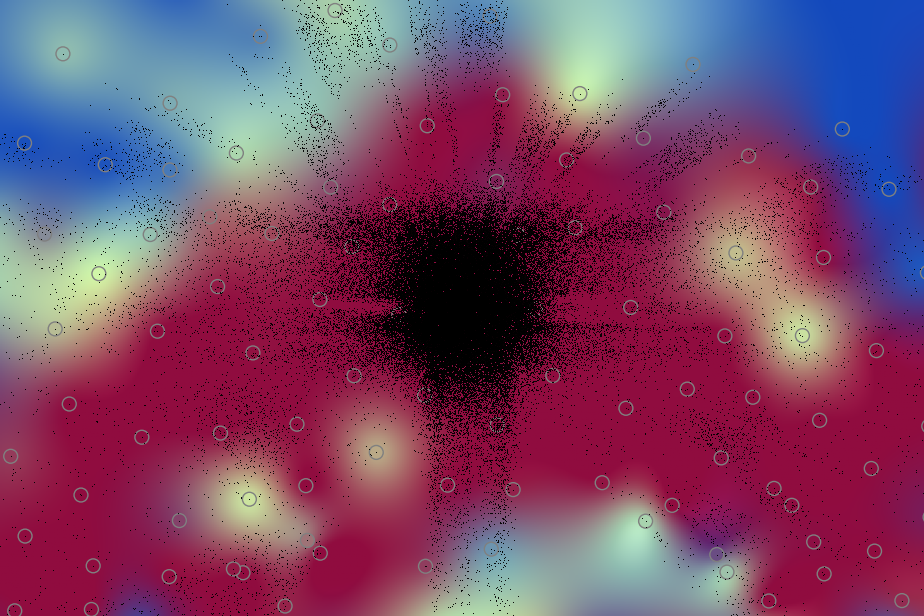}}
            \caption{$\sigma=3.0$}
        \end{subfigure}
        &
        \begin{subfigure}{0.24\linewidth}
            \centering
            \frame{\includegraphics[width=\linewidth]{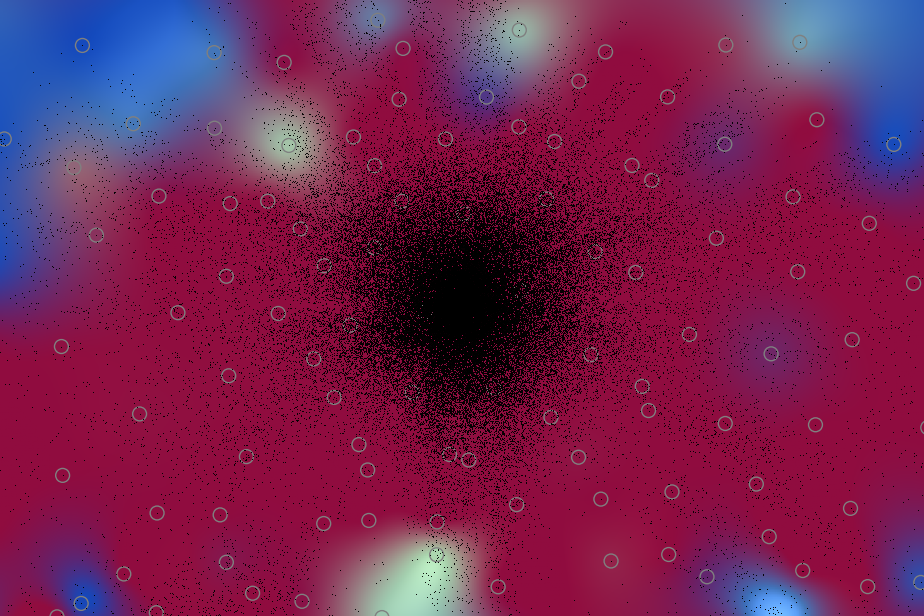}}
            \caption{$\sigma=7.0$}
        \end{subfigure}
        &
        \begin{subfigure}{0.24\linewidth}
            \centering
            \frame{\includegraphics[width=\linewidth]{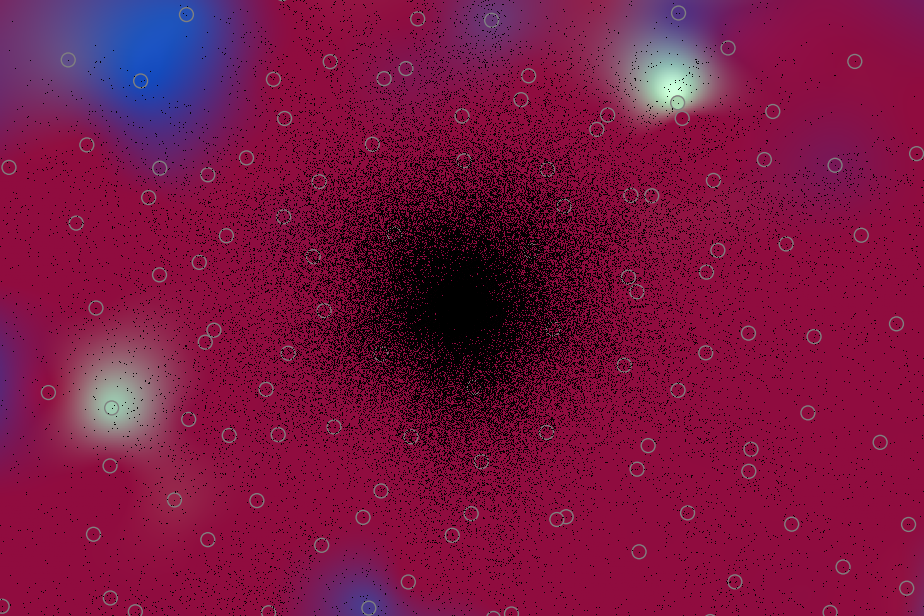}}
            \caption{$\sigma=10.0$}
        \end{subfigure}
        &
        \hspace{1cm}
        \makecell[l]{
            \textcolor{real_img}{\rule{2.4mm}{2.4mm}} \Real{} \\
            \textcolor{syn_img}{\rule{2.4mm}{2.4mm}} \Syn{} \\
            \textcolor{misc_img}{\rule{2.4mm}{2.4mm}} \Misc{}
        }

    \end{tabular}

    \caption{
        \textbf{CARLA Sample with Gaussian Noise}:
        Example image series of a single CARLA sample with additive Gaussian noise of varying standard deviation $\sigma$.
        The colors show the probabilities for each category at each local region.
        The numbers above each image show the mean score per category for the entire scene.
        When adding noise with small $\sigma$, the sample appears more realistic but when the noise gets too strong, the data does not contain any structuring anymore.
    }
    \label{fig:qualitative_carla_noise}

\end{figure*}

\begin{figure*}
    \centering

    \begin{tabular}{cccc}

        & Scene 1 & Scene 2 & Scene 3
        \\
        \\
        %
        \rotatebox{90}{Ground Truth}
        &
        \frame{\includegraphics[width=0.24\linewidth]{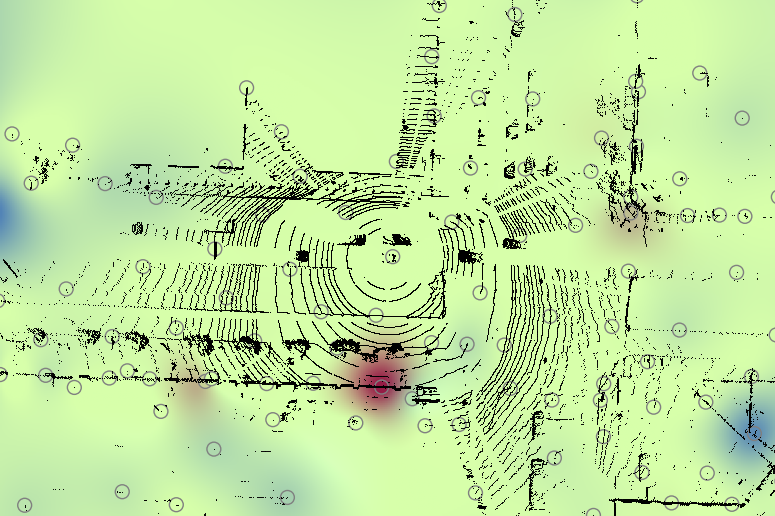}}
        &
        \frame{\includegraphics[width=0.24\linewidth]{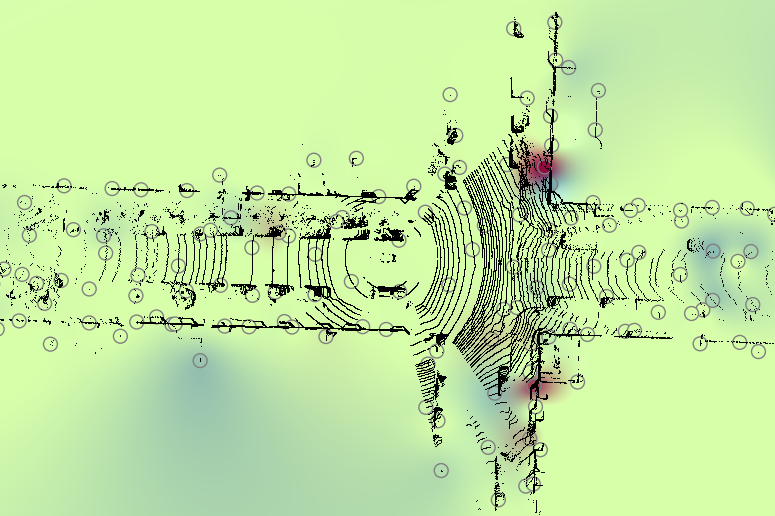}}
        &
        \frame{\includegraphics[width=0.24\linewidth]{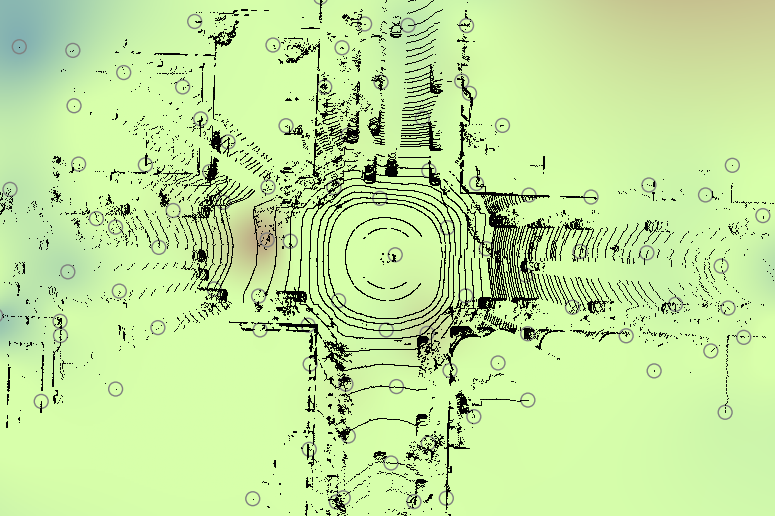}}
        \\
        %
        \rotatebox{90}{L1-CNN}
        &
        \frame{\includegraphics[width=0.24\linewidth]{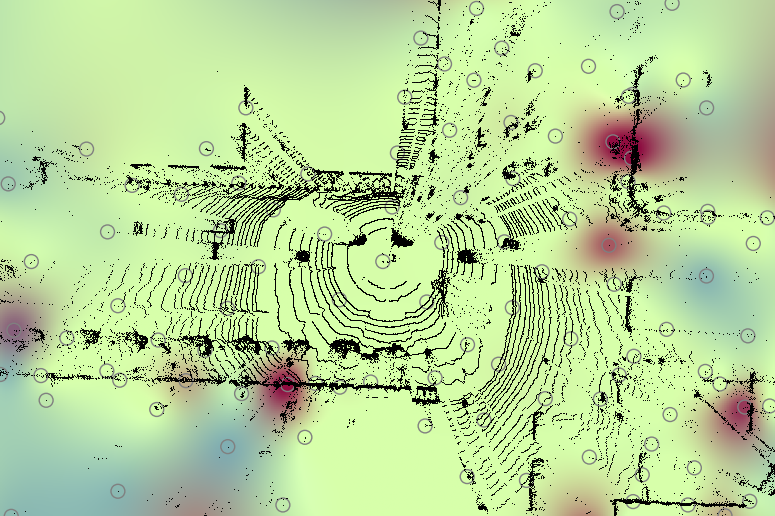}}
        &
        \frame{\includegraphics[width=0.24\linewidth]{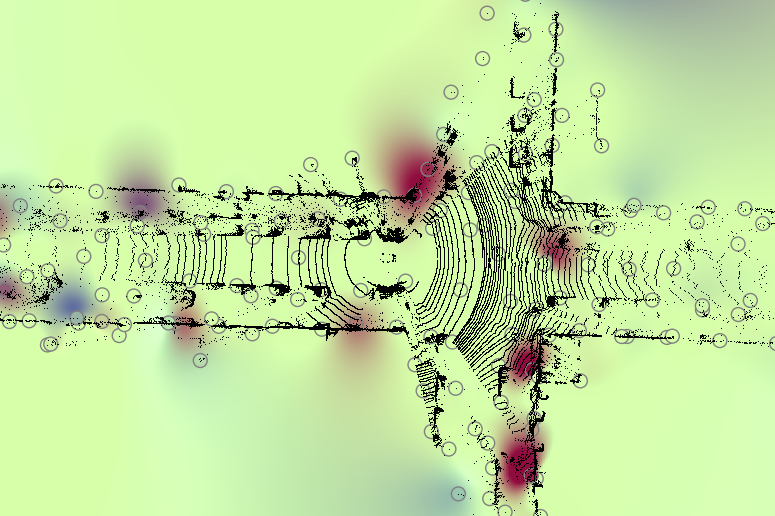}}
        &
        \frame{\includegraphics[width=0.24\linewidth]{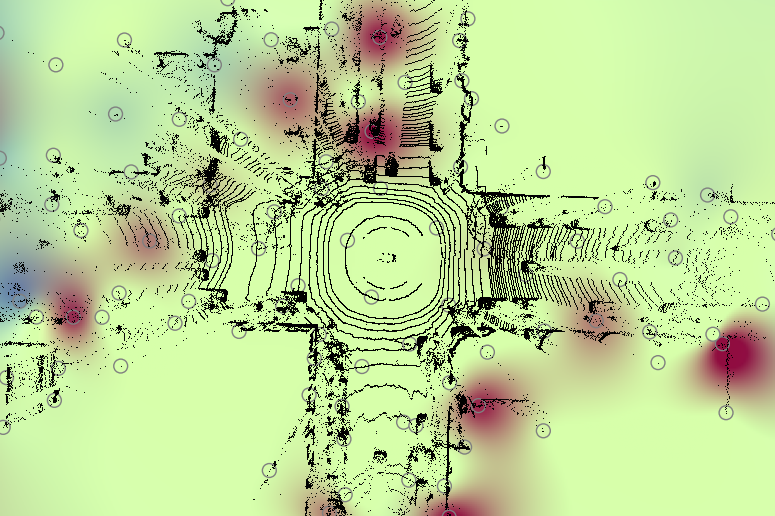}}
        \\
        %
        \rotatebox{90}{Bilinear}
        &
        \frame{\includegraphics[width=0.24\linewidth]{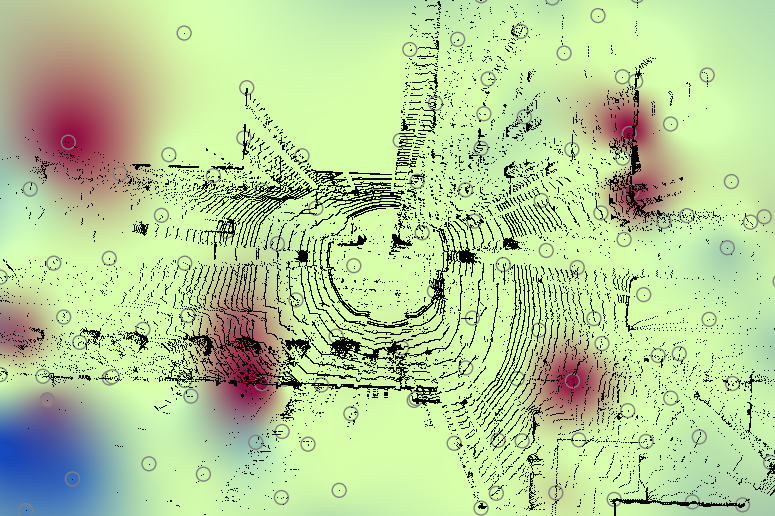}}
        &
        \frame{\includegraphics[width=0.24\linewidth]{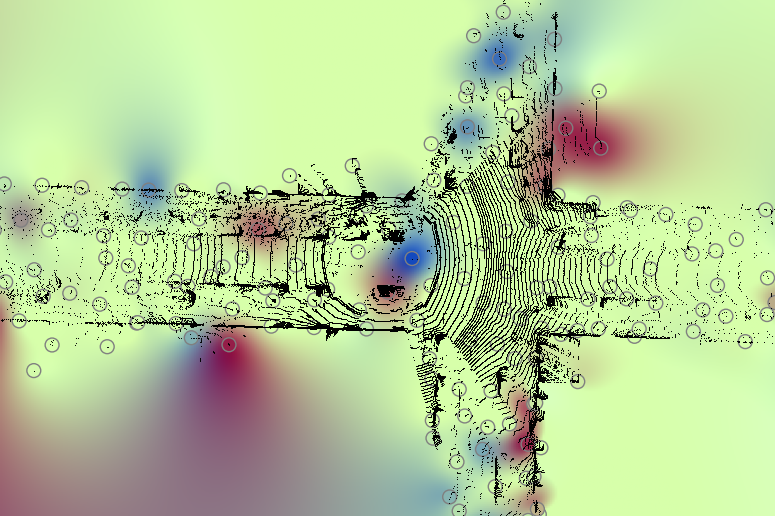}}
        &
        \frame{\includegraphics[width=0.24\linewidth]{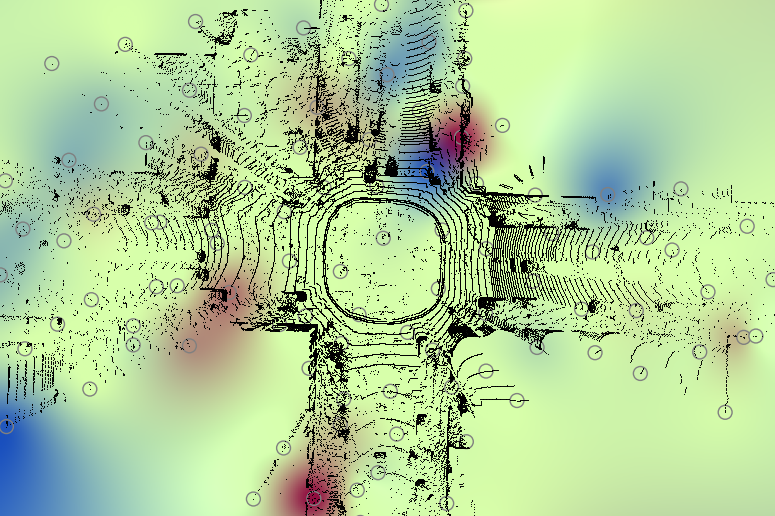}}
        \\
        %
        \rotatebox{90}{L2-CNN}
        &
        \frame{\includegraphics[width=0.24\linewidth]{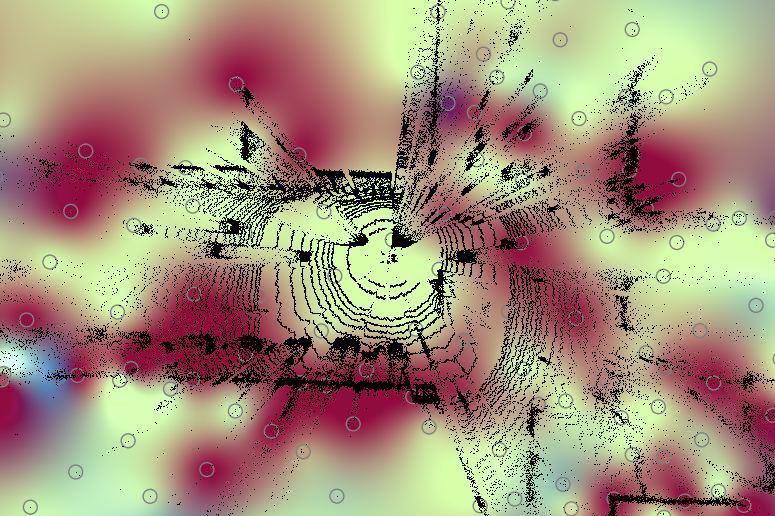}}
        &
        \frame{\includegraphics[width=0.24\linewidth]{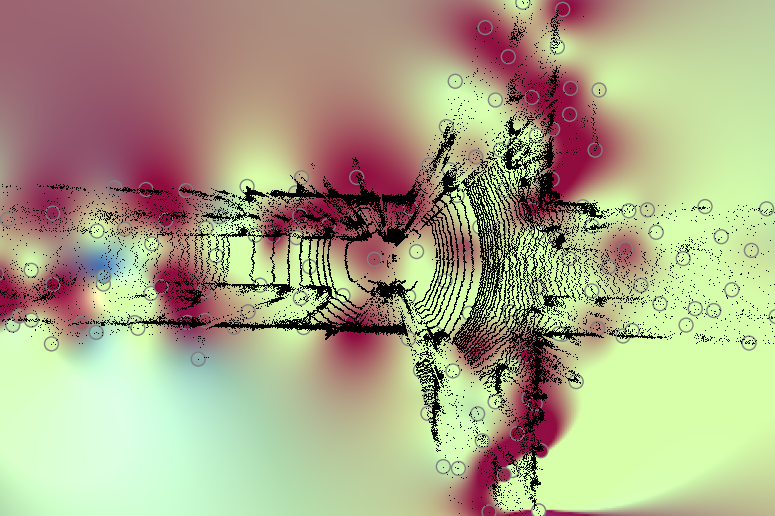}}
        &
        \frame{\includegraphics[width=0.24\linewidth]{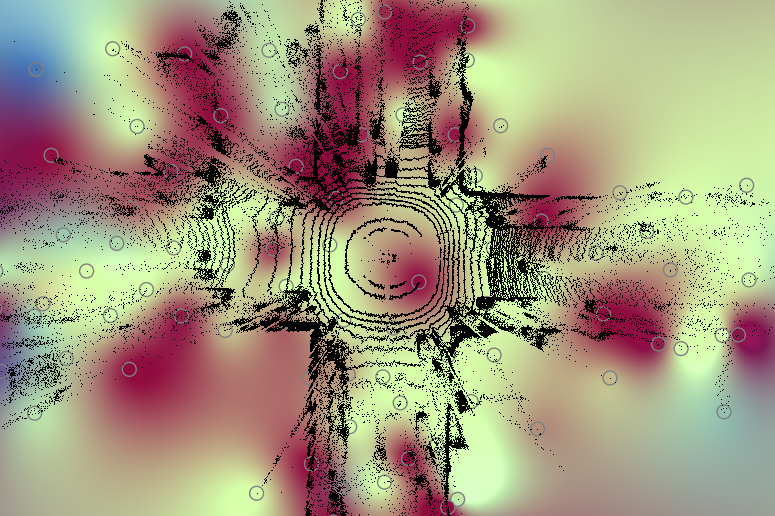}}
        \\
        %
        \rotatebox{90}{GAN}
        &
        \frame{\includegraphics[width=0.24\linewidth]{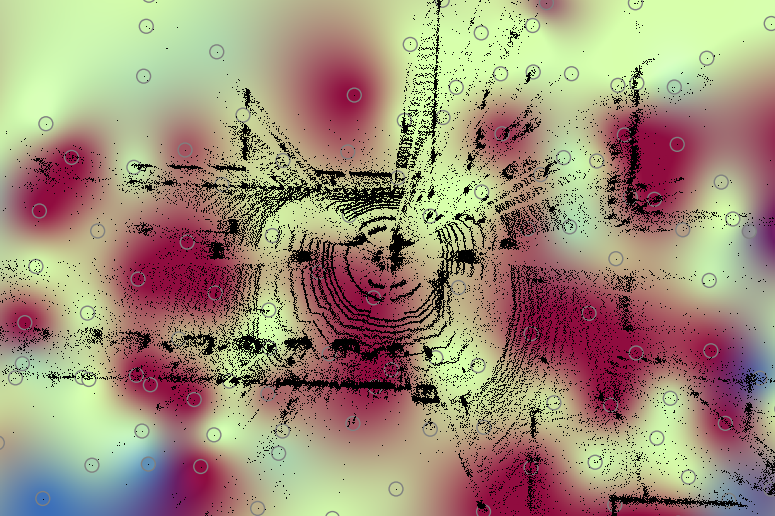}}
        &
        \frame{\includegraphics[width=0.24\linewidth]{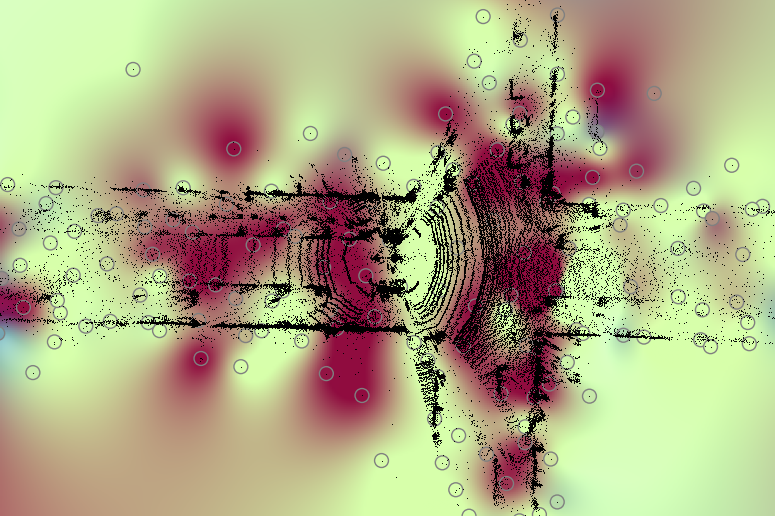}}
        &
        \frame{\includegraphics[width=0.24\linewidth]{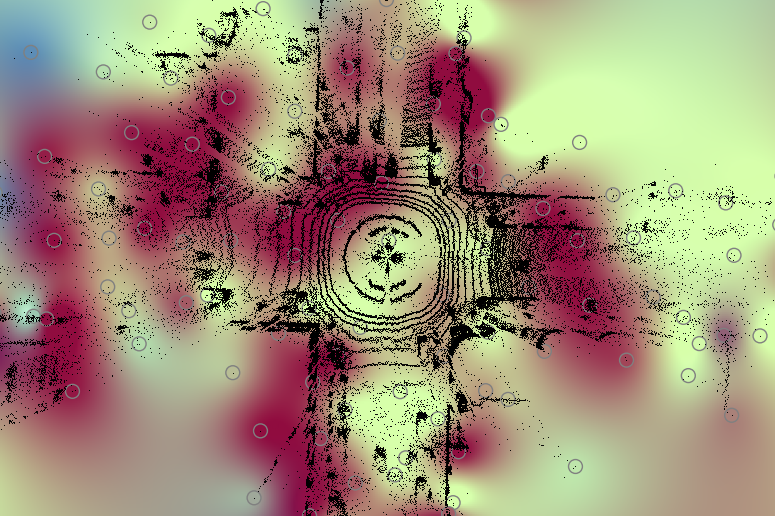}}

    \end{tabular}

    \caption{
        \textbf{PandaSet Up-Sampling}:
        Shown are three example scenes from the PandaSet test split.
        The first row shows the high-resolution Ground Truth, \ie target.
        The four rows below show the corresponding reconstructions of different techniques for $4\!~\!\times$ up-sampling.
        Note that the query points are sampled at different locations for each image, leading to varying score computation for similar regions, and that the height above ground of the query points is not encoded in this visualization.
        (\textcolor{real_img}{\rule{2.4mm}{2.4mm}}~\Real{}, \textcolor{syn_img}{\rule{2.4mm}{2.4mm}}~\Syn{}, \textcolor{misc_img}{\rule{2.4mm}{2.4mm}}~\Misc{})
    }
    \label{fig:qualitative_pandaset_upsampling}

\end{figure*}

\begin{figure*}
    \centering

    \begin{tabular}{ccc}

        \frame{\includegraphics[width=0.3\linewidth]{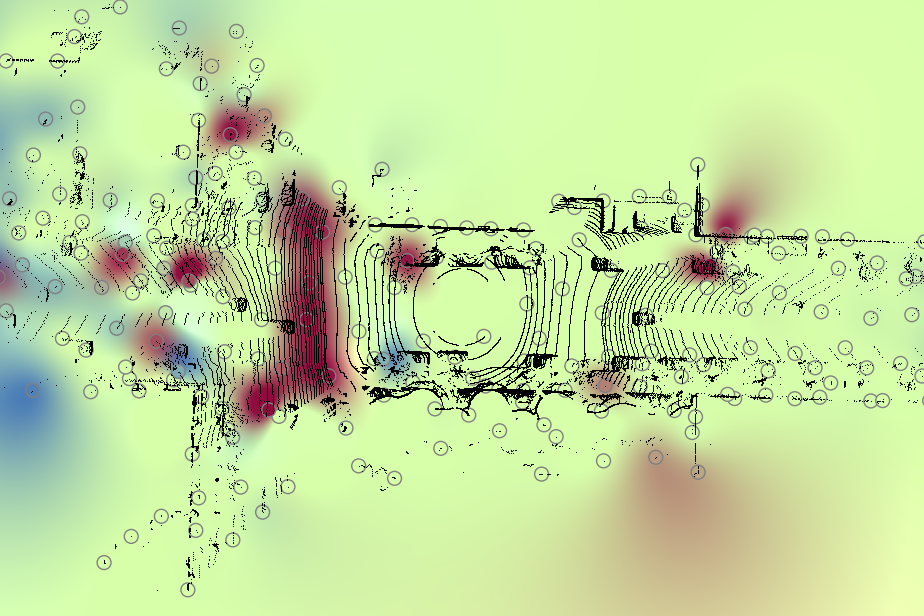}} &
        \frame{\includegraphics[width=0.3\linewidth]{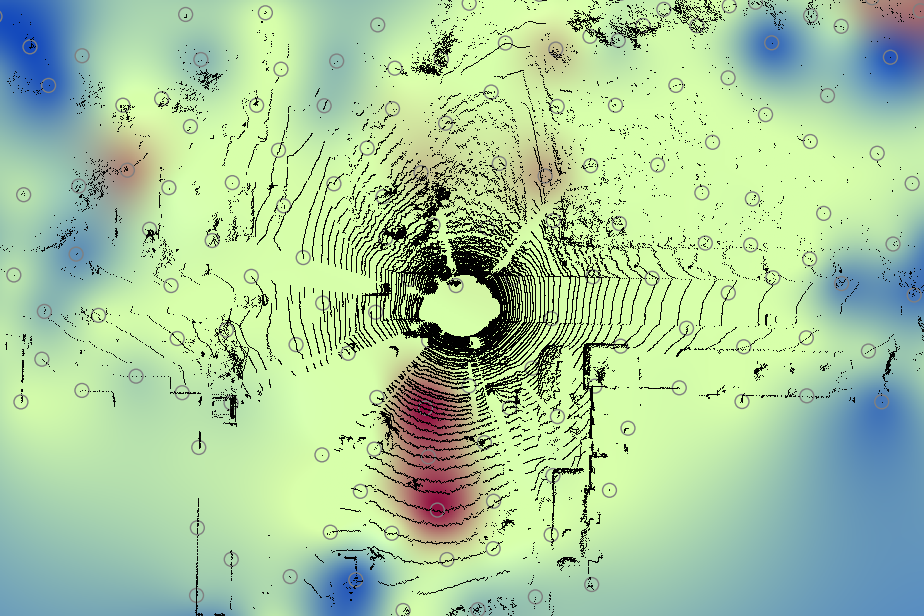}} &
        \frame{\includegraphics[width=0.3\linewidth]{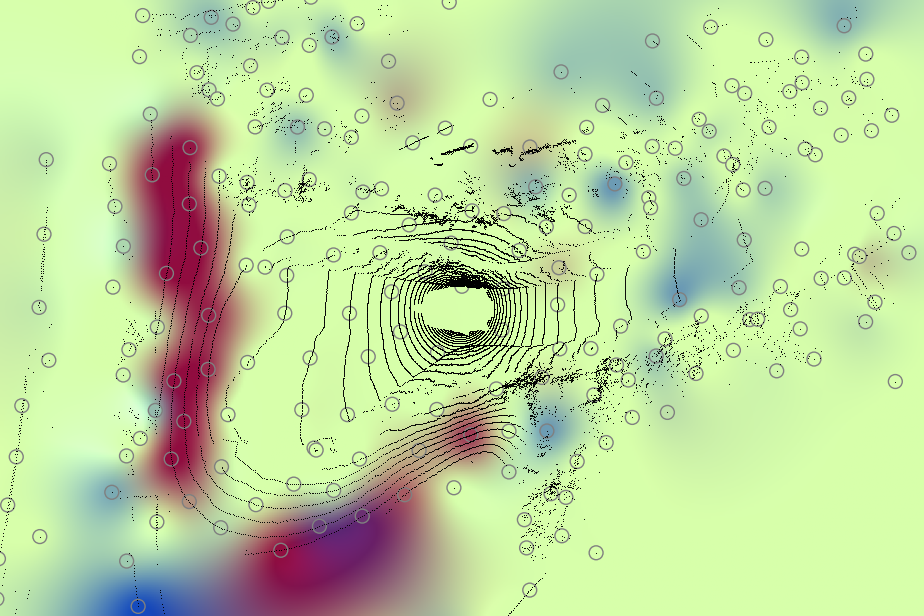}} \\

        \frame{\includegraphics[width=0.3\linewidth]{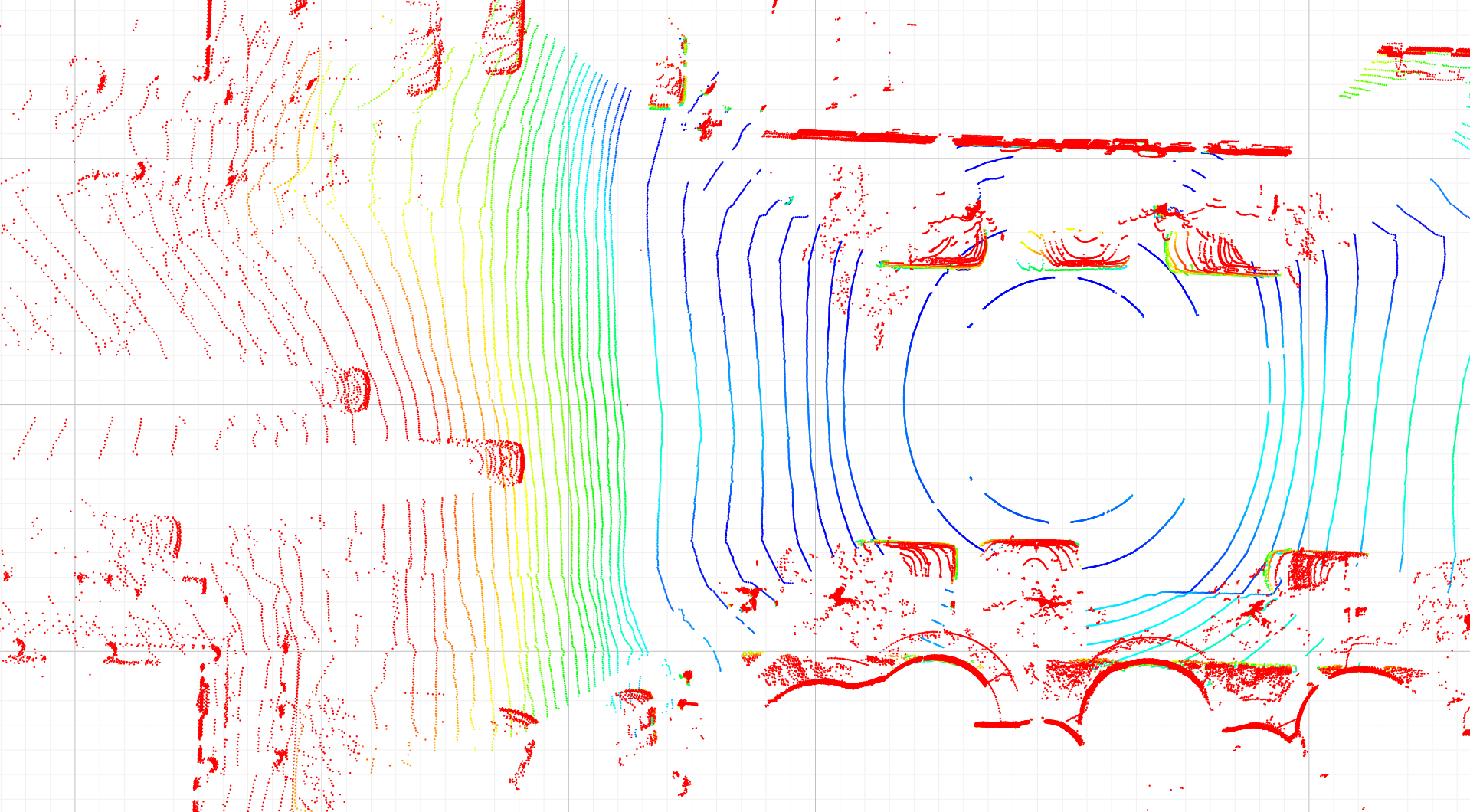}} &
        \frame{\includegraphics[width=0.3\linewidth]{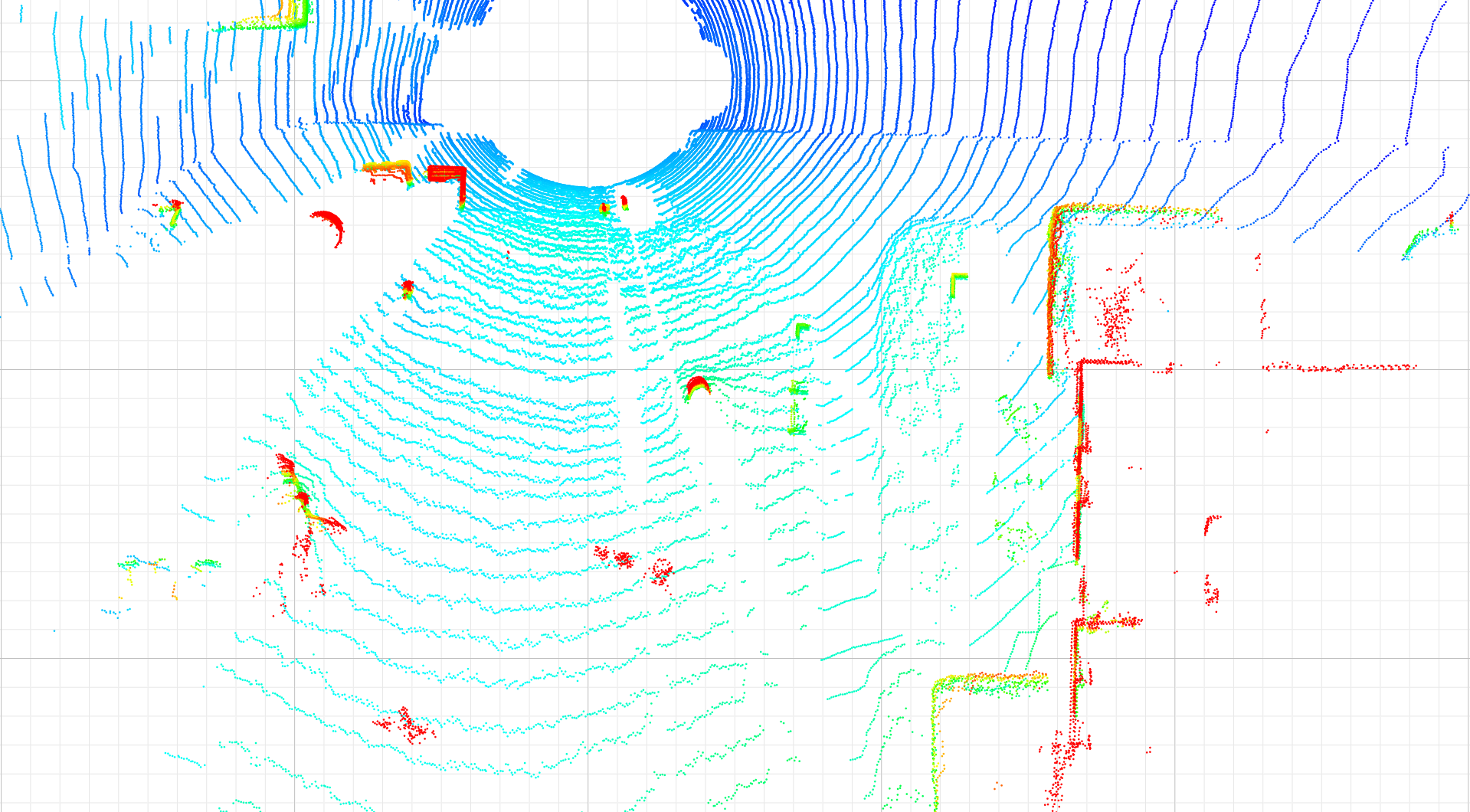}} &
        \frame{\includegraphics[width=0.3\linewidth]{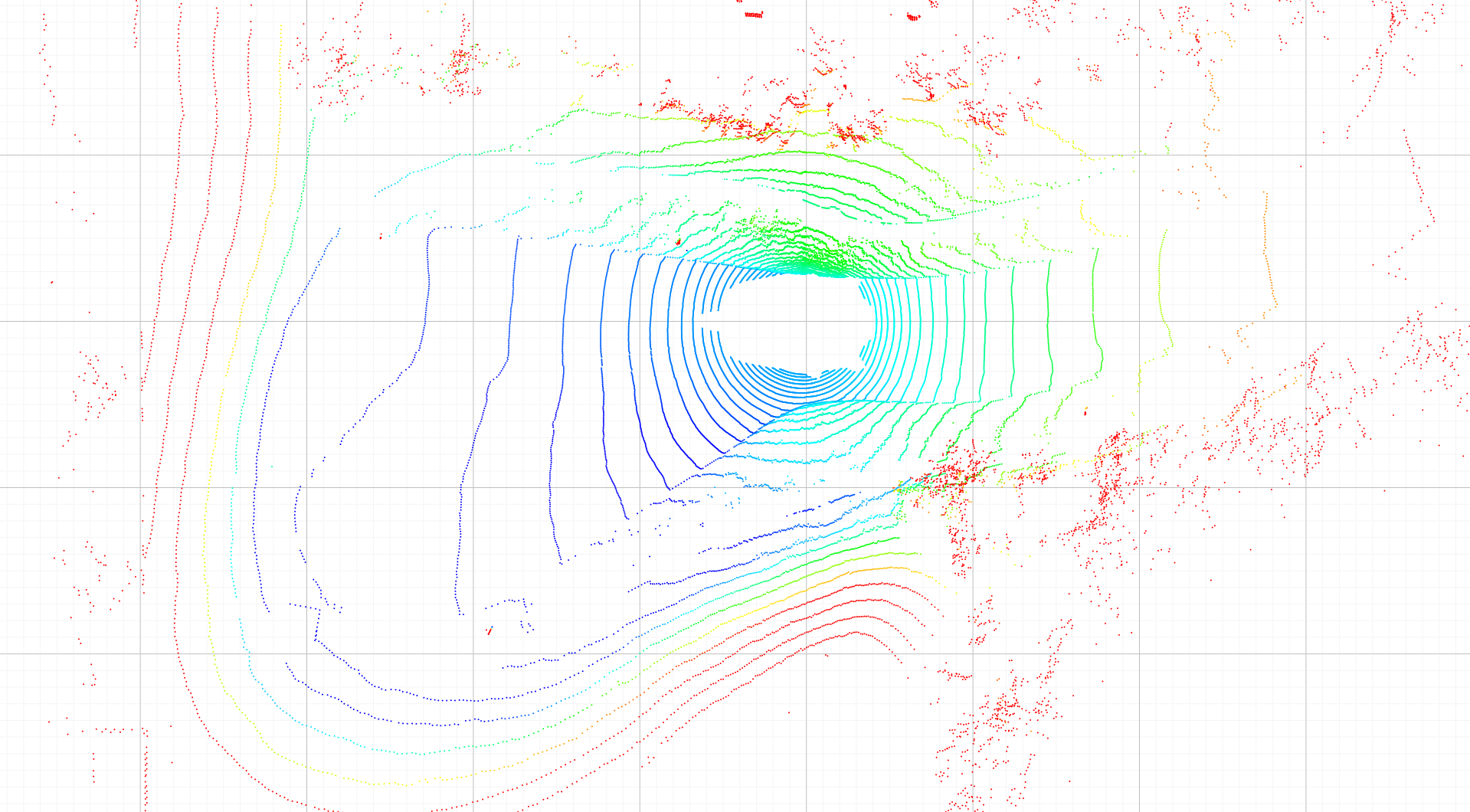}} \\

        \frame{\includegraphics[width=0.3\linewidth]{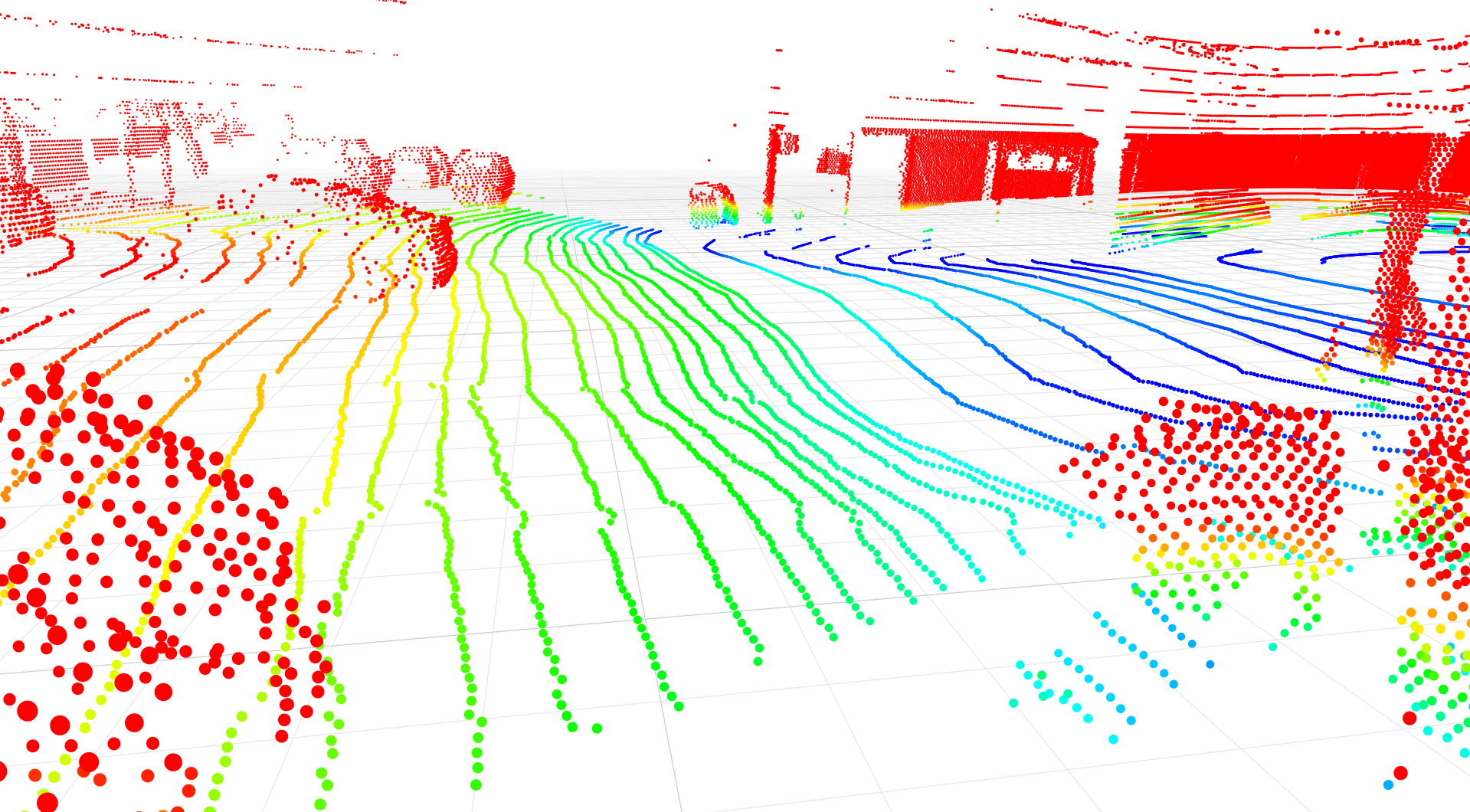}} &
        \frame{\includegraphics[width=0.3\linewidth]{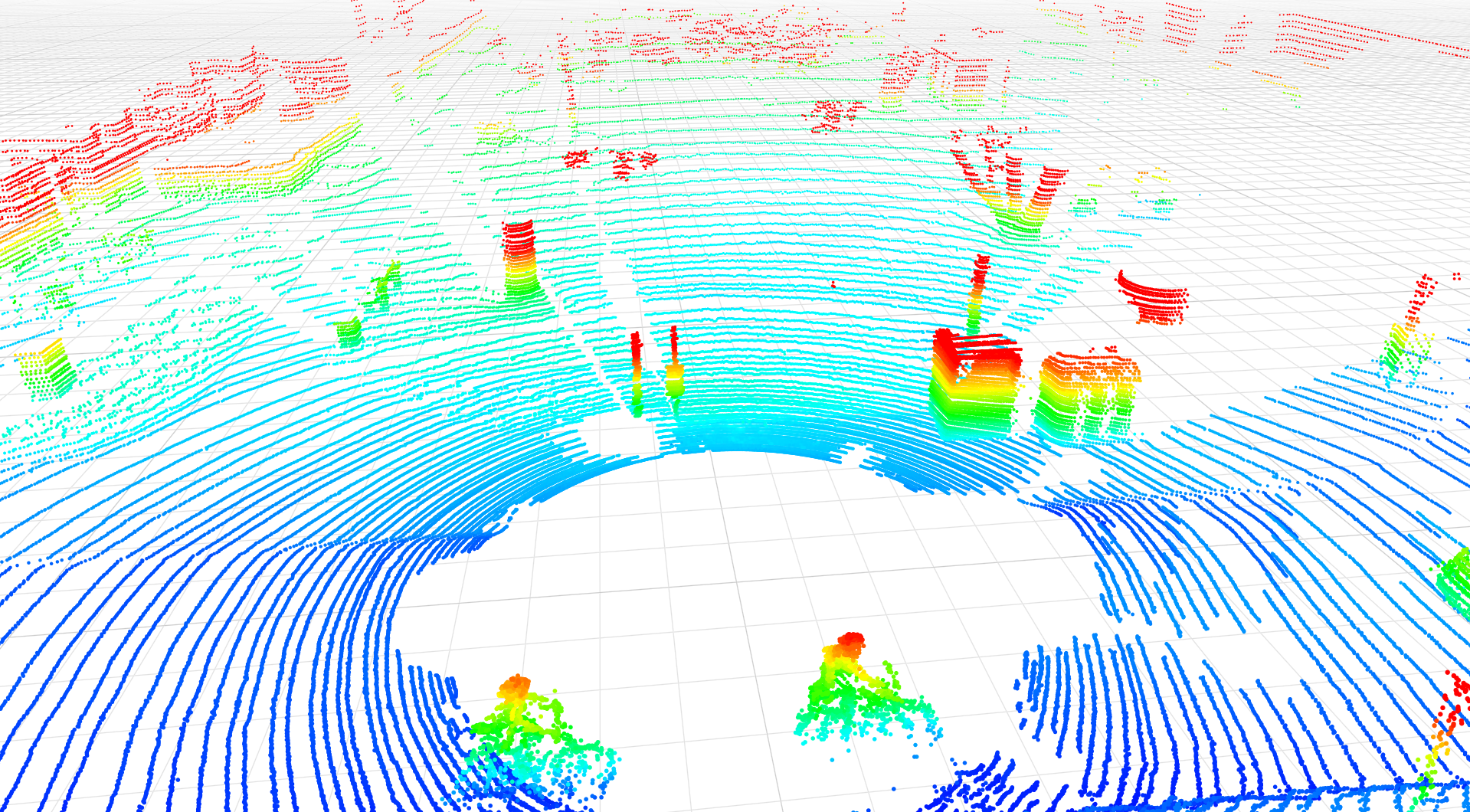}} &
        \frame{\includegraphics[width=0.3\linewidth]{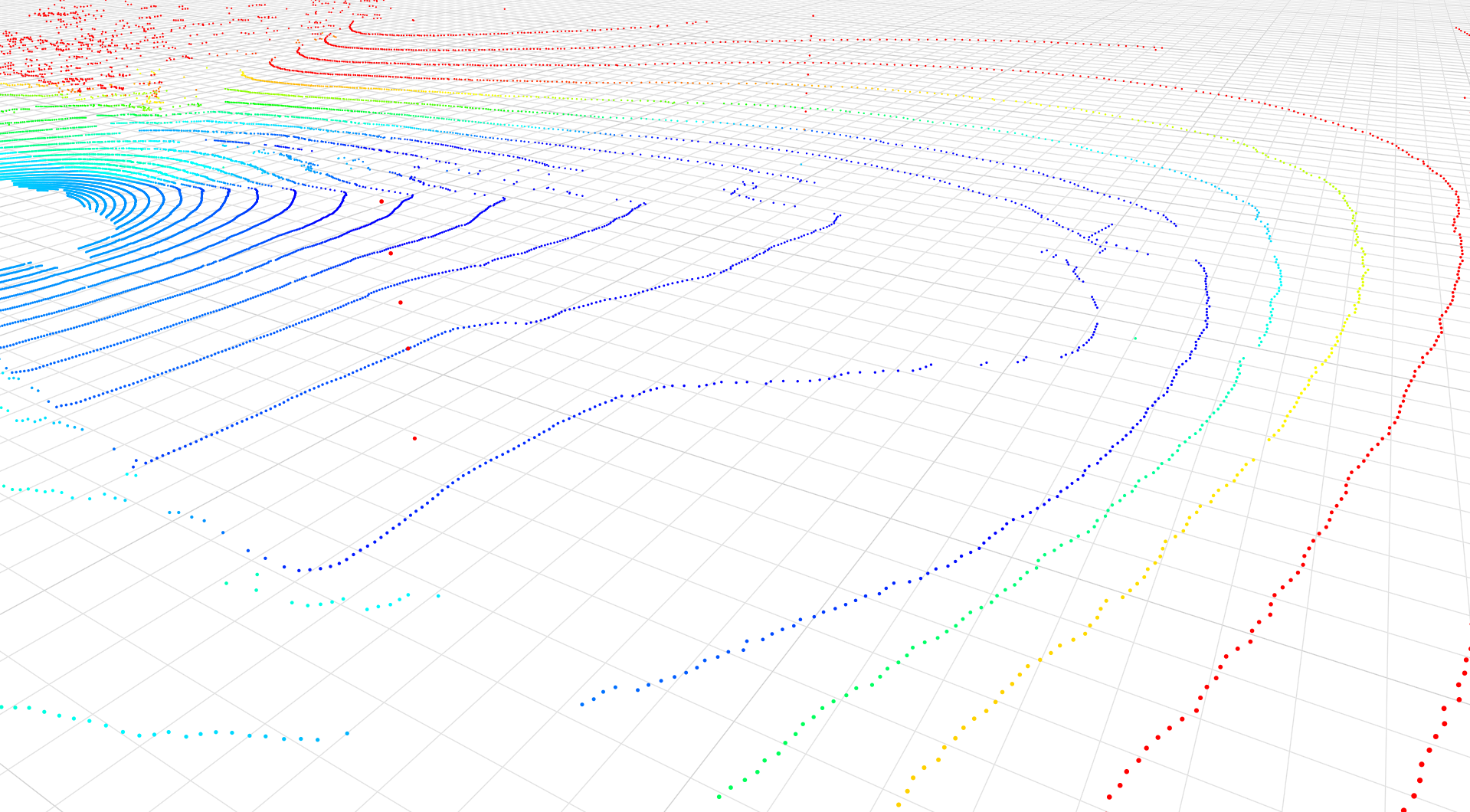}} \\

        \begin{subfigure}{0.3\linewidth}
            \centering
            \frame{\includegraphics[width=\linewidth]{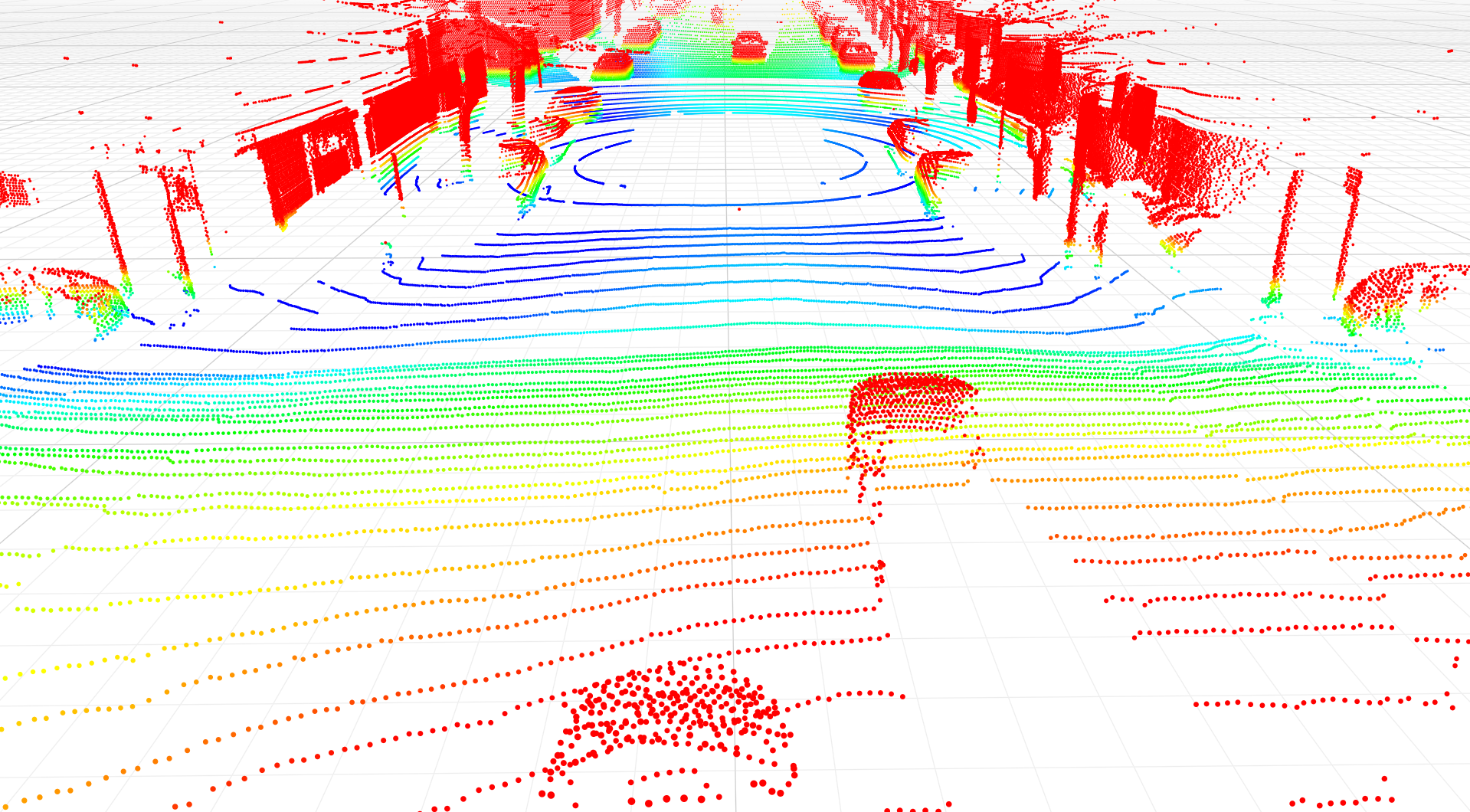}}
            \caption{\label{fig:experiments_anomaly_3d_pandaset}PandaSet}
        \end{subfigure}
        &
        \begin{subfigure}{0.3\linewidth}
            \centering
            \frame{\includegraphics[width=\linewidth]{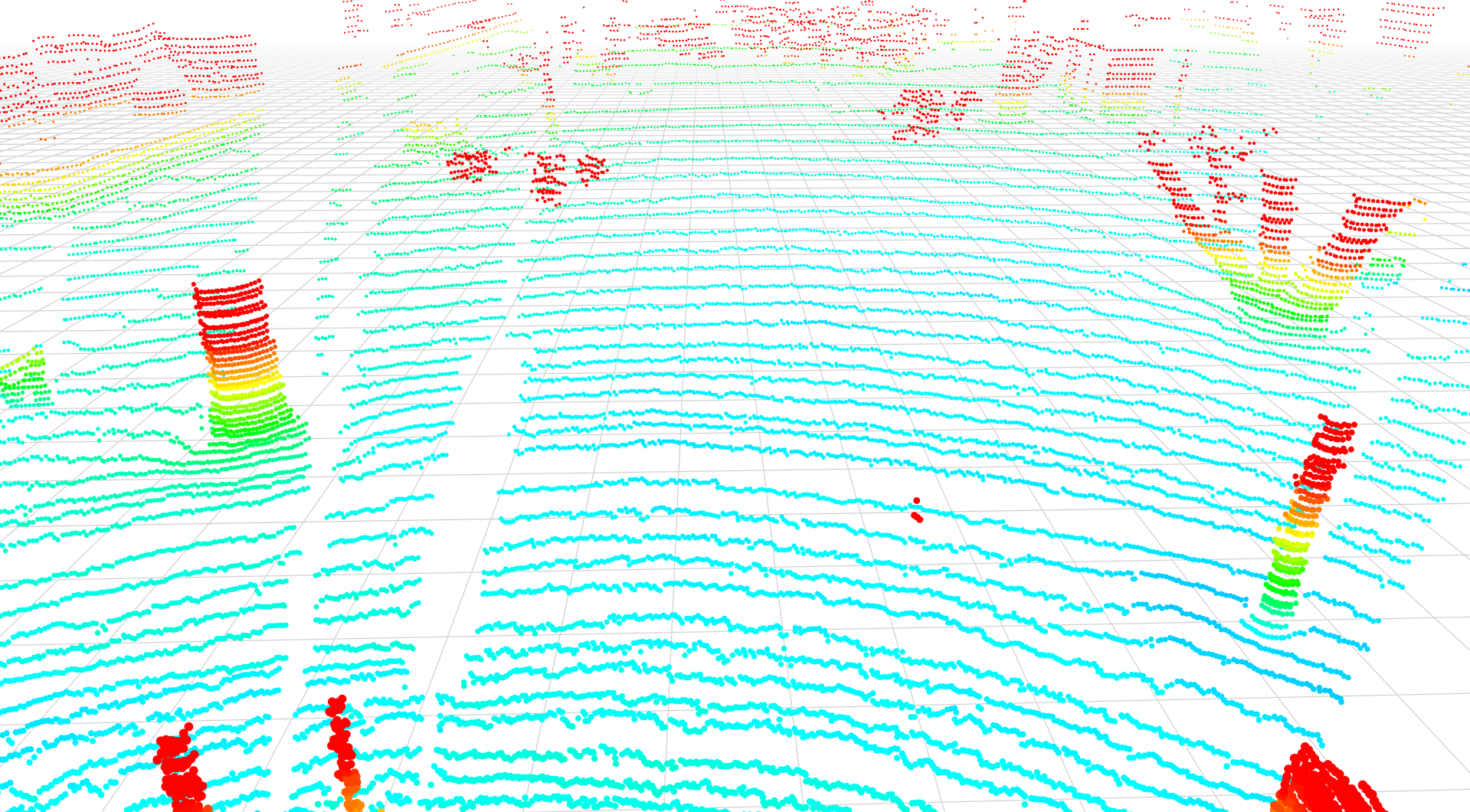}}
            \caption{\label{fig:experiments_anomaly_3d_kitti}KITTI}
        \end{subfigure}
        &
        \begin{subfigure}{0.3\linewidth}
            \centering
            \frame{\includegraphics[width=\linewidth]{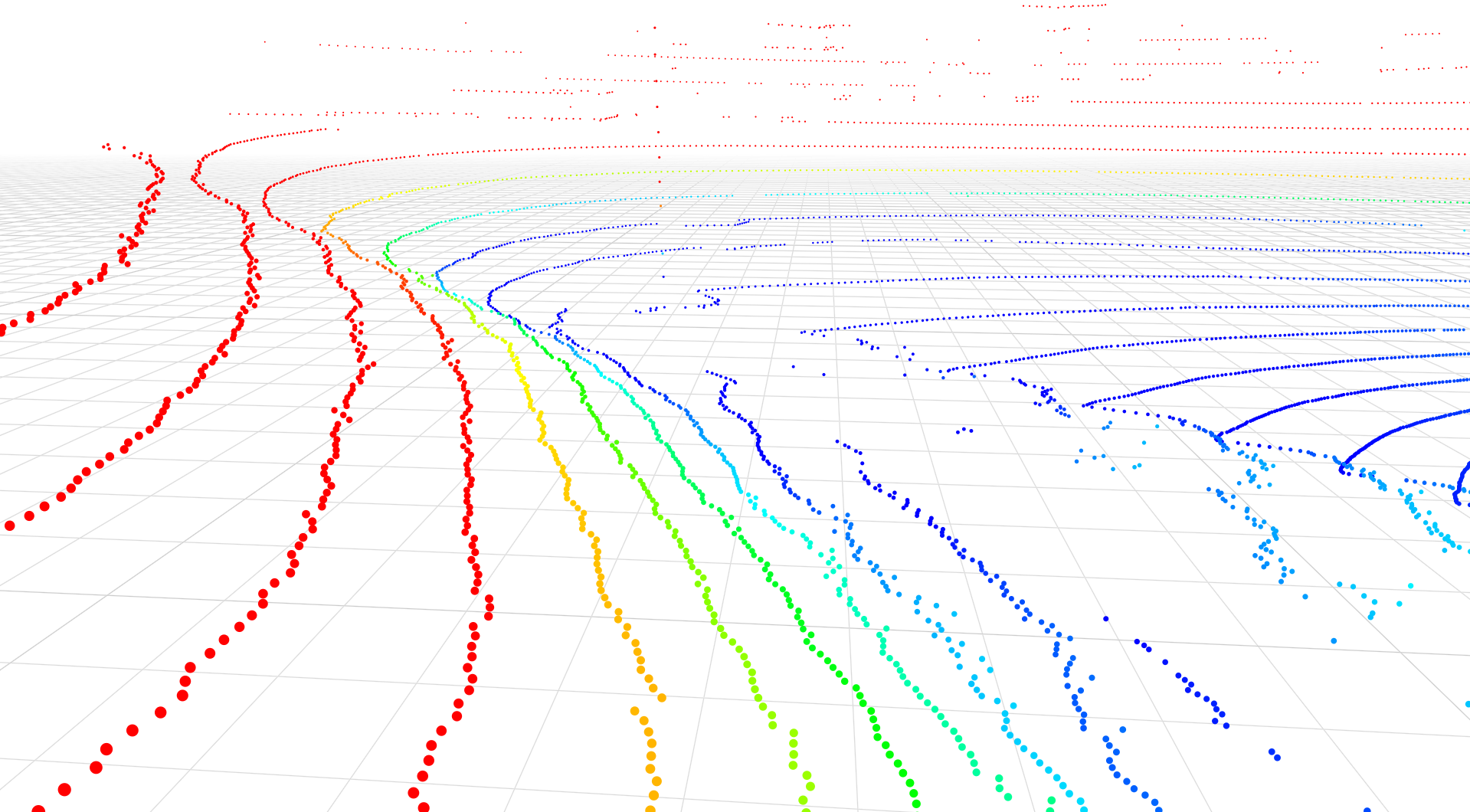}}
            \caption{\label{fig:experiments_anomaly_3d_nuscenes}nuScenes}
        \end{subfigure}

    \end{tabular}

    \caption{
        \textbf{Localization of Anomalies}:
        Shown are the same example scenes from the main paper with low \Real{} scores.
        The upper images show the color-coded metric results, the lower images are 3D visualization of the same scene for better understanding of the scene contents.
        In (\subref{fig:experiments_anomaly_3d_pandaset}), the metric marks a road section with extreme elevation changes.
        The street is covered with bumps.
        Additionally, there is a steep elevation gain in the rear of the ego-vehicle (left side).
        In the lower half of (\subref{fig:experiments_anomaly_3d_kitti}), the metric highlights seemingly floating branches of two trees that enter the LiDAR field-of-view from above.
        The height-encoded color shows the red branch clusters floating above the cyan road surface.
        (\subref{fig:experiments_anomaly_3d_nuscenes})~shows an unusual scene in a dead end road with steep hills surrounding the car.
    }
    \label{fig:experiments_anomaly_3d}

\end{figure*}

This section provides additional visualizations.
\figref{fig:qualitative_carla_noise}~shows the qualitative results of the realism metric when being applied to one CARLA sample with varying additive noise.
The quantitative results over the whole test split are presented in~\secref{sec:experiments_transition}.
\figref{fig:qualitative_pandaset_upsampling}~shows example scenes from the reconstruction experiments that are presented in~\secref{sec:experiments_reconstruction}.
\figref{fig:experiments_anomaly_3d}~illustrates the example scenes from \secref{sec:experiments_anomaly} in 3D for better visualization of the detected anomalies.

\end{document}